\documentclass[11pt]{article}
\usepackage[utf8]{inputenc}
\usepackage[margin=1in]{geometry}
\usepackage{graphicx} 
\usepackage{amsfonts}
\usepackage{amsmath}
\usepackage{dirtytalk}
\usepackage{algorithm}
\usepackage{mathtools}
\usepackage{xcolor}
\usepackage{amssymb}
\usepackage[title]{appendix}
\usepackage[noend]{algpseudocode}
\usepackage{amsthm}
\usepackage{bigints}
\usepackage{thm-restate}
\usepackage[]{hyperref}
\usepackage{cleveref}
\usepackage{appendix}
\usepackage{makecell}
\usepackage{mathtools}
\usepackage{subfigure}
\usepackage{booktabs}
\usepackage{bbding}
\usepackage{wrapfig}
\usepackage{bbm}
\usepackage{authblk}
\usepackage[framemethod=TikZ]{mdframed}
\mdfdefinestyle{MyFrame2}{%
    linecolor=white,
    outerlinewidth=1pt,
    roundcorner=2pt,
    innertopmargin=5pt, %
    innerbottommargin=5pt, %
    innerrightmargin=10pt,
    innerleftmargin=10pt,
    backgroundcolor=black!3!white}
\declaretheorem[name=Theorem,numberwithin=section]{thm}

\declaretheorem[name=Claim,sibling=thm]{claim}

\usepackage[numbers]{natbib}
\newcommand{\man}{\mathcal{M}}
\newcommand{\norm}[1]{\left\lVert#1\right\rVert}

\title{\textrm{EmbedOR}: Provable Cluster-Preserving Visualizations with Curvature-Based Stochastic Neighbor Embeddings}

\author[1]{Tristan Luca Saidi\thanks{\url{tls2160@columbia.edu}}}
\author[2]{Abigail Hickok\thanks{\url{abigail.hickok@yale.edu}}}
\author[3]{Bastian Rieck\thanks{\url{bastian@rieck.me}}}
\author[4,1,5]{Andrew J. Blumberg\thanks{\url{andrew.blumberg@columbia.edu}}}

\affil[1]{Department of Computer Science, Columbia University}
\affil[2]{Department of Statistics \& Data Science, Wu Tsai Institute, Yale University}
\affil[3]{Department of Machine Learning, University of Fribourg}
\affil[4]{Department of Mathematics, Columbia University}
\affil[5]{Irving Institute for Cancer Dynamics, Columbia University}

\begin{document}

\maketitle

\begin{abstract}
    Stochastic Neighbor Embedding (SNE) algorithms like UMAP and tSNE often produce visualizations that do not preserve the geometry of noisy and high dimensional data. In particular, they can spuriously separate connected components of the underlying data submanifold and can fail to find clusters in well-clusterable data. To address these limitations, we propose EmbedOR, a SNE algorithm that incorporates  discrete graph curvature. Our algorithm stochastically embeds the data using a curvature-enhanced distance metric that emphasizes underlying cluster structure. Critically, we prove that the EmbedOR distance metric extends consistency results for tSNE
    to a much broader class of datasets. We also describe extensive experiments on synthetic and real data that demonstrate the visualization and geometry-preservation capabilities of EmbedOR. We find that, unlike other SNE algorithms and UMAP, EmbedOR is much less likely to fragment continuous, high-density regions of the data. Finally, we demonstrate that the EmbedOR distance metric can be used as a tool to annotate existing visualizations to identify fragmentation and provide deeper insight into the underlying geometry of the data.
\end{abstract}

\maketitle


\section{Introduction}
Stochastic Neighbor Embedding (SNE) algorithms like UMAP \cite{umap} and tSNE \cite{tsne} have gained a lot of attention in the last decade due to their ability to produce visualizations of high-dimensional data. They have thus found applications in a range of of fields, including single-cell genomics \cite{becht2019dimensionality, kobak2019art}, neural-network interpretability \cite{carter2019exploring}, time-series analysis \cite{ali2019timecluster} and neuroscience \cite{parmar2021visualizing, liu2022improved}. Despite their popularity, theoretical guarantees concerning these methods are highly limited: results have been established only under very stringent assumptions about the data for the tSNE algorithm \cite{linderman2019clustering, pmlr-v75-arora18a}, and no significant results have been established about UMAP to the best of our knowledge. Furthermore, many empirical studies find that, in many cases, UMAP and tSNE do not recover underlying geometry by either (1) failing to find clusters in well-clusterable data \cite{yang2021t}, or (2) introducing fragmentation of underlying connected regions of the data \cite{meilua2024manifold, chari2023specious, moon2019visualizing}.

To address these shortcomings, we incorporate information captured by discrete-graph curvature. Ollivier-Ricci curvature (ORC) \cite{ollivier2007ricci}, a form of curvature for finite metric spaces, has been shown to detect community structure in networks and manifold structure in graphs sampled from manifolds \cite{sia2019ollivier, ni2019community, saidi2025recovering}. We extend these results to describe the interaction of curvature with cluster structure in the face of noise. To this end, we adopt a noisy manifold model of the data, hypothesizing that it consists of several underlying connected components. Under these assumptions, we show two foundational results. First, we establish that for all pairs of points in the same underlying component, a positive-curvature path exists through the induced nearest neighbor graph that closely approximates the geodesic path with high probability. Second, we show that all edges that bridge different connected components have highly negative curvature with high probability. 

Guided by these results, we use ORC to create a distance metric that highlights underlying cluster structure.
We weight the edges of a nearest-neighbor graph by a function of curvature so that shortest (weighted) paths are encouraged to take positive-curvature edges, which are more likely than negative-curvature edges to connect nearby points on the underlying manifold. This produces a pairwise distance metric that
converges in probability at an exponential rate (in the number of samples) to the regime where theoretical results have been established for tSNE
\cite{pmlr-v75-arora18a}. We then use this distance metric in a custom SNE framework, combining elements of both tSNE and UMAP, to create a non-linear dimensionality reduction algorithm that we call \underline{Embed}ding via \underline{O}llivier-\underline{R}icci curvature-based metric learning (EmbedOR). Overall, we find that EmbedOR achieves the delicate balance of highlighting cluster structure in data \textit{without} introducing undesirable fragmentation.\footnote{Our software is available on GitHub: \url{https://github.com/TristanSaidi/embedor}}

\subsection{Contributions}
\begin{enumerate}
    \item We propose EmbedOR, a novel Stochastic Neighbor Embedding (SNE) algorithm that uses discrete graph curvature to obtain a cluster-enhancing distance metric. Critically, we provide theoretical results that show that EmbedOR's distance metric extends results for tSNE \cite{pmlr-v75-arora18a} to a much broader class of datasets.
    \item We provide experimental results that demonstrate that EmbedOR better preserves the underlying connected components and topology of both synthetic and real-world data across a variety of domains. 
    \item We demonstrate that the EmbedOR distance metric can be used as a tool to supplement \textit{existing} data embeddings to identify fragmentation and shed light on the underlying geometry of the data.
\end{enumerate}

\subsection{Related Work}
In a previous paper \cite{saidi2025recovering}, we used ORC to prune ``shortcuts''---edges that bridge points that are close within the ambient Euclidean space but far with respect to geodesic distance---in nearest-neighbor graphs. The algorithm from that paper, ORC-ManL, identifies shortcut edges by thresholding on ORC and graph distance. The pruned nearest-neighbor graph can then be passed downstream to standard manifold learning algorithms; we studied the resultant low-dimensional embeddings in \cite{saidi2025recovering}. While ORC-ManL is highly effective in many circumstances, the use of binary thresholds means that shortcut edges near the thresholds can be missed. The method we construct in the present paper is more robust because we assign edge weights using a \emph{continuous} function of ORC. We go a step further by designing a custom SNE algorithm, rather than simply putting our EmbedOR metric into a standard nonlinear dimension reduction algorithm.

\section{Preliminaries} \label{sec: preliminaries}
\subsection{Differential Geometry} In this section we provide a brief overview of relevant concepts from differential geometry.

\textbf{Manifolds.} A manifold is a generalization of the notion of a surface---it is a topological space that locally looks like Euclidean space. Concretely, a manifold $\man$ is an $d$-dimensional space such that for every point $x \in \man$ there is a neighborhood $U \subseteq \man$ such that $U$ is homeomorphic to $\mathbb{R}^d$. In this paper we will work with submanifolds of Euclidean space that inherit the ambient metric on $\mathbb{R}^D$. Additionally, we will assume that $\man$ is compact. For a more detailed treatment of differential and Riemannian geometry, we direct readers to \cite{prasolov2022differential} and \cite{lee2018introduction}. 

\textbf{Geodesics.} Recall that the length of a continuously differentiable path $\gamma:[a,b] \rightarrow \mathbb{R}^D$ is $L(\gamma) = \int_a^b \|\gamma'(t)\|_2 \, dt$. The \textit{geodesic} distance between two points $x$ and $y$ in a submanifold $\man$ of $\mathbb{R}^D$ is the minimum length over all continuously differentiable paths connecting $x$ and $y$. In this paper, we consider the distance metric $d_{\man}(a,b)$ as the length of the shortest geodesic path through $\man$ connecting $a \in \man$ to $b \in \man$.

\subsection{Nearest Neighbor Graphs}\label{sec: nng}
When approximating underlying manifold structure from samples, it is commonplace to build a nearest neighbor graph. Seminal work in the manifold learning field established that shortest path distances through nearest neighbor graphs are a suitable approximation to manifold geodesic distances \cite{bernstein2000graph}; this idea is central to the Isomap algorithm, a popular nonlinear dimension reduction technique \cite{tenenbaum2000global}. Nearest neighbor graphs use connectivity rules of two flavors: $\epsilon$-radius, or $k$-nearest neighbor ($k$-NN) \cite{bernstein2000graph}. The $\epsilon$-radius connectivity scheme asserts that for any two vertices $a$ and $b$, an edge exists between them if $\|a-b\|_2 \leq \epsilon$. The $k$-NN rule, on the other hand, asserts that the edge exists only if $b$ is one of the $k$ nearest neighbors of $a$, or vice versa. While the $\epsilon$-radius rule is more amenable to theoretical analysis, the $k$-NN rule is used more often in practice. Our theoretical results assume $\epsilon$-radius connectivity because this drastically simplifies the proof technique, but our experiments use $k$-NN connectivity.

\subsection{Ollivier-Ricci Curvature}\label{sec: orc definition}

Ollivier-Ricci curvature (ORC) was proposed as a measure of curvature for discrete spaces \citep{ollivier2007ricci} by leveraging the connection between optimal transport and Ricci curvature of Riemannian manifolds. While there are many subtle variations of ORC, we will describe the one used by \cite{saidi2025recovering}. 
\begin{restatable}[Ollivier-Ricci Curvature]{defn}{orc} \label{def: orc}
Define $\mu_x$ and $\mu_y$ to be the discrete uniform probability measures over the $1$-hop neighborhoods of $x$ and $y$ (with $x$ and $y$ excluded from both sets), respectively. The ORC of the edge $(x,y)$, denoted $\kappa(x,y)$, is defined as
\[\kappa(x,y) \overset{\text{def}}{=} 1 - W(\mu_x, \mu_y)\]
where $W(\mu_x, \mu_y)$ is the 1-Wasserstein distance between the measures $\mu_x$ and $\mu_y$, with respect to the unweighted shortest-path metric $d_G(\cdot \,,\, \cdot)$. 
\end{restatable}
The unweighted shortest-path metric $d_G(x,y)$ is simply the total number of hops in a hop-minimizing path from $x$ to $y$ through $G$. The $1$-Wasserstein distance is computed by solving the following optimal transport problem, 
\[W(\mu_x, \mu_y) = \inf_{\rho \in \Pi(\mu_x, \mu_y)} \sum_{(a,b) \, \in \, V \times V}d_G(a,b)\rho(a,b) \]
where $\Pi(\mu_x, \mu_y)$ is the set of all measures on $V \times V$ with marginals $\mu_x$ and $\mu_y$. Intuitively, ORC quantifies the local structure of $G$; negative curvature implies that the edge is a \say{bottleneck}, while positive curvature indicates that the edge is present in a highly connected community.

\begin{figure}
    \centering
    \includegraphics[width=0.6\linewidth]{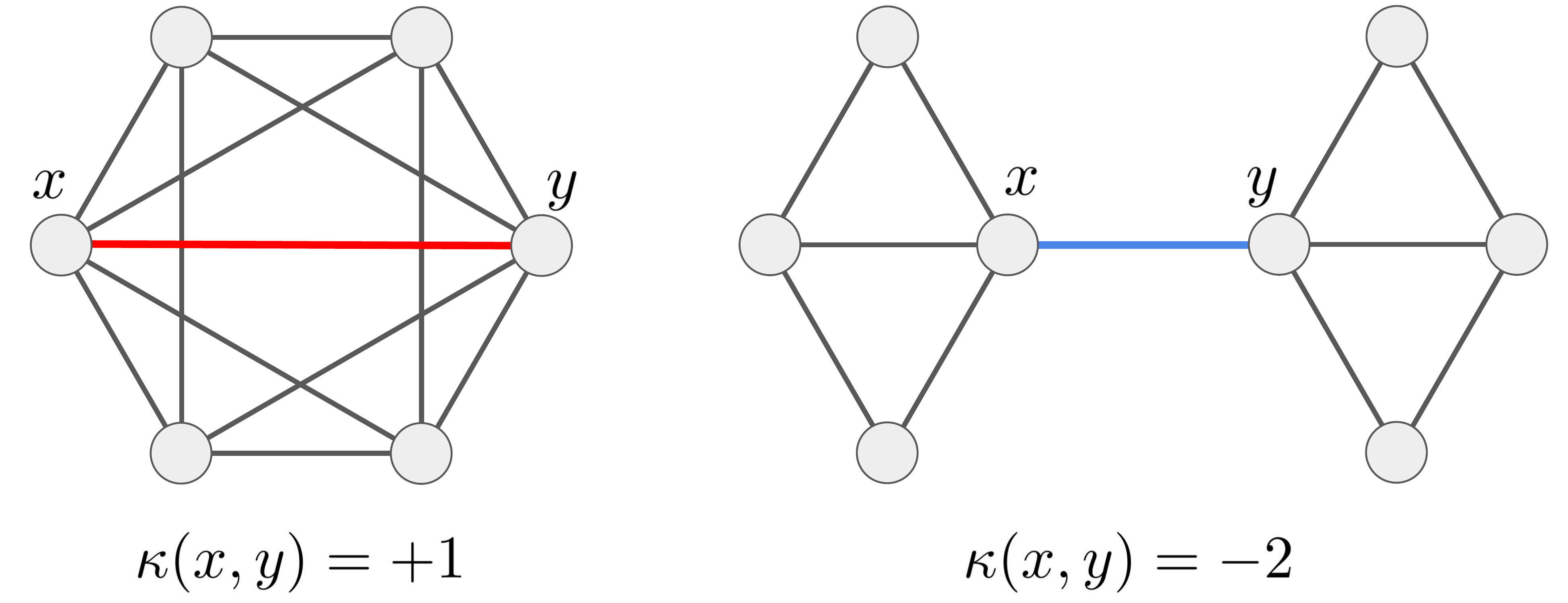}
    \caption{An example of edges with ORC $+1$ (left) and $-2$ (right).}
    \label{fig: orc visualization}
\end{figure}

In our setting, the vertices of the graph $G$ are points in $\mathbb{R}^D$, and edges connect points $(x, y)$ such that $\|x - y\|_2 \leq \epsilon$, where $\epsilon$ is a user-chosen connectivity threshold. We note that other common definitions of ORC take on the form $\kappa(x,y) = 1 - W(\mu_x, \mu_y)/d_G(x,y)$ and use the \textit{weighted} shortest path metric. In this paper, we use unweighted edges in computing the ORC, as it forces the ORC to reflect only the local connectivity, makes it invariant to scale, and restricts the ORC values to lie between $-2$ and $+1$. This notion of unweighted ORC was 
also used
in \cite{saidi2025recovering}. \Cref{fig: orc visualization} depicts examples where the extreme ORC values are attained; in particular, an ORC value of $+1$ indicates that the edge endpoints share neighborhoods, while an ORC value of $-2$ indicates that edge endpoints have completely disjoint \textit{and} unconnected neighborhoods. 
\begin{restatable}[From \cite{saidi2025recovering}]{prop}{uworc}\label{prop: unweighted orc range}
    For any edge $(x,y)$  in an unweighted graph, $-2 \leq \kappa(x,y) \leq 1$.
\end{restatable}

\subsection{Stochastic Neighbor Embeddings} Stochastic Neighbor Embedding (SNE) methods \cite{tsne, umap, hinton2002stochastic} are a family of widely adopted approaches that have been shown to produce impressive visualization of high-dimensional data when cluster structure is known to exist. These methods all share a similar two-step approach for producing and optimizing an embedding of a dataset: (1) for each pair of points, assign a probability $\in [0,1]$ (decreasing in distance) that represents the likelihood that this pair of points are \say{neighbors} in the original data space $\mathbb{R}^D$, then (2) optimize a set of low dimensional points in $\mathbb{R}^m$ in attempt to preserve these pairwise probabilities. In this paper we focus on visualizations via dimension reduction, so we consider the setting where $m = 2$.

Although the SNE framework has been extremely successful, it remains poorly understood. Traditional dimension reduction methods like PCA \cite{mackiewicz1993principal}, MDS \cite{torgerson1952multidimensional}, and Laplacian Eigenmaps \cite{belkin2003laplacian} (among others) cast their embedding optimization problems as convex programs, and in particular, ones that allow for applications of eigendecomposition techniques; this means that the properties of the solutions to the optimization problems can be characterized and understood very well. The SNE framework, on the other hand, uses a highly non-convex objective function. Optimizing such an objective is achieved via gradient descent or its variants, making it particularly challenging to understand and analyze the properties of the local and global minima of the neighborhood-preserving objective.

\begin{figure}[h]
    \centering
    \includegraphics[width=0.6\linewidth]{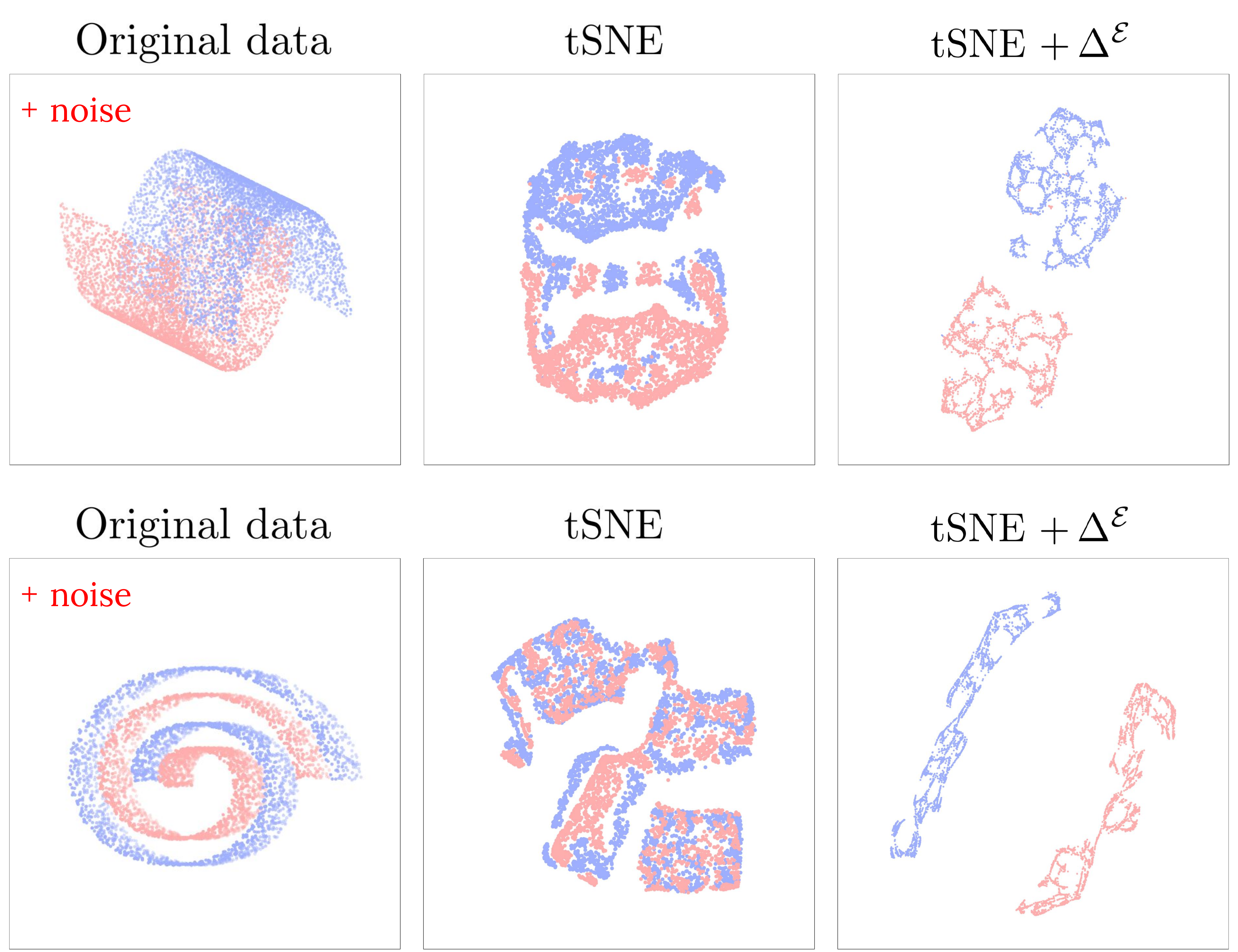}
    \caption{Examples of datasets that violate the well-separated and spherical conditions, as well as embeddings produced by tSNE \cite{tsne} and tSNE using the ORC-enhanced metric $\Delta^{\mathcal{E}}$ described in  \Cref{alg: EmbedOR}.}
    \label{fig: arora condition violation}
\end{figure}

A recent line of work \cite{pmlr-v75-arora18a, linderman2019clustering} attempted to tackle this problem by understanding sufficient conditions under which tSNE produces a correct visualization of clusterable data. The sufficient conditions on the data are as follows. 
\begin{restatable}[Well-separated and spherical data, informal \cite{pmlr-v75-arora18a}]{defn}{gamma-sperical-separable} \label{def: separated spherical data (informal)}
    Let $X \subset \mathbb{R}^D$ be a dataset with a desired clustering $C: X \rightarrow [n_C]$, where $n_C$ is the number of underlying clusters. The dataset $X$ is said to be spherical and well-separated with respect to a metric $d_X$ and the clustering $C$ if all pairwise intra-cluster distances concentrate around some value, and all pairwise inter-cluster distances are bounded from below by something much larger than that value.
\end{restatable}
A formal definition of the conditions is given in \Cref{def: separated spherical data}. From here, \cite{pmlr-v75-arora18a} establishes the following result. Note that we restate it very informally. 
\begin{restatable}[Informal, see Theorem 3.1 of \cite{pmlr-v75-arora18a}]{thm}{aroratsne}\label{thm: arora tsne result}
      Let $X \subset \mathbb{R}^D$ be a dataset with a desired clustering $C: X \rightarrow [n_C]$, where $n_C$ is the number of underlying clusters. If the data satsifies \Cref{def: separated spherical data (informal)}, then a particular instantiation of the tSNE algorithm will produce a cluster-preserving visualization.  
\end{restatable}
While this theoretical result represents an important contribution to the understanding of the SNE framework, the settings for which it is applicable are quite limited. In particular, one can construct many simple datasets that do not satisfy the requirements of the theoretical results---two such examples are shown in \Cref{fig: arora condition violation}. Neither of these datasets is well-separated or spherical. However, as we will show, one can use Ollivier-Ricci curvature to construct a metric that emphasizes this cluster structure, thereby relaxing the assumptions of \Cref{thm: arora tsne result} to more flexible setting such as those shown in \Cref{fig: arora condition violation}.

\section{Method and Theoretical Results} \label{sec: method and theory}

\subsection{Setting and assumptions}

We suppose that our data $X = \{x_1, \dots, x_N\}$ is drawn uniformly from a $d$-dimensional submanifold $\man \subset \mathbb{R}^D$ consisting of $k$ connected components. Note that we will use the words connected component and cluster interchangeably. Equivalently, $\man$ can be represented as the union of $n_C$ nonintersecting submanifolds $\man_1, \dots, \man_{n_C}$ of the same dimension. We will now define the clustering function $C: \man \rightarrow [n_C]$ which maps points in $\man$ to their corresponding cluster label $\in [n_C]$. Furthermore, we will define $C_i = \{p \in \man \, | \, C(p) = i\}$ to be the set of all points in the $i$-th cluster. We also assume that underlying clusters are well separated with respect to the nearest neighbor graph parameter $\epsilon$, stated formally below.

\begin{restatable}[]{asmpt}{assumption} \label{assumption: cluster separation}
    Let $C_i$, $C_j \subset \man$ be any two disjoint connected components of $\man$. Define
    \begin{equation}\label{eq: branch separation}
        c_{ij} = \inf_{(a,b)\in C_i\times C_j} \|a-b\|_2.
    \end{equation}
    We assume that $c_{ij} > \epsilon$ for all $i$,$j$, where $\epsilon$ is the connection threshold used to build the nearest-neighbor graph. We will also define 
    \begin{equation}\label{eq: global branch separation}
        c_{\man} = \inf_{i,j} c_{ij}
    \end{equation}
    to be the global minimum separation between any two connected components of $\man$. 
\end{restatable}

For our theoretical analysis, we adopt a noise model that randomly adds edges to the nearest neighbor graph built from noiseless samples from $\man$. 
We stipulate that for any fixed point $x$, the probability that $x$ is connected to another point $y$ is a monotonic function of the distance between them. This means, for an appropriate choice of model parameters, the marginal probability that a pair of points are connected can be made to match the same quantity under a model that applies perturbations to the datapoint positions. The noise model gives rise to connections between points in different connected components, a property that we see in almost all real-world data. For a detailed treatment of the noise model we use, we refer readers to \Cref{sec: noise model} and \Cref{sec: noise model experiments}. In those sections, we describe in detail the proposed noise model that we employ for theoretical analysis and verify empirically that it coincides with the output of a more common noise model that utilizes ambient perturbations of the data. We also include a discussion of the key limitations of the model in \Cref{sec: noise limitations}.

\subsection{The EmbedOR algorithm}

The EmbedOR algorithm consists of two stages: (1) Compute a cluster-enhancing distance metric using Ollivier-Ricci curvature, and then  (2) apply an augmented SNE algorithm to optimize a low-dimensional embedding to preserve similarities with respect to this metric. In this section, we describe both stages and provide theoretical results that analyze the algorithm's performance under the noise model described in \Cref{sec: noise model}.

\subsubsection{The metric} \label{sec: the metric}

The results of \cite{saidi2025recovering, sia2019ollivier} and \cite{ni2019community} indicate that the edges in graphs that bridge distant neighborhoods of the underlying manifold or disjoint clusters tend to have negative ORC. Thus, we desire a pairwise distance metric that reflects the lengths of paths that are incentivized to \textit{avoid} negative curvature edges. In the event that no such path exists between a pair, we want the metric to assign a large distance between that pair. To achieve this, we assign to each edge an \say{energy} value $\mathcal{E}$ depending on its ORC. Our energy function, defined below, scales the Euclidean distance of an edge in the construction of the weighted nearest-neighbor graph of the data; the metric is then obtained by computing shortest weighted path distances in the graph.
Before describing the exact form of this \say{energy} function $\mathcal{E}$, we list some desiderata. In particular, we want the energy to be monotonically \textit{decreasing} in the edge curvature $\kappa$; this will encourage pairwise shortest-path distances to traverse more positively-curved edges. We also want the bounding conditions $\mathcal{E} = 1$ when $\kappa = 1$, and $\mathcal{E} \rightarrow \infty$ as $\kappa \rightarrow -2$. Finally, we constrain the family of functions further by stipulating that $\mathcal{E} = 2$ when $\kappa = 0$. With these desiderata in mind, we construct an energy function parameterized by a hyperparameter $p$ that controls the degree of repulsion between endpoints of negatively curved edges.
\begin{restatable}[Edge energy]{defn}{energy} \label{def: energy function}
We define the energy of an edge with curvature $\kappa$ to be
\[\mathcal{E}(\kappa;p) \overset{\text{def}}{=} \biggl(-\frac{1}{\log(3/2)}\Bigl(\log(\kappa+2) - \log(2)\Bigr)+1\biggr)^{p} + 1\]
where $p$ is a user parameter controlling the degree of repulsion.
\end{restatable}

\begin{figure}
    \centering
    \includegraphics[width=0.6\linewidth]{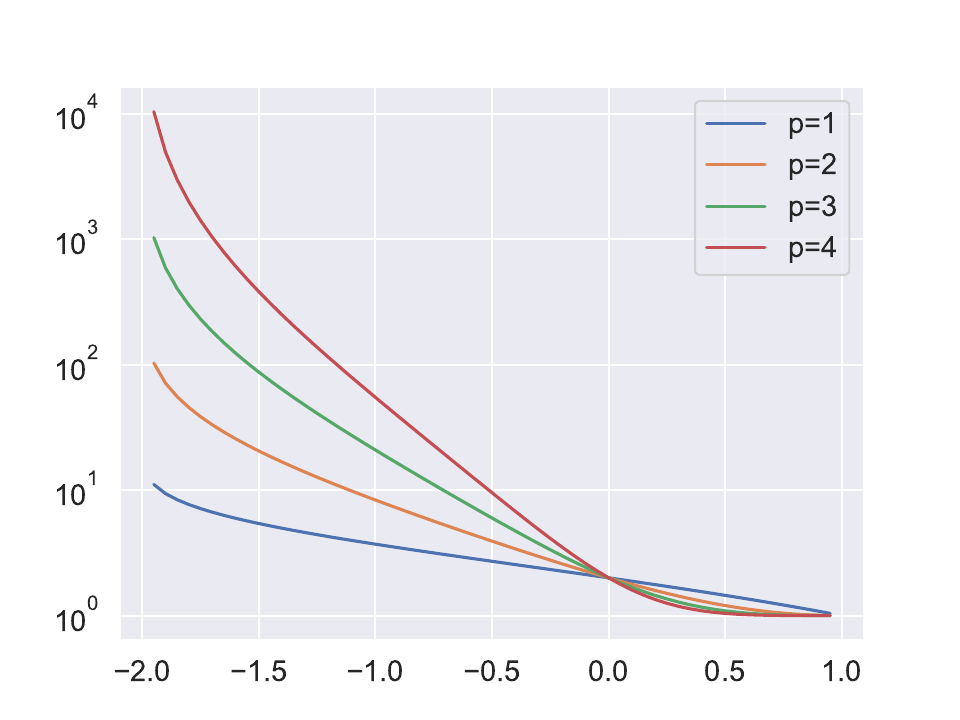}
    \caption{Plot of $\mathcal{E}(\kappa;p)$ (log scale) versus $\kappa$ for several values of $p$.}
    \label{fig: energy plot}
\end{figure}

\Cref{fig: energy plot} provides a plot of $\mathcal{E}$ for several values of $p$. Observe that for any $p > 0$, we have $\mathcal{E}(1;p) = 1$ and $\lim_{\kappa \to -2} \mathcal{E}(\kappa;p) = \infty$ as desired, and also note that $p$ controls the rate at which the energy asymptotes to $\infty$ as the curvature $\kappa$ approaches $-2$. \Cref{alg: EmbedOR metric} describes how the EmbedOR metric is constructed. Put simply, it amounts to the shortest path distances through the nearest-neighbor graph with respect to a weighting function,
\begin{equation}\label{eq: weight}
    w(x_i, x_j) = \frac{1}{7}\|x_i-x_j\|_2\cdot\mathcal{E}(\kappa(x_i, x_j);p)
\end{equation}
which consists of the product of the Euclidean distance of the edge and its energy. Notice that the selection of $p=0$ recovers a scaled version of the Isomap metric. As one increases the $p$ parameter, the metric emphasizes cluster structure by avoiding negative-curvature edges. 

\begin{algorithm}[H]
        \begin{algorithmic}[1]
        \caption{Metric $\Delta^{\mathcal{E}}$}
        \label{alg: EmbedOR metric}
        \Require $X = \{x_1,\dots,x_N\} \subset \mathbb{R}^{D}$, $\epsilon$ or $k$ (graph parameter), $p$ (repulsion factor)
        \State $G \gets \text{NearestNeighborGraph}(X, \epsilon \text{ or } k)$
        \For{$(x_i,x_j)$ in $E$}
            \State $\kappa(x_i,x_j) \gets \text{OllivierRicci}(G,(x_i,x_j))$
            \State $w(x_i,x_j) \gets \frac{1}{7}\|x_i-x_j\|_2 \cdot \mathcal{E}(\kappa(x_i,x_j); p)$
        \EndFor
        \State  $\Delta^{\mathcal{E}} \gets \text{AllPairsShortestPaths}(G, w)$ 
        \State \Return $\Delta^{\mathcal{E}}$
    \end{algorithmic}
\end{algorithm}

Next we provide theoretical results that establish that \Cref{alg: EmbedOR metric} gives rise to more clusterable metric structure when the underlying data comes from a submanifold of $\mathbb{R}^D$ with several connected components. Our first result establishes that the pairwise distance under $\Delta^{\mathcal{E}}$ between \textit{all} points in the \textit{same} connected component falls somewhere between the manifold geodesic distance and a fraction of the ambient Euclidean distance with high probability. 

\begin{restatable}[Intra-cluster distances]{thm}{similarities}
\label{thm: similar pairs}
    Suppose $X$ consists of $N$ i.i.d. samples from a uniform distribution over $\man$, and let $\Delta^{\mathcal{E}}$ be the metric created using \Cref{alg: EmbedOR metric} from $X$'s nearest neighbor graph corrupted with noisy connections as described in \Cref{eq: noise model}. Then we have for all $a,b$ such that $C(a) = C(b)$, \[\|a-b\|_2/7 \leq  \Delta^{\mathcal{E}}(a,b) \leq d_{\man}(a,b)\]
    with probability at least $ 1 - \mathcal{O}(N^{5/2+\frac{1}{2d}})e^{-\Omega(N^{1/2})}$.
\end{restatable}
\noindent The proof of \Cref{thm: similar pairs} is provided in \Cref{sec: proof of theorems}. Our second result is complementary, in the sense that it lower bounds the pairwise distance between all points in \textit{different} connected components.

\begin{restatable}[Inter-cluster distances]{thm}{dissimilarities}\label{thm: dissimilar pairs}
      Suppose $X$ consists of $N$ i.i.d. samples from a uniform distribution over $\man$, and let $\Delta^{\mathcal{E}}$ be the metric created using \Cref{alg: EmbedOR} from $X$'s nearest neighbor graph corrupted with noisy connections as described in \Cref{eq: noise model} from \Cref{sec: noise model}.  If $s_{\max} < \frac{3}{8e}$ from \Cref{eq: noise model}, then we have for all $a,b$ such that $C(a) \neq C(b)$,
    \[\Delta^{\mathcal{E}}(a,b) > \frac{c_{\man}}{7}\mathcal{E}(-1/2; p) \]
     with probability at least $1-\mathcal{O}(N^4)e^{-\Omega(N)}$, where $p$ is the hyperparameter in \Cref{def: energy function}, and $c_{\man}$, defined in \Cref{eq: global branch separation}, is the minimum separation between any two connected components of $\man$. 
\end{restatable}
\noindent Again, the proof of \Cref{thm: dissimilar pairs} is provided in \Cref{sec: proof of theorems}. Observe that the lower bound on distance between inter-cluster points is monotonically \textit{increasing} in the user-selected parameter $p$, whereas the upper bound on distances between intra-cluster points is \textit{independent} of $p$. Thus, the $p$ parameter allows the user to manually select the degree of cluster emphasis, manifested through the distance metric $\Delta^{\mathcal{E}}$.

For a sufficiently large choice of parameter $p$, the EmbedOR metric satisfies the requirements of \cite{pmlr-v75-arora18a}---described informally in \Cref{def: separated spherical data (informal)}---that guarantee a cluster-preserving visualization by the tSNE algorithm \cite{tsne}. Critically, this means that the EmbedOR metric relaxes the requirements for the theoretical results of \cite{pmlr-v75-arora18a}, allowing them to apply to a broader range of datasets. We refer back to \Cref{fig: arora condition violation}, which illustrates examples of datasets that violate these conditions, resulting in undesirable embeddings from tSNE. We also see that the substitution of the Euclidean metric with the EmbedOR metric $\Delta^{\mathcal{E}}$ (\Cref{alg: EmbedOR metric}, described in \Cref{sec: the metric}) enables tSNE to produce a cluster-preserving visualization of the data. 

Our third major result formalizes the notion that tSNE operating on the EmbedOR metric $\Delta^{\mathcal{E}}$ returns a cluster-preserving visualization for any data adhering to the model described in \Cref{sec: noise model}. We will define what it means to have a cluster-preserving visualization, then we will state our final result, \Cref{thm: tsne convergence (informal)}.

\begin{restatable}[Full visualization \cite{pmlr-v75-arora18a}]{defn}{full-visualization} \label{def: full visualization}
Let $Y = \{y_i\}_{i=1}^N$ be a $2$-dimensional embedding of a dataset $X$ with ground truth clustering $C_1^{X}, \dots, C_{n_C}^X$ where $C_i^X = X \cap C_i$. The embedding $Y$ is said to be a \textit{full visualization} if there exists a partition $\mathcal{P}_1, \dots, \mathcal{P}_{n_C}$ of $\{y_i\}_{i=1}^N$ such that,
\begin{enumerate}
    \item $\mathcal{P}_i = C_i^X$
    \item for every $y_j, y_j' \in \mathcal{P}_i$ and for every $y_l \in Y \setminus \mathcal{P}_i$ we have $\|y_j - y_{j'}\|_2 \leq \frac{1}{2} \|y_j -y_l\|_2$.
\end{enumerate}
\end{restatable}

\begin{restatable}[Convergence of tSNE with $\Delta^{\mathcal{E}}$]{thm}{tsne}\label{thm: tsne convergence (informal)}
    Suppose $X$ consists of $N$ i.i.d. samples from a uniform distribution over $\man$, and let $\Delta^{\mathcal{E}}$ be the metric created using \Cref{alg: EmbedOR metric} from $X$'s nearest neighbor graph corrupted with noisy connections as described in \Cref{eq: noise model}. Then with high probability over the choice of initialization and the configurations of $X$ and for a sufficiently large choice of $p$, tSNE after $\mathcal{O}(\frac{\sqrt{N}}{n_C^2})$ iterations using the metric $\Delta^{\mathcal{E}}$ returns a full visualization of $X$. 
\end{restatable}
\noindent 
This result is a corollary of the previous theorems. \Cref{thm: similar pairs} and \Cref{thm: dissimilar pairs} establish that we can choose a sufficiently large $p$ such that \Cref{def: separated spherical data (informal)} holds with high probability.
From there, the convergence of tSNE follows by Theorem 3.1 of \cite{pmlr-v75-arora18a}. 

\begin{restatable}[Choice of $p$]{rmk}{p parameter (main text)}
    In \Cref{eq: choice of algorithm parameter p}, we derive a lower bound on the required size of $p$. First, we note that the expression for $p$ is increasing in $N$, which is a potentially undesirable property. However, this property arises from the results of \cite{pmlr-v75-arora18a}, where they require increasingly well-behaved data as $N$ increases. Second, we note that this bound on $p$ is dependent on $c_{\man}$, which we would not expect to be able to estimate in practice. With that being said, this bound establishes that for a \textit{fixed} point cloud generated from some underlying manifold $\man$, there exists a sufficiently large $p$ such that the requirements on the metric hold. This would \textit{not} be true for a more naive choice of metric, like the ambient Euclidean metric, for example. 
\end{restatable}

\Cref{thm: tsne convergence (informal)} motivates our use of the metric $\Delta^{\mathcal{E}}$ in our Stochastic Neighbor Embedding algorithm, which we define in the next subsection. However, we clarify here that our algorithm is a custom SNE that blends elements of tSNE and UMAP, and thus differs from what is suggested by \Cref{thm: tsne convergence (informal)}. We recognize that this raises a natural question about the applicability of Theorem 3.3 to our method. To address this we will remark on two things. Firstly, prior work has demonstrated strong empirical similarities between UMAP and tSNE \cite{böhm2022attractionrepulsionspectrumneighborembeddings}, suggesting that incorporating UMAP-style components into tSNE does not fundamentally alter the algorithmic behavior. This lends credence to the idea that insights from the tSNE analysis remain informative. Secondly, the specific tSNE instantiation analyzed in \cite{pmlr-v75-arora18a} is not directly aligned with widely used implementations such as scikit-learn’s \cite{scikit-learn} and FIt-SNE’s \cite{linderman2019fast}, which adopt choices optimized for speed and quality and diverge from the theoretical regime. In our view, such theoretical analyses (including ours) are best seen as descriptive and motivating guides. Thus, while Theorem 3.3 does not apply directly to the exact algorithm we propose, it provides theoretical motivation for why the EmbedOR metric $\Delta^{\mathcal{E}}$ should be effective in a SNE algorithm. For this reason, rather than adhering strictly to the theoretical setting, we designed EmbedOR to use the metric $\Delta^{\mathcal{E}}$ and retain desirable scaling properties - we include further discussion regarding this decision to blend elements of UMAP and tSNE in \Cref{sec: rmk on design choices}, after we describe the EmbedOR algorithm.

\subsubsection{Our Stochastic Neighbor Embedding} \label{sec: sne construction}

To construct a SNE algorithm using the metric $\Delta^{\mathcal{E}}$, we pull elements from both tSNE \cite{tsne} and UMAP \cite{umap}. The tSNE algorithm considers all pairwise interactions, but uses the ambient Euclidean metric to do so. The UMAP algorithm, on the other hand, only optimizes for preserving the similarities between a point and its $k$ nearest neighbors. Furthermore,
there is an \emph{equal} repulsive force between every unconnected pair of points
in the original nearest neighbor graph, a choice that does not reflect the geometry of the data. 

What UMAP lacks conceptually, it compensates for computationally. Namely, the effective loss function that UMAP optimizes (which is different from the purported loss, as studied by \cite{damrich2021umap}) allows for easy application of stochastic gradient descent \cite{robbins1951stochastic}. Furthermore, it avoids the need for normalizing the low dimensional affinities as tSNE does. In constructing our SNE, we will draw on the strengths of both UMAP and tSNE. 

Our high-dimensional affinities will capture all pairwise interactions as tSNE does, but we will utilize our \textit{intrinsic} metric $\Delta^{\mathcal{E}}$. We define
\begin{equation} \label{eq: sym affinities}
    p_{ij}\overset{\text{def}}{=} \frac{1}{2}\exp\Biggl\{ -\biggl(\frac{\Delta^{\mathcal{E}}(x_i, x_j)}{\sigma_i}\biggr)^2\Biggr\} + \frac{1}{2}\exp\Biggl\{ -\biggl(\frac{\Delta^{\mathcal{E}}(x_i, x_j)}{\sigma_j}\biggr)^2\Biggr\}
\end{equation}
where $\sigma_i$ and $\sigma_j$ are chosen to match a desired perplexity \cite{tsne}. Since we have $p_{ij} \in [0,1]$ we can interpret it as the probability $x_i$ is a neighbor of $x_j$ (with respect to $\man$). Let $y_i$ denote the low-dimensional embedding of point $x_i$, which we will calculate by solving an optimization problem. The low-dimensional affinities are chosen to follow a Student t-distribution with one degree of freedom
\begin{equation}\label{eq:fij}
    f_{ij}\overset{\text{def}}{=} \frac{1}{1+\|y_i-y_j\|^2}\,,
\end{equation}
as is done in both UMAP and tSNE. Since $f_{ij}$ also lies in $[0,1]$ we can interpret it as the probability that $y_i$ and $y_j$ are neighbors in our embedding. As UMAP does, we can interpret $\{p_{ij}\}_{ij}$ and $\{f_{ij}\}_{ij}$ as the membership functions for fuzzy sets in the space of all possible edges, $\mathcal{Z} = X \times X$. A fuzzy set $A$ in $\mathcal{Z}$ is characterized by a membership function $f_A: \mathcal{Z} \rightarrow [0,1]$ with the value representing the \say{grade} of the membership of the element in $A$ \cite{ZADEH1965338}. Now we can optimize $Y$ to minimize an augmented version of the \say{fuzzy cross entropy} between the fuzzy sets $f_{ij}$ and $p_{ij}$ \cite{li2015fuzzy},

\begin{equation}\label{eq: fuzzy cross entropy objective}
    \mathcal{L}(Y) = \sum_{j < i}\biggl[p_{ij}\log\biggl(\frac{p_{ij}}{f_{ij}}\biggr) + \lambda\cdot(1-p_{ij})\log\biggl(\frac{1-p_{ij}}{1-f_{ij}}\biggr) \biggr]
\end{equation}

\noindent where $\lambda$ controls the relative weight of the two terms.  The key difference here from UMAP is that for pairs $(x_i,x_j) \notin E$, we don't have $p_{ij} = 0$. That means a point $x_j$ just outside of the $k$-th nearest neighbor of point $x_i$ will experience repulsive forces determined by $\Delta^{\mathcal{E}}(x_i, x_j)$. In UMAP, this point would be repelled with the same force as any other non-adjacent point.

Now we can rewrite our optimization problem to facilitate the application of stochastic gradient descent (SGD). In particular, if we define discrete distributions over attractive and repulsive interactions interactions respectively 
\[p^+_{ij} \overset{\text{def}}{=}\frac{p_{ij}}{\sum_{j' < i'}p_{i'j'}} \qquad p^-_{ij} \overset{\text{def}}{=}\frac{1-p_{ij}}{\sum_{j' < i'}(1-p_{i'j'})}\] 
then we can say,
\begin{equation}\label{eq: objective}
    Y^* =  \arg \max_{Y} \,  \underbrace{\mathbb{E}_{(i,j) \sim p^+}\Bigl[\log(f_{ij})\Bigr]}_{\mathclap{\text{attractive term}}} + \Bigg(\frac{\frac{N^2-N}{2}-Z}{Z}\Bigg)\Bigg(\frac{1}{N^2}\Biggr) \underbrace{\mathbb{E}_{(i,j) \sim p^-}\Bigl[\log(1-f_{ij})\Bigr]}_{\mathclap{\text{repulsive term}}}
\end{equation}
where $Z \overset{\text{def}}{=} \sum_{i<j}p_{ij}$. Note that this objective uses $\lambda = 1/N^2$, which is the parameter setting we opt for in our experiments. The derivation of the simplified objective shown above is provided in \Cref{sec: objective function derivation}. This derivation gives rise to \Cref{alg: EmbedOR}, the full EmbedOR algorithm. 

\subsubsection{A note on our design choices} \label{sec: rmk on design choices}

We chose to design a custom SNE, rather than directly using the metric $\Delta^{\mathcal{E}}$ with tSNE, which is the setting studied in \Cref{thm: tsne convergence (informal)}. The results of \Cref{thm: tsne convergence (informal)} (which are derived from Theorem 3.1 in \cite{pmlr-v75-arora18a}) establish guarantees for a \textit{very specific} instantiation of tSNE, one that deviates from widely adopted and optimized versions of tSNE such as the default implementation from scikit-learn \cite{scikit-learn} and FIt-SNE \cite{linderman2019fast}. Furthermore, the default settings of these popular implementations largely reflect parameter choices that consistently yield a strong tradeoff between speed and quality. Thus, instead of constraining ourselves to the exact setting of the theorem, we chose to construct a custom SNE algorithm that operates on the EmbedOR metric $\Delta^{\mathcal{E}}$ in an effort to optimize for empirical performance and speed.

The SNE stage of the EmbedOR algorithm blends elements of both UMAP and tSNE; here, we briefly provide justification for our choices. Firstly, we chose to optimize the UMAP objective as it avoids the need to re-normalize the low-dimensional affinities after each update, which speeds up the implementation and saves memory. On the other hand, we chose to compute similarities using the tSNE perplexity procedure, as the UMAP procedure for doing so is slightly more complicated and heuristic. Furthermore, given that rigorous theoretical results have been established for the tSNE framework but not the UMAP framework, we wanted to adhere closely to the tSNE procedure when possible.

\subsubsection{Computational and Memory Complexity} We provide an analysis and discussion of the memory and computational complexity of the EmbedOR algorithm in \Cref{sec: comp and memory complexity}. In this section we show that the computational complexity of the exact algorithm scales as $\mathcal{O}\big(N^2(k+\log N) + Nk^4 + T\big)$ when one uses a $k$-NN graph. That being said, we also show that one can employ approximations that minimally affect embedding quality that bring the runtime to sub-quadratic in $N$.

\begin{algorithm}[h]
        \begin{algorithmic}[1]
        \caption{EmbedOR}
        \label{alg: EmbedOR}
        \Require $X = \{x_1,\dots,x_N\} \subset \mathbb{R}^{D}$, $\epsilon$ (graph parameter), $\tau$ (perplexity), $p$ (repulsion factor), $d$ (embedding dimension), $T$ (iterations), $\eta$ (learning rate)
        \State  $\Delta^{\mathcal{E}} \gets \text{Metric}(X, \epsilon, p)$ \Comment{\Cref{alg: EmbedOR metric}}
        \State $\{\sigma_i\}_{i=1,\dots,N} \gets \text{MatchPerplexity}(\Delta^{\mathcal{E}}, \tau)$ \Comment{Subroutine from \cite{tsne}}
        \State $\{p_{ij}\}_{ij} \gets \text{ComputeSymAffinities}(\Delta^{\mathcal{E}}, \{\sigma_i\}_{i=1,\dots,N})$ \Comment{\Cref{eq: sym affinities}}
        \State $Y \gets \text{Laplacian Eigenmaps}(\{p_{ij}\}_{ij}, d)$ \Comment{Initialize embedding}
        \For{$t=1$ to $T$}
            \State Sample $(i,j) \sim p^+$ \Comment{Positive pair}
            \State $y_i \gets y_i + \eta\nabla_{y_i}\log(f_{ij})$
            \State $y_j \gets y_j + \eta\nabla_{y_j}\log(f_{ij})$
            \State Sample $(i',j') \sim p^-$ \Comment{Negative pair}
            \State $y_{i'} \gets y_{i'} + \eta\Big(\frac{\frac{N^2-N}{2}-Z}{Z}\Big)\Big(\frac{1}{N^2}\Bigr)\nabla_{y_{i'}}\log(1-f_{i'j'})$
            \State $y_{j'} \gets y_{j'} + \eta\Big(\frac{\frac{N^2-N}{2}-Z}{Z}\Big)\Big(\frac{1}{N^2}\Bigr)\nabla_{y_{j'}}\log(1-f_{i'j'})$
            
        \EndFor
        \State \Return $Y$
    \end{algorithmic}
\end{algorithm}

\section{Experiments} \label{sec: all experiments}

\begin{figure*}[h]
    \centering
    \includegraphics[width=0.95\linewidth]{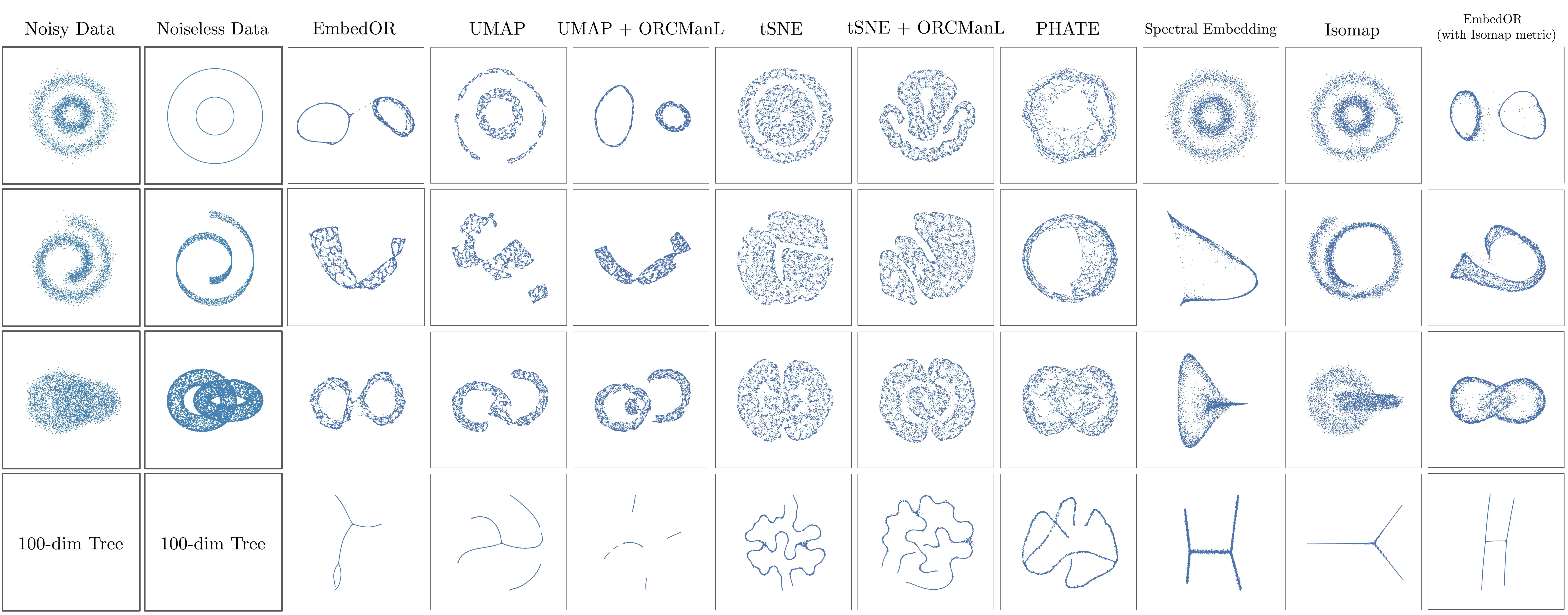}
    \caption{Embeddings produced by various non-linear dimension reduction techniques on synthetic datasets using default parameters.}
    \label{fig: synthetic data}
\end{figure*}

In this section we present a range of experiments on synthetic and real data to assess the performance of the EmbedOR algorithm. Through our experiments, we seek to justify the following two claims.

\begin{claim}
    EmbedOR produces visualizations of noisy data that separate underlying connected components \textit{and} avoid fragmentation of said components. \label{goal: visualization}
\end{claim}

This claim is supported theoretically by \Cref{thm: tsne convergence (informal)}, which proves that tSNE used with our EmbedOR metric $\Delta^{\mathcal{E}}$ produces such cluster-preserving visualizations. To provide empirical evidence for \Cref{goal: visualization}, we evaluate EmbedOR visualizations against a set of non-linear dimensionality reduction techniques and their variants that have been adopted in the literature:
\begin{enumerate}
    \item The most widely adopted SNE algorithms,  UMAP \cite{umap} and tSNE \cite{tsne}. 
    \item UMAP and tSNE in conjunction with the ORCManL algorithm \cite{saidi2025recovering}, a method that prunes nearest neighbor graphs using ORC.
    \item  PHATE \cite{moon2019visualizing}, a method for visualizing trajectory data that relies on principles from diffusion geometry.
    \item Seminal dimension reduction techniques Isomap \cite{tenenbaum2000global} and Laplacian Eigenmaps \cite{belkin2003laplacian}. We note that the visualizations produced by these two methods are deferred to \Cref{fig: (f)mnist extra}, \Cref{fig: iPSCs extra}, \Cref{fig: retinal extra} and \Cref{fig: chimp extra}.
    \item EmbedOR where the metric (line 1 of \Cref{alg: EmbedOR}) is replaced with the Euclidean-weighted graph geodesic distance metric used in Isomap. Observe that this is equivalent to setting $p=0$ in \Cref{alg: EmbedOR metric} (up to a scaling factor). 
\end{enumerate}

Our second claim is that one can use the EmbedOR metric $\Delta^{\mathcal{E}}$ to identify fragmentation in any visualization---even ones produced by other algorithms.
\begin{claim}
    Edges with short $\Delta^{\mathcal{E}}$ distance are more likely to connect points that are close together within the underlying manifold $\man$. Therefore, the EmbedOR metric $\Delta^{\mathcal{E}}$ can be reliably used to supplement visualizations to identify fragmentation. \label{goal: supplementation}
\end{claim}
\noindent By visualizing the shortest edges with respect to $\Delta^{\mathcal{E}}$, we can see if any of these edges have been stretched too far in the embedding. \Cref{goal: supplementation} says that this indicates likely fragmentation, because the endpoints of the edges should be close together in the original manifold space $\man$.

\Cref{goal: supplementation} is supported theoretically by \Cref{thm: similar pairs} and \Cref{thm: dissimilar pairs}, which together establish the fact that edges that are close with respect to $\Delta^{\mathcal{E}}$ connect points that are close with respect to $\man$. In the following subsections, we provide empirical evidence for \Cref{goal: supplementation} by showing that short $\Delta^{\mathcal{E}}$ edges are more likely to connect points in the same ground-truth cluster.

We note that for all experiments in this section, we construct a $k$-NN graph ($k = 15$) instead of an $\epsilon$-radius graph (line 1 of \Cref{alg: EmbedOR metric}). Furthermore, we use $p = 3$ and $\tau = 150$ for all experiments in this section. Parameter ablations for $p$ and $\tau$ are provided in \Cref{sec: parameter ablations}.

\subsection{Synthetic data}

\begin{table*}[h]
\small
\centering
 \begin{tabular}{l | >{\centering\arraybackslash}p{0.10\textwidth}  >{\centering\arraybackslash}p{0.10\textwidth}  >{\centering\arraybackslash}p{0.10\textwidth} >{\centering\arraybackslash}p{0.10\textwidth} | >{\centering\arraybackslash}p{0.10\textwidth} } 
  \toprule  & Circles &  Swiss Roll & Tori & Tree & Avg. Placement \\ 
\midrule 
 \hline
 EmbedOR (ours)  
 & $\underline{0.78} \pm 0.15$ & $\underline{0.92} \pm 0.04$ & $\mathbf{0.83} \pm 0.02 $ & $0.81 \pm 0.12$ & $\mathbf{2.25}$ \\ 
 UMAP* \cite{umap} 
 & $0.48 \pm0.03$ & $0.84 \pm 0.13$ & $0.68 \pm 0.03$ & $\underline{0.92} \pm 0.01$ & $3.5$ \\ UMAP + ORCManL \cite{umap, saidi2025recovering} & $\mathbf{0.91} \pm 0.01$ & $\mathbf{0.98} \pm 0.01$ & $\underline{0.79} \pm 0.10$ & $0.70 \pm 0.12$ & $\underline{2.75}$\\
 tSNE* \cite{tsne} 
 & $0.47 \pm 0.01$ & $0.78 \pm 0.01$ & $0.62 \pm 0.08$ & $0.84 \pm 0.01$ & $4.75$\\ 
 tSNE + ORCManL \cite{tsne, saidi2025recovering} & $0.60 \pm 0.16$ & $0.86 \pm 0.05$ & $0.62 \pm 0.09$ & $0.63 \pm 0.01$ & $4.50$ \\ 
 PHATE \cite{moon2019visualizing} 
 & $0.25 \pm 0.09$ & $0.27 \pm 0.02$ & $0.43\pm 0.01$ & $0.49 \pm 0.06$ & $8.75$ \\ Laplacian \cite{belkin2003laplacian} 
 & $0.27 \pm 0.00$ & $0.22 \pm 0.01$ & $0.48 \pm 0.01$ & $0.79 \pm 0.00$ & $7.50$\\ Isomap \cite{tenenbaum2000global} 
 & $0.50 \pm 0.01$ & $0.31 \pm 0.02$ & $0.40 \pm 0.01$ & $\mathbf{0.93} \pm 0.01$ & $5.00$ \\ EmbedOR (w/ Isomap metric) 
 & $0.47 \pm 0.20$ & $0.72 \pm 0.07$ & $0.59 \pm 0.02$ & $0.80 \pm 0.12$ & $6.00$\\
 \bottomrule
 \end{tabular}
 \caption{Mean and standard deviation of the Spearman correlation coefficient between embedding distances and estimated geodesic distance for the synthetic datasets over $10$ random trials. Note that the embeddings were computed on noisy data, while the ground-truth geodesic distances were estimated on the noiseless data. For datasets with more than 1 connected component, ground-truth distances between points in different connected components were set to a large constant value. Bold text indicates the best performance, while underlined text indicates the second best. Note that, due to parameter sensitivity, the performance associated with the starred methods were chosen from an extensive parameter search (the results of which are provided in Tables \ref{table: geodesic distances umap ablation} and \ref{table: geodesic distances tsne ablation}). The parameters for all other methods were chosen to be the default algorithm parameters.  }
 \label{table: geodesic distances}
\end{table*}

\Cref{fig: synthetic data} displays embeddings produced by EmbedOR and baseline algorithms for a variety of synthetic datasets with non-trivial geometry and topology. We find that EmbedOR consistently demonstrates superior or tied-for-best performance in terms of (1) separation of the underlying connected components, and (2) lack of fragmentation of continuous paths. We find that UMAP introduces fragmentation on all datasets, while tSNE fails to unroll the Swiss Roll and fails to highlight cluster structure. UMAP + ORCManL exhibits the best performance among baselines, comparable to that of EmbedOR (though fragmentation still appears with the tree dataset), but tSNE + ORCManL does not seem to improve over tSNE alone. We attribute this to the fact that, in the \textit{unbounded} noise regime tested in this experiment (which differs from the \textit{bounded} noise tested in \cite{saidi2025recovering}), the ORCManL algorithm fails to prune many shortcutting edges. PHATE and Laplacian Eigenmaps also struggle with highlighting cluster structure, though they avoid the fragmentation issues present in the UMAP embeddings. Similarly, Isomap and EmbedOR with the Isomap metric fail to highlight cluster structure in the datasets, but avoid fragmentation issues. Overall, we find that EmbedOR achieves the delicate balance of highlighting cluster structure \textit{without} introducing fragmentation, qualities not present among any of the baselines tested. 

To make the results quantitative, \Cref{table: geodesic distances} provides Spearman correlation coefficients between pairwise embedding distances and estimated geodesic distances (obtained from the noiseless datasets). Here we also see that EmbedOR has the best placement among baseline methods on average (right column of \Cref{table: geodesic distances}). We also note that, due to parameter sensitivity, the results for UMAP and tSNE in \Cref{table: geodesic distances} were obtained by varying algorithm hyperparameters over a large range and cherry-picking the best performance \textit{per dataset}---this procedure was not done for EmbedOR. We provide full UMAP and tSNE results across a wide range of parameter settings in \Cref{table: geodesic distances umap ablation} and \Cref{table: geodesic distances tsne ablation}.


To provide 
empirical
support of 
\Cref{goal: supplementation},
we present the results of experiments on synthetic data in \Cref{table: low-energy graph}. In these experiments, we find that the $33\%$ shortest edges under the metric $\Delta^{\mathcal{E}}$ have a more than $10$-fold drop in the number of connected-component-bridging edges, showing that short $\Delta^{\mathcal{E}}$ edges, as compared to other edges in the nearest-neighbor graph, are much more likely to connect points in the same connected component. This suggests that the EmbedOR metric indeed captures information about the connected components of the underlying data manifold $\man$. In the subsequent section we repeat this experiment on labeled single-cell RNA sequencing (scRNAseq) to support the notion that \Cref{goal: supplementation} extends beyond synthetic data.

\begin{table}[H]
\small
\centering
 \begin{tabular}{l | >{\centering\arraybackslash}p{0.15\textwidth}  >{\centering\arraybackslash}p{0.15\textwidth}  >{\centering\arraybackslash}p{0.15\textwidth} } 
  \toprule  & Circles &  Tori & Moons \\ 
\midrule 
 \hline
 Low $\Delta^{\mathcal{E}}$ graph & $1.9 \pm1.6$ & $54.8 \pm 23.1$ & $10.2 \pm 6.37$\\ 
 Full graph & $139.7 \pm 33.2$ & $623.0 \pm 70.6$ & $166.8 \pm 38.4$\\ 
 \bottomrule
 \end{tabular}
 \caption{Number of connected-component-bridging edges in a full $k$-NN graph (bottom row), versus the graph consisting of the shortest $33\%$ of edges w.r.t. the metric $\Delta^{\mathcal{E}}$ (top row). Each entry reports the mean and standard deviation across 10 independent trials.}
 \label{table: low-energy graph}
\end{table}

\subsection{Real data} \label{sec: real data}

\begin{figure}[h]
    \centering
    \includegraphics[width=\linewidth]{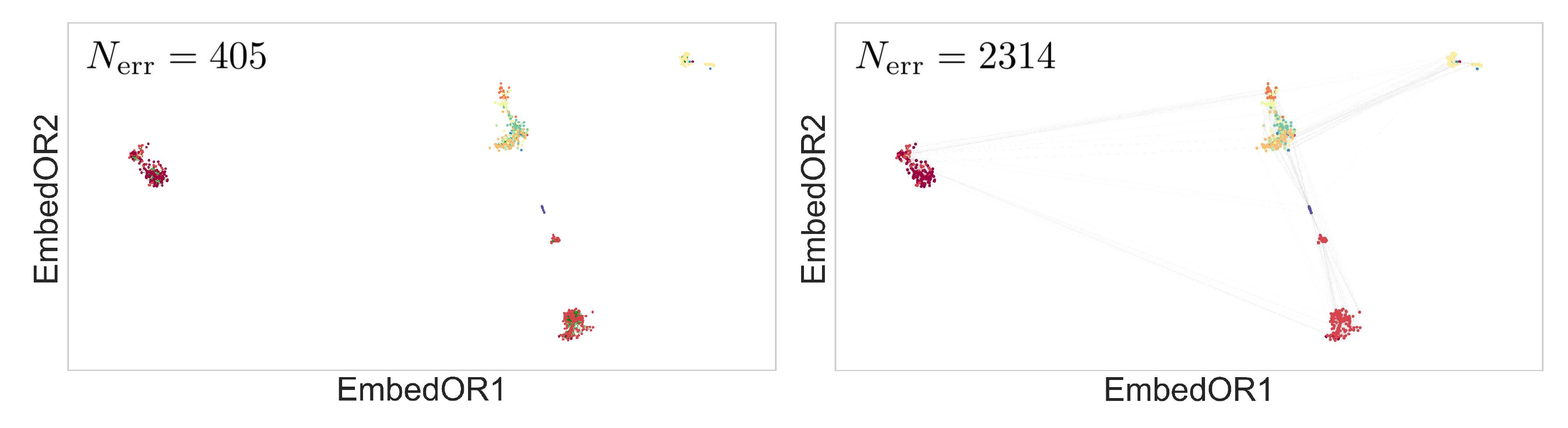}
    \caption{EmbedOR visualization of peripheral blood mononuclear cell data with the full $k$-NN graph (right) and the $33\%$ shortest edges under the metric $\Delta^{\mathcal{E}}$ (left). The edges of the graph in the left pane are colored green for visibility purposes. The quantity $N_{\text{err}}$ denotes the number of edges in the visualization where the cell type of one endpoint does not match the cell type of the other.}
    \label{fig: pbmc}
\end{figure}

We now turn our attention to evaluating EmbedOR on real datasets. We begin by providing further experimental support for \Cref{goal: supplementation}: $\Delta^{\mathcal{E}}$ can be used to identify fragmentation.
To do so, we repeat the experiment from \Cref{table: low-energy graph} on a benchmark single-cell RNA sequencing (scRNAseq) dataset consisting of peripheral blood mononuclear cells available from 10XGenomics. The results of this experiment are shown in \Cref{fig: pbmc}, where we find that the $33\%$ shortest edges under the metric $\Delta^{\mathcal{E}}$ have a nearly $7$-fold drop in the number of cell-type bridging edges compared to the full $k$-NN graph. Like the results in 
\Cref{table: low-energy graph},
this suggests that \Cref{goal: supplementation} applies beyond the synthetic data regime.

In the remaining experiments, we turn our attention back to \Cref{goal: visualization} and present visualization results on benchmark nonlinear dimension reduction datasets, such as MNIST \cite{6296535} and FashionMNIST \cite{xiao2017fashionmnistnovelimagedataset}. We also present results on single-cell RNA sequencing (scRNA-seq) datasets that have been analyzed using tSNE and UMAP in the literature. Guided by the established experimental and theoretical support of \Cref{goal: supplementation}, we use the EmbedOR metric to supplement all visualizations (including baselines) to identify fragmentation. We also quantify the degree to which EmbedOR, UMAP and tSNE avoid fragmentation; to do so we compare the average normalized distances of embedded edges that are deemed close by the EmbedOR metric $\Delta^{\mathcal{E}}$. Finally, we complement this comparison with tests of statistical significance---we describe our testing procedure in \Cref{sec: statistical tests desc}.

\begin{table*}[h]
\footnotesize
\centering
 \begin{tabular}{l | >{\centering\arraybackslash}p{0.125\textwidth}  >{\centering\arraybackslash}p{0.125\textwidth}  >{\centering\arraybackslash}p{0.125\textwidth} >{\centering\arraybackslash}p{0.125\textwidth} >{\centering\arraybackslash}p{0.125\textwidth} } 
  \toprule  & MNIST &  fMNIST & \makecell{Develop. \\ Trajectories \\ (DT) } & Retinal Cells (RC) & \makecell{Chimp Brain \\ Organoids \\ (CBO)} \\ \midrule 
 \hline
 EmbedOR (ours)  & $\underline{-0.37} \pm 0.13$ & $\mathbf{-0.48}\pm 0.22$ &  $\mathbf{-0.62}\pm0.28$ & $\mathbf{-0.27} \pm 0.30$ & $\mathbf{-0.44} \pm 0.16$\\ 
  UMAP \cite{umap}  & $-0.28 \pm 0.16$ & $-0.28\pm0.16$ & $\underline{-0.37}\pm 0.39$ & $-0.19 \pm 0.24$ & $-0.29 \pm 0.22$\\ 
  UMAP + ORCManL \cite{ saidi2025recovering} & $-0.28 
  \pm 0.13$ & $-0.26 \pm 0.32$ & $-0.30 \pm 0.32$ & $-0.22 \pm 0.29$ & $-0.21 \pm 0.26$ \\
  tSNE \cite{tsne} & $\mathbf{-0.37} \pm 0.19$ & $\underline{-0.45}\pm 0.41$ & $-0.23 \pm 0.43$ & $\underline{-0.24} \pm 0.44$ & $\underline{-0.43} \pm 0.31$\\ 
  tSNE + ORCManL \cite{ saidi2025recovering}
 & $-0.32 \pm 0.27$ & $-0.35 \pm 0.34$ & $-0.18 \pm 0.38$ & $-0.24 \pm 0.44$ & $-0.34 \pm 0.28$ \\ \bottomrule
 \end{tabular}
 \caption{Mean and standard deviation of $z$-scored edge lengths among the $33\%$ lowest distance edges (as measured by the EmbedOR metric), reported for stochastic neighbor embedding algorithms. Bold text indicates the best performance, while underlined text indicates the second best.}
 \label{table: low distance edge distortions}
\end{table*}

\subsubsection{MNIST and Fashion-MNIST}

\begin{figure*}[h] 
    \centering
    \includegraphics[width=\linewidth]{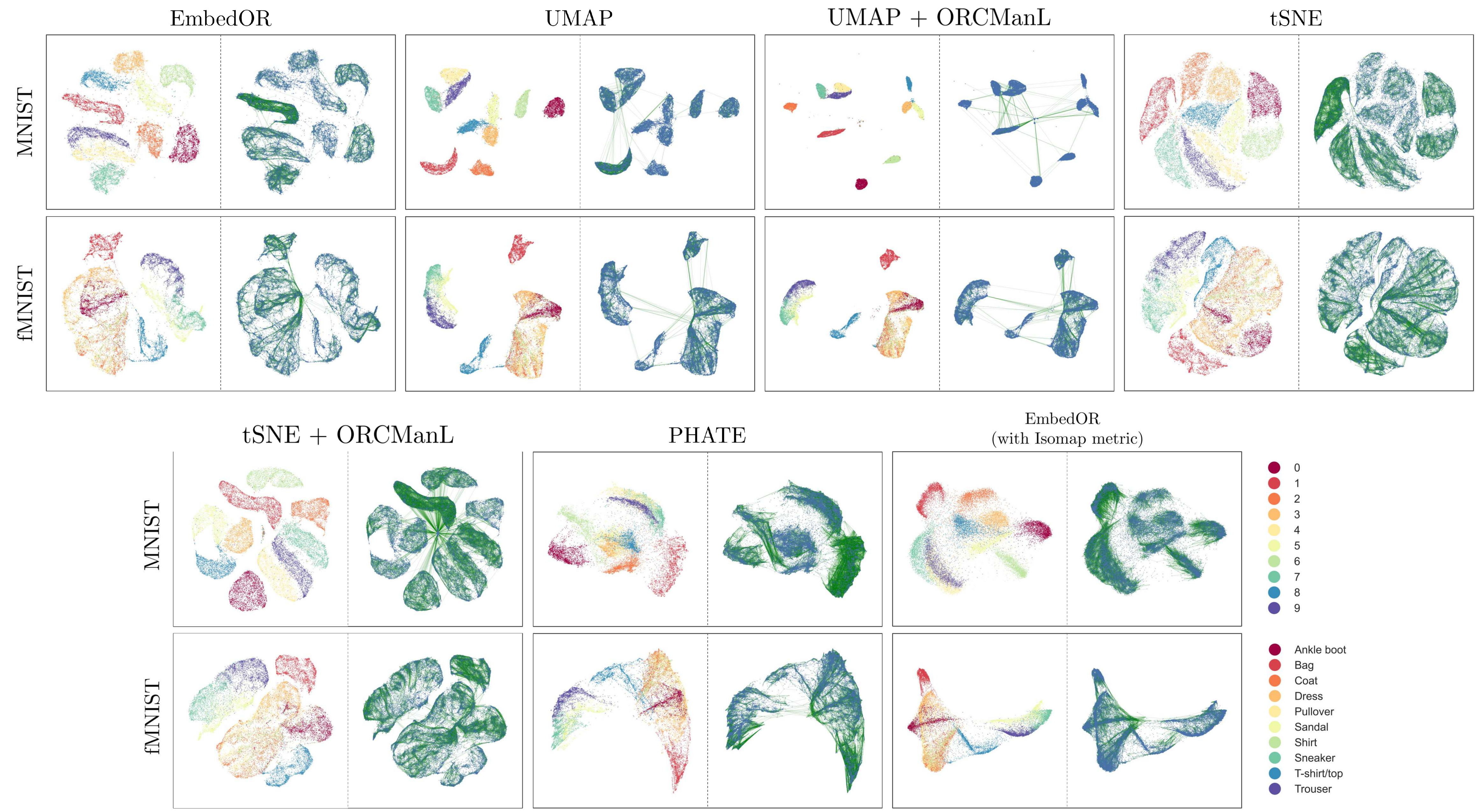}
    \caption{Embeddings produced by various non-linear dimension reduction techniques (using default parameters) of $25,000$ datapoints from the MNIST \cite{6296535} and Fashion-MNIST \cite{xiao2017fashionmnistnovelimagedataset} datasets. Visualizations with class annotations are shown on the left of each pane, while visualizations annotated with edges that have the $33\%$ smallest distances under $\Delta^{\mathcal{E}}$ are shown on the right of each pane.} \label{fig: (f)mnist}
\end{figure*}

\Cref{fig: (f)mnist} illustrates embeddings produced by EmbedOR and baseline algorithms for the MNIST \cite{6296535} and Fashion-MNIST \cite{xiao2017fashionmnistnovelimagedataset} datasets. We find that the SNE algorithms (EmbedOR, UMAP and tSNE) provide the most pronounced separation of the classes. That being said, from visual inspection we find that UMAP embeddings tend to drastically separate some short (w.r.t. $\Delta^{\mathcal{E}}$) edges; we provide a rigorous quantitative analysis of this claim in the following paragraph. Qualitatively, we see that UMAP drastically separates (1) $4$s and $1$s and (2) $7$s and $1$s, while our short $\Delta^{\mathcal{E}}$ edge annotations correctly indicate that those pairs of classes can be quite similar and easily confused. We also see that UMAP + ORCManL and tSNE + ORCManL introduce an extra cluster consisting of several different digits. Finally, we observe that PHATE and EmbedOR with the Isomap metric avoid fragmentation, but do not uncover cluster structure in the data to the same extent as other methods. Overall, we find that EmbedOR strikes a good balance between class visualization and preservation of low-distance edges. 

The expansion of low $\Delta^{\mathcal{E}}$ distance edges by UMAP and tSNE is quantitatively captured in \Cref{table: low distance edge distortions}. This table reports the average $z$-scored distance between pairs of points that the EmbedOR metric, $\Delta^{\mathcal{E}}$, determines to be close. Instead of using raw embedding distances, we choose to compute $z$-scored distances, as many of the dimension reduction techniques (including ours) do not preserve scale. \Cref{table: low distance edge distortions} makes it clear that EmbedOR produces embeddings that minimally expand distances between pairs of points deemed close by $\Delta^{\mathcal{E}}$. \Cref{table: statistical tests} provides statistical support for this claim using the testing framework described in \Cref{sec: statistical tests desc}---we find that the test supports the notion that EmbedOR expands short $\Delta^{\mathcal{E}}$ edges less than UMAP for both the MNIST and the fMNIST datasets, and it supports the same notion for tSNE with the fMNIST dataset. This provides strong statistical support for the claim that UMAP has a stronger tendency to expand edges with a small distance under $\Delta^{\mathcal{E}}$; with that being said, since tSNE expands small $\Delta^{\mathcal{E}}$ edges less than EmbedOR on the MNIST dataset, we are not yet in a position to make an analogous claim about tSNE. However, we will find that tests on subsequent datasets consistently support the claim that tSNE indeed tends to expand these edges more than EmbedOR.

\subsubsection{Single-cell RNA sequencing data}

\begin{figure*}[h]
    \centering
    \includegraphics[width=\linewidth]{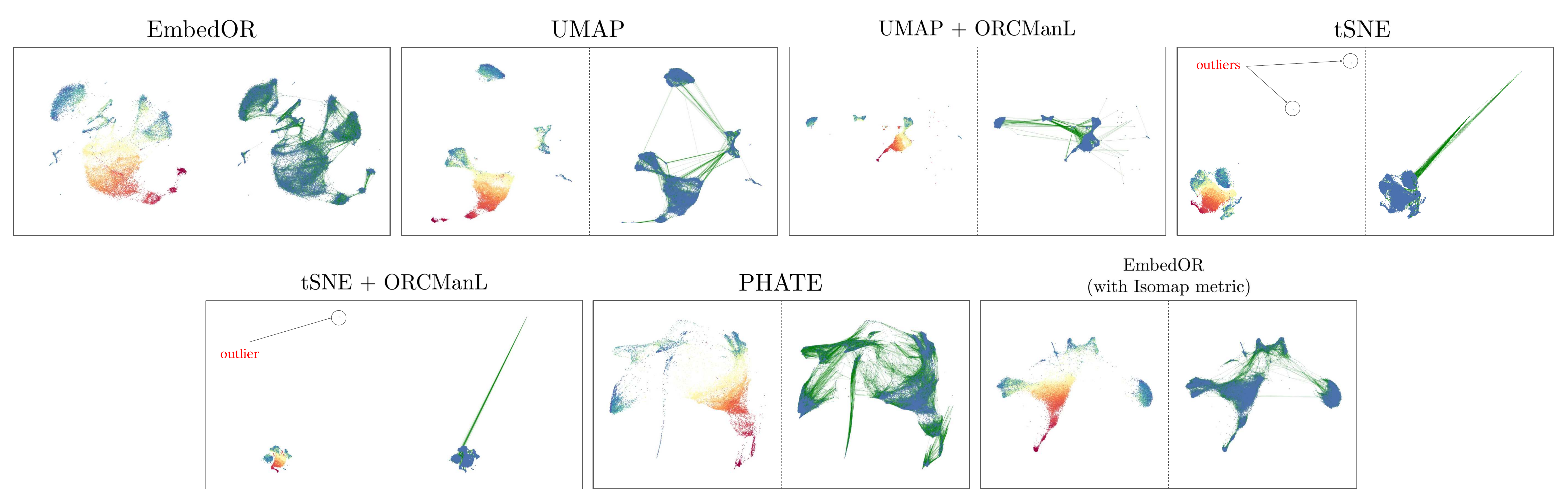}
    \caption{Embeddings produced by various non-linear dimension reduction techniques (using default parameters) of $25,000$ datapoints from a dataset of induced pluripotent stem cells. Visualizations with time annotation (indicated by color) are shown on the left of each pane, while visualizations annotated with edges that have the $33\%$ smallest distances under $\Delta^{\mathcal{E}}$ are shown on the right of each pane.}
    \label{fig: iPSCs}
\end{figure*}

To explore another application domain, we perform the same experiments on publicly available single-cell RNA sequencing (scRNAseq) datasets. In particular, we use data from a time-course of induced pluripotent stem cells from \cite{schiebinger2019optimal} (denoted DT for \say{developmental trajectories}), retinal cells from \cite{macosko2015highly} (denoted RT) and chimpanzee brain organoid cells from \cite{kanton2019organoid} (denoted CBO).

\begin{table}[h]
\small
\centering
 \begin{tabular}{l | >{\centering\arraybackslash}p{0.15\textwidth}  >{\centering\arraybackslash}p{0.15\textwidth}  >{\centering\arraybackslash}p{0.15\textwidth} >{\centering\arraybackslash}p{0.15\textwidth}  >{\centering\arraybackslash}p{0.15\textwidth} } 
  \toprule  & EmbedOR & UMAP* \cite{umap} & tSNE* \cite{tsne} & PHATE \cite{moon2019visualizing} \\ \midrule 
 \hline 
 Spearman  & \rule{0pt}{1ex} $\mathbf{0.6498}$ & $0.6085$ & $\underline{0.6288}$ & $0.5690$\\ 
 Pearson & $\mathbf{0.6709}$ & $0.6195$ & $\underline{0.6452}$ & $0.5669$ \\ 
 \bottomrule
 \end{tabular}
 \caption{Spearman and Pearson correlation coefficients between embedded Euclidean distances and temporal difference for the iPSC developmental trajectories dataset. Bold indicates best, underline indicates second best. Performance of starred methods were obtained by varying hyperparameters and choosing best result, described in \Cref{fig: umap tsne developmental temporal ablation}.} \label{table: temporal correlation}
\end{table}

\Cref{fig: iPSCs} shows the resulting embeddings of the iPSC developmental trajectories dataset. We find that EmbedOR does the best job at preserving the time parameter, which is supported quantitatively by \Cref{table: temporal correlation} - this table establishes that EmbedOR embedded distances correlate more strongly with differences in time than any other method. We note that the UMAP and tSNE results were obtained via a hyperparameter search (the full results of which are displayed in \Cref{fig: umap tsne developmental ablation}), while the EmbedOR results use default parameters described early in \Cref{sec: all experiments}. We also find that EmbedOR even improves upon PHATE \cite{moon2019visualizing}, a method designed specifically for visualizing trajectory data. Returning to the visualization, we highlight the fact that UMAP, tSNE, UMAP + ORCManL and tSNE+ORCManL induce severe fragmentation in the visualization, as indicated by the extreme length of dark green edges (which have small distance as measured by $\Delta^{\mathcal{E}} $)---EmbedOR, however, does not. In particular, we find that UMAP introduces discontinuities in the embedding at the latter stage of the developmental process (identified by the cooler colored datapoints). This fragmentation induced by UMAP and tSNE is again captured in the third column of \Cref{table: low distance edge distortions}, which illustrates that EmbedOR produces embeddings that minimally expand distances between pairs of points deemed close by $\Delta^{\mathcal{E}}$. Statistical significance tests (via the permutation testing framework described in \Cref{sec: statistical tests desc}) of the results shown in \Cref{table: low distance edge distortions} strongly support this claim as well. We find that the tests reject the null hypotheses in favor of the alternate hypotheses that EmbedOR expands short $\Delta^{\mathcal{E}}$ edges \textit{less} than both UMAP and tSNE. Furthermore, these conclusions from the significance tests also hold across parameter ablations of both UMAP and tSNE (\Cref{fig: umap tsne developmental ablation}), providing further support for the claim.

\begin{figure*}[h]
    \centering
    \includegraphics[width=\linewidth]{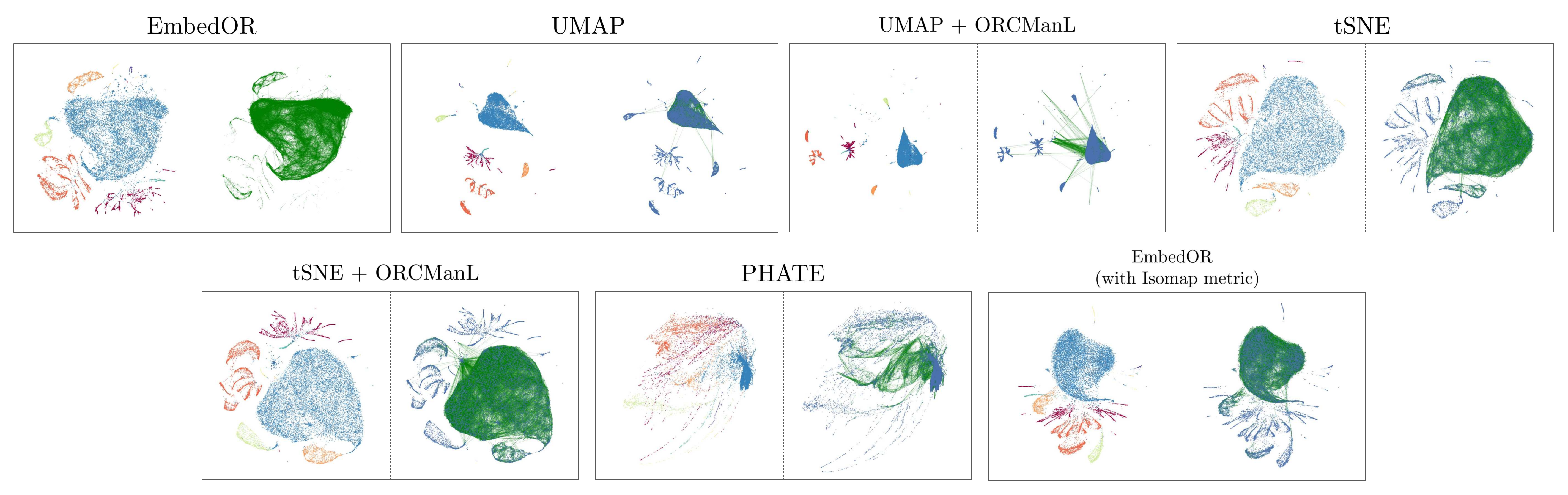}
    \caption{Embeddings produced by various non-linear dimension reduction techniques (using default parameters) of a dataset of $25,000$ retinal cells. Visualizations with cell type annotations (indicated by color) are shown on the left of each pane, while visualizations annotated with edges that have the $33\%$ smallest distances under $\Delta^{\mathcal{E}}$ are shown on the right of each pane.}
    \label{fig: retinal}
\end{figure*}

We now move on to the embeddings of the retinal cell dataset, which are shown in \Cref{fig: retinal}. We find that EmbedOR highlights clusters that align with cell types, while doing well to minimally expand distances between pairs deemed close by the metric. On the other hand, UMAP and UMAP + ORCManL substantially separate cell type clusters that the EmbedOR metric predicts should be close. This is captured in \Cref{table: low distance edge distortions}, which indicates that UMAP, tSNE and their ORCManL augmented counterparts do worse in this regard than EmbedOR, with statistical support indicated again by \Cref{table: statistical tests}. To assess the sensitivity of these conclusions to UMAP and tSNE parameter choices, we re-run the statistical tests across a range of parameters in \Cref{fig: umap tsne macosko ablation}, for which the conclusions are identical across all settings. Jumping back to the visualizations, we observe that EmbedOR with the Isomap metric does a similarly strong job at highlighting clusters that align with cell type annotations, while the remaining embedding approaches largely fail to do so. We thus find that many methods recover many smaller peripheral clusters, suggesting potentially erroneous assumptions about the underlying geometry of this dataset.

\begin{figure*}[h]
    \centering
    \includegraphics[width=\linewidth]{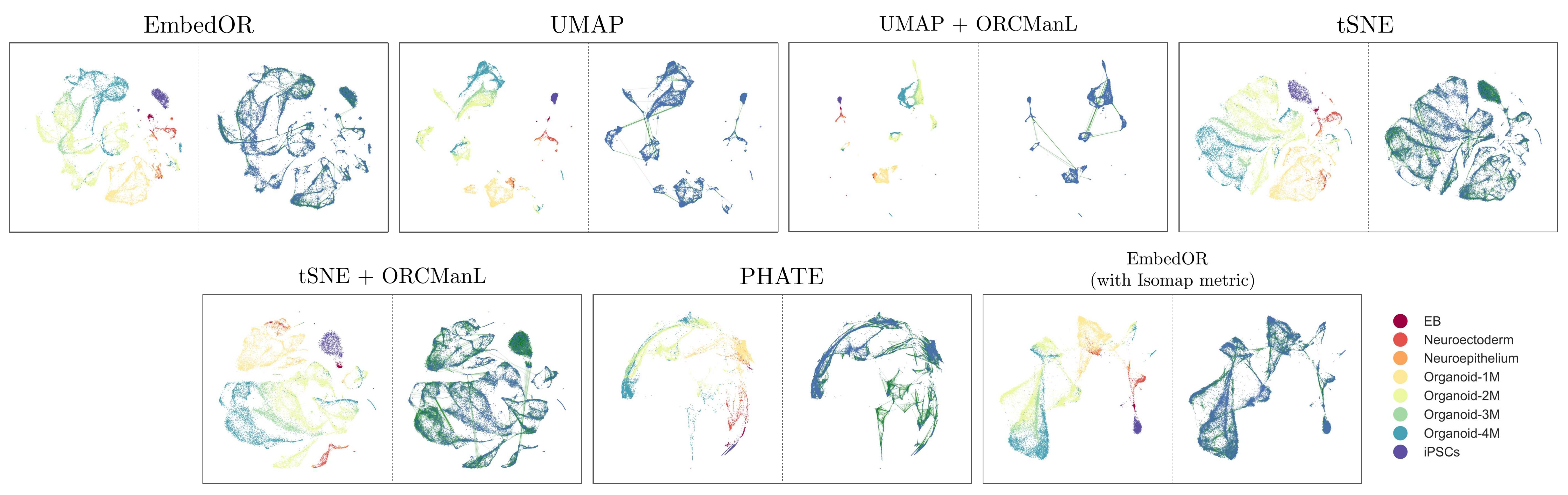}
    \caption{Embeddings produced by various non-linear dimension reduction techniques (using default parameters) of $25,000$ datapoints from a dataset of cells sampled from chimpanzee brain organoids. Visualizations with time annotation (indicated by color) are shown on the left of each pane, while visualizations annotated with edges that have the $33\%$ smallest distances under $\Delta^{\mathcal{E}}$ are shown on the right of each pane.}
    \label{fig: chimp}
\end{figure*}

Pivoting to the chimp organoid cell dataset (shown in \Cref{fig: chimp}), we find that all embedding methods produce fragmented visualizations that make interpretation challenging. However, when supplemented with low-distance edge annotations, the results become more interpretable. In particular, we are able to identify data clusters that were unnecessarily pulled apart by UMAP and UMAP + ORCManL. We also see that most other methods do \textit{not} introduce severe fragmentation in any region under the metric $\Delta^{\mathcal{E}}$, allowing a biologist performing downstream analyses to conclude with higher confidence that the smaller visualized clusters are indeed meaningful. These conclusions are also supported by \Cref{table: low distance edge distortions}, where it becomes clear that EmbedOR and tSNE minimally expand distances between pairs of points deemed close by $\Delta^{\mathcal{E}}$, while UMAP and UMAP + ORCManL do not. Furthermore, we find that these results in \Cref{table: low distance edge distortions} are indeed significant, as we find statistically significant differences in the distribution of $z$-scored distances of small $\Delta^{\mathcal{E}}$ edges (see \Cref{table: statistical tests}). We also note that these results hold across many parameter settings of UMAP and $9$ of $13$ tested parameter settings of tSNE; we provide the results of these experiments and tests in \Cref{fig: umap tsne chimp ablation}. 

\begin{figure}[h]
    \centering
    \includegraphics[width=0.8\linewidth]{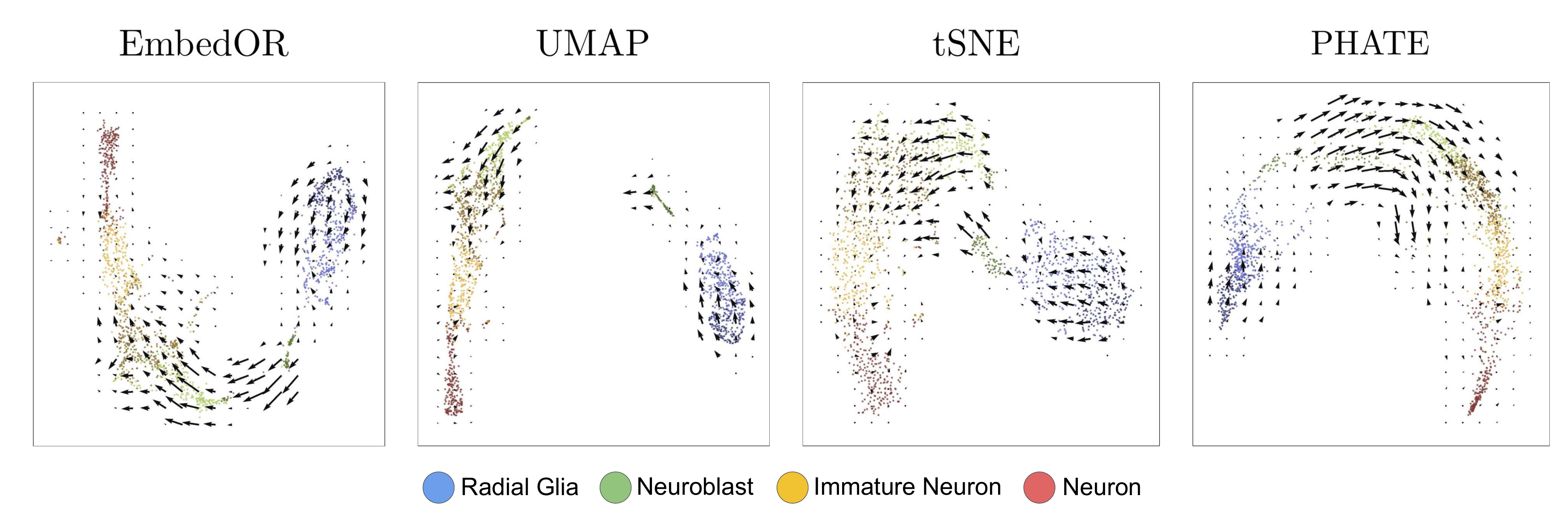}
    \caption{Embeddings produced by various non-linear dimension reduction techniques (using default parameters) of a dataset of human forebrain data. Included are visualizations of the estimated RNA velocity using the {\fontfamily{cmtt}\selectfont velocyto}  package \cite{la2018rna}.}
    \label{fig: trajectory analysis}
\end{figure}

As a final experiment, we reproduce the results from the work of \cite{chari2023specious} that analyzes UMAP and tSNE for trajectory inference in a human forebrain dataset used in \cite{la2018rna}. In this experiment, RNA \say{velocity} is estimated per-cell based on spliced and unspliced gene counts. Then, given an embedding, a projection of the velocity vectors into the embedding space is estimated. The results of this experiment are shown in \Cref{fig: trajectory analysis}. We find that tSNE and especially UMAP induce fragmentation of the underlying trajectory, resulting in the prediction of a discontinuous vector field. We find that EmbedOR produces an embedding comparable to that of PHATE, a method designed explicitly for trajectory preservation. Both methods produce unfragmented embeddings that preserve the principal underlying degree of freedom of the data, with estimated velocity fields that match.

\section{Conclusion}

In this work, we presented EmbedOR, a SNE algorithm that leverages ORC to produce low-dimensional visualizations that better preserve the geometry and topology of the underlying manifold. We provided theoretical results that showed that the use of EmbedOR's distance metric extends results for the SNE framework \cite{pmlr-v75-arora18a} to a broader class of datasets. We supported up these theoretical results by empirically demonstrating that EmbedOR better preserves the topology and underlying connected components of synthetic and real data. 
Our algorithm reveals clear cluster structure while minimizing fragmentation.
Finally, we showed that the EmbedOR distance metric can be used to supplement visualizations of high-dimensional data to identify fragmentation and to better understand the data geometry.

\section{Acknowledgments}
The authors would like to thank Gabriel Guo for his insightful comments about the manuscript and the work.

\bibliography{Research_report}

\begin{thebibliography}{48}
\providecommand{\natexlab}[1]{#1}
\providecommand{\url}[1]{\texttt{#1}}
\expandafter\ifx\csname urlstyle\endcsname\relax
  \providecommand{\doi}[1]{doi: #1}\else
  \providecommand{\doi}{doi: \begingroup \urlstyle{rm}\Url}\fi

\bibitem[McInnes et~al.(2018)McInnes, Healy, and Melville]{umap}
Leland McInnes, John Healy, and James Melville.
\newblock Umap: Uniform manifold approximation and projection for dimension reduction.
\newblock \emph{arXiv preprint arXiv:1802.03426}, 2018.

\bibitem[Van~der Maaten and Hinton(2008)]{tsne}
Laurens Van~der Maaten and Geoffrey Hinton.
\newblock Visualizing data using t-{SNE}.
\newblock \emph{Journal of machine learning research}, 9\penalty0 (11), 2008.

\bibitem[Becht et~al.(2019)Becht, McInnes, Healy, Dutertre, Kwok, Ng, Ginhoux, and Newell]{becht2019dimensionality}
Etienne Becht, Leland McInnes, John Healy, Charles-Antoine Dutertre, Immanuel~WH Kwok, Lai~Guan Ng, Florent Ginhoux, and Evan~W Newell.
\newblock Dimensionality reduction for visualizing single-cell data using umap.
\newblock \emph{Nature biotechnology}, 37\penalty0 (1):\penalty0 38--44, 2019.

\bibitem[Kobak and Berens(2019)]{kobak2019art}
Dmitry Kobak and Philipp Berens.
\newblock The art of using t-sne for single-cell transcriptomics.
\newblock \emph{Nature communications}, 10\penalty0 (1):\penalty0 5416, 2019.

\bibitem[Carter et~al.(2019)Carter, Armstrong, Schubert, Johnson, and Olah]{carter2019exploring}
Shan Carter, Zan Armstrong, Ludwig Schubert, Ian Johnson, and Chris Olah.
\newblock Exploring neural networks with activation atlases.
\newblock \emph{Distill.}, 2019.

\bibitem[Ali et~al.(2019)Ali, Jones, Xie, and Williams]{ali2019timecluster}
Mohammed Ali, Mark~W Jones, Xianghua Xie, and Mark Williams.
\newblock Timecluster: dimension reduction applied to temporal data for visual analytics.
\newblock \emph{The Visual Computer}, 35\penalty0 (6):\penalty0 1013--1026, 2019.

\bibitem[Parmar et~al.(2021)Parmar, Nutter, Long, Antani, and Mitra]{parmar2021visualizing}
Harshit Parmar, Brian Nutter, Rodney Long, Sameer Antani, and Sunanda Mitra.
\newblock Visualizing temporal brain-state changes for fmri using t-distributed stochastic neighbor embedding.
\newblock \emph{Journal of Medical Imaging}, 8\penalty0 (4):\penalty0 046001--046001, 2021.

\bibitem[Liu and Vinck(2022)]{liu2022improved}
Jinke Liu and Martin Vinck.
\newblock Improved visualization of high-dimensional data using the distance-of-distance transformation.
\newblock \emph{PLoS computational biology}, 18\penalty0 (12):\penalty0 e1010764, 2022.

\bibitem[Linderman and Steinerberger(2019)]{linderman2019clustering}
George~C Linderman and Stefan Steinerberger.
\newblock Clustering with t-sne, provably.
\newblock \emph{SIAM journal on mathematics of data science}, 1\penalty0 (2):\penalty0 313--332, 2019.

\bibitem[Arora et~al.(2018)Arora, Hu, and Kothari]{pmlr-v75-arora18a}
Sanjeev Arora, Wei Hu, and Pravesh~K. Kothari.
\newblock An analysis of the t-{SNE} algorithm for data visualization.
\newblock In Sébastien Bubeck, Vianney Perchet, and Philippe Rigollet, editors, \emph{Proceedings of the 31st Conference On Learning Theory}, volume~75 of \emph{Proceedings of Machine Learning Research}, pages 1455--1462. PMLR, 06--09 Jul 2018.
\newblock URL \url{https://proceedings.mlr.press/v75/arora18a.html}.

\bibitem[Yang et~al.(2021)Yang, Chen, and Corander]{yang2021t}
Zhirong Yang, Yuwei Chen, and Jukka Corander.
\newblock T-sne is not optimized to reveal clusters in data.
\newblock \emph{arXiv preprint arXiv:2110.02573}, 2021.

\bibitem[Meil{\u{a}} and Zhang(2024)]{meilua2024manifold}
Marina Meil{\u{a}} and Hanyu Zhang.
\newblock Manifold learning: What, how, and why.
\newblock \emph{Annual Review of Statistics and Its Application}, 11\penalty0 (1):\penalty0 393--417, 2024.

\bibitem[Chari and Pachter(2023)]{chari2023specious}
Tara Chari and Lior Pachter.
\newblock The specious art of single-cell genomics.
\newblock \emph{PLOS Computational Biology}, 19\penalty0 (8):\penalty0 e1011288, 2023.

\bibitem[Moon et~al.(2019)Moon, Van~Dijk, Wang, Gigante, Burkhardt, Chen, Yim, Elzen, Hirn, Coifman, et~al.]{moon2019visualizing}
Kevin~R Moon, David Van~Dijk, Zheng Wang, Scott Gigante, Daniel~B Burkhardt, William~S Chen, Kristina Yim, Antonia van~den Elzen, Matthew~J Hirn, Ronald~R Coifman, et~al.
\newblock Visualizing structure and transitions in high-dimensional biological data.
\newblock \emph{Nature biotechnology}, 37\penalty0 (12):\penalty0 1482--1492, 2019.

\bibitem[Ollivier(2007)]{ollivier2007ricci}
Yann Ollivier.
\newblock Ricci curvature of metric spaces.
\newblock \emph{Comptes Rendus Mathematique}, 345\penalty0 (11):\penalty0 643--646, 2007.

\bibitem[Sia et~al.(2019)Sia, Jonckheere, and Bogdan]{sia2019ollivier}
Jayson Sia, Edmond Jonckheere, and Paul Bogdan.
\newblock {O}llivier-{R}icci curvature-based method to community detection in complex networks.
\newblock \emph{Scientific reports}, 9\penalty0 (1):\penalty0 9800, 2019.

\bibitem[Ni et~al.(2019)Ni, Lin, Luo, and Gao]{ni2019community}
Chien-Chun Ni, Yu-Yao Lin, Feng Luo, and Jie Gao.
\newblock Community detection on networks with {R}icci flow.
\newblock \emph{Scientific reports}, 9\penalty0 (1):\penalty0 9984, 2019.

\bibitem[Saidi et~al.(2025)Saidi, Hickok, and Blumberg]{saidi2025recovering}
Tristan~Luca Saidi, Abigail Hickok, and Andrew~J. Blumberg.
\newblock Recovering manifold structure using {O}llivier {R}icci curvature.
\newblock In \emph{The Thirteenth International Conference on Learning Representations}, 2025.
\newblock URL \url{https://openreview.net/forum?id=aX7X9z3vQS}.

\bibitem[Prasolov(2022)]{prasolov2022differential}
Victor~V Prasolov.
\newblock \emph{Differential Geometry}, volume~8.
\newblock Springer Nature, 2022.

\bibitem[Lee(2018)]{lee2018introduction}
John~M Lee.
\newblock \emph{Introduction to Riemannian manifolds}, volume~2.
\newblock Springer, 2018.

\bibitem[Bernstein et~al.(2000)Bernstein, De~Silva, Langford, and Tenenbaum]{bernstein2000graph}
Mira Bernstein, Vin De~Silva, John~C Langford, and Joshua~B Tenenbaum.
\newblock Graph approximations to geodesics on embedded manifolds.
\newblock Technical report, Technical report, Department of Psychology, Stanford University, 2000.

\bibitem[Tenenbaum et~al.(2000)Tenenbaum, Silva, and Langford]{tenenbaum2000global}
Joshua~B Tenenbaum, Vin~de Silva, and John~C Langford.
\newblock A global geometric framework for nonlinear dimensionality reduction.
\newblock \emph{science}, 290\penalty0 (5500):\penalty0 2319--2323, 2000.

\bibitem[Hinton and Roweis(2002)]{hinton2002stochastic}
Geoffrey~E Hinton and Sam Roweis.
\newblock Stochastic neighbor embedding.
\newblock \emph{Advances in neural information processing systems}, 15, 2002.

\bibitem[Ma{\'c}kiewicz and Ratajczak(1993)]{mackiewicz1993principal}
Andrzej Ma{\'c}kiewicz and Waldemar Ratajczak.
\newblock Principal components analysis (pca).
\newblock \emph{Computers \& Geosciences}, 19\penalty0 (3):\penalty0 303--342, 1993.

\bibitem[Torgerson(1952)]{torgerson1952multidimensional}
Warren~S Torgerson.
\newblock Multidimensional scaling: I. theory and method.
\newblock \emph{Psychometrika}, 17\penalty0 (4):\penalty0 401--419, 1952.

\bibitem[Belkin and Niyogi(2003)]{belkin2003laplacian}
Mikhail Belkin and Partha Niyogi.
\newblock Laplacian eigenmaps for dimensionality reduction and data representation.
\newblock \emph{Neural computation}, 15\penalty0 (6):\penalty0 1373--1396, 2003.

\bibitem[Böhm et~al.(2022)Böhm, Berens, and Kobak]{böhm2022attractionrepulsionspectrumneighborembeddings}
Jan~Niklas Böhm, Philipp Berens, and Dmitry Kobak.
\newblock Attraction-repulsion spectrum in neighbor embeddings, 2022.
\newblock URL \url{https://arxiv.org/abs/2007.08902}.

\bibitem[Pedregosa et~al.(2011)Pedregosa, Varoquaux, Gramfort, Michel, Thirion, Grisel, Blondel, Prettenhofer, Weiss, Dubourg, Vanderplas, Passos, Cournapeau, Brucher, Perrot, and Duchesnay]{scikit-learn}
F.~Pedregosa, G.~Varoquaux, A.~Gramfort, V.~Michel, B.~Thirion, O.~Grisel, M.~Blondel, P.~Prettenhofer, R.~Weiss, V.~Dubourg, J.~Vanderplas, A.~Passos, D.~Cournapeau, M.~Brucher, M.~Perrot, and E.~Duchesnay.
\newblock Scikit-learn: Machine learning in {P}ython.
\newblock \emph{Journal of Machine Learning Research}, 12:\penalty0 2825--2830, 2011.

\bibitem[Linderman et~al.(2019)Linderman, Rachh, Hoskins, Steinerberger, and Kluger]{linderman2019fast}
George~C Linderman, Manas Rachh, Jeremy~G Hoskins, Stefan Steinerberger, and Yuval Kluger.
\newblock Fast interpolation-based t-sne for improved visualization of single-cell rna-seq data.
\newblock \emph{Nature methods}, 16\penalty0 (3):\penalty0 243--245, 2019.

\bibitem[Damrich and Hamprecht(2021)]{damrich2021umap}
Sebastian Damrich and Fred~A Hamprecht.
\newblock On umap's true loss function.
\newblock \emph{Advances in Neural Information Processing Systems}, 34:\penalty0 5798--5809, 2021.

\bibitem[Robbins and Monro(1951)]{robbins1951stochastic}
Herbert Robbins and Sutton Monro.
\newblock A stochastic approximation method.
\newblock \emph{The annals of mathematical statistics}, pages 400--407, 1951.

\bibitem[Zadeh(1965)]{ZADEH1965338}
L.A. Zadeh.
\newblock Fuzzy sets.
\newblock \emph{Information and Control}, 8\penalty0 (3):\penalty0 338--353, 1965.
\newblock ISSN 0019-9958.
\newblock \doi{https://doi.org/10.1016/S0019-9958(65)90241-X}.
\newblock URL \url{https://www.sciencedirect.com/science/article/pii/S001999586590241X}.

\bibitem[Li(2015)]{li2015fuzzy}
Xiang Li.
\newblock Fuzzy cross-entropy.
\newblock \emph{Journal of Uncertainty Analysis and Applications}, 3:\penalty0 1--6, 2015.

\bibitem[Deng(2012)]{6296535}
Li~Deng.
\newblock The {MNIST} database of handwritten digit images for machine learning research [best of the web].
\newblock \emph{IEEE Signal Processing Magazine}, 29\penalty0 (6):\penalty0 141--142, 2012.
\newblock \doi{10.1109/MSP.2012.2211477}.

\bibitem[Xiao et~al.(2017)Xiao, Rasul, and Vollgraf]{xiao2017fashionmnistnovelimagedataset}
Han Xiao, Kashif Rasul, and Roland Vollgraf.
\newblock Fashion-{MNIST}: a novel image dataset for benchmarking machine learning algorithms, 2017.
\newblock URL \url{https://arxiv.org/abs/1708.07747}.

\bibitem[Schiebinger et~al.(2019)Schiebinger, Shu, Tabaka, Cleary, Subramanian, Solomon, Gould, Liu, Lin, Berube, et~al.]{schiebinger2019optimal}
Geoffrey Schiebinger, Jian Shu, Marcin Tabaka, Brian Cleary, Vidya Subramanian, Aryeh Solomon, Joshua Gould, Siyan Liu, Stacie Lin, Peter Berube, et~al.
\newblock Optimal-transport analysis of single-cell gene expression identifies developmental trajectories in reprogramming.
\newblock \emph{Cell}, 176\penalty0 (4):\penalty0 928--943, 2019.

\bibitem[Macosko et~al.(2015)Macosko, Basu, Satija, Nemesh, Shekhar, Goldman, Tirosh, Bialas, Kamitaki, Martersteck, et~al.]{macosko2015highly}
Evan~Z Macosko, Anindita Basu, Rahul Satija, James Nemesh, Karthik Shekhar, Melissa Goldman, Itay Tirosh, Allison~R Bialas, Nolan Kamitaki, Emily~M Martersteck, et~al.
\newblock Highly parallel genome-wide expression profiling of individual cells using nanoliter droplets.
\newblock \emph{Cell}, 161\penalty0 (5):\penalty0 1202--1214, 2015.

\bibitem[Kanton et~al.(2019)Kanton, Boyle, He, Santel, Weigert, Sanch{\'\i}s-Calleja, Guijarro, Sidow, Fleck, Han, et~al.]{kanton2019organoid}
Sabina Kanton, Michael~James Boyle, Zhisong He, Malgorzata Santel, Anne Weigert, F{\'a}tima Sanch{\'\i}s-Calleja, Patricia Guijarro, Leila Sidow, Jonas~Simon Fleck, Dingding Han, et~al.
\newblock Organoid single-cell genomic atlas uncovers human-specific features of brain development.
\newblock \emph{Nature}, 574\penalty0 (7778):\penalty0 418--422, 2019.

\bibitem[La~Manno et~al.(2018)La~Manno, Soldatov, Zeisel, Braun, Hochgerner, Petukhov, Lidschreiber, Kastriti, L{\"o}nnerberg, Furlan, et~al.]{la2018rna}
Gioele La~Manno, Ruslan Soldatov, Amit Zeisel, Emelie Braun, Hannah Hochgerner, Viktor Petukhov, Katja Lidschreiber, Maria~E Kastriti, Peter L{\"o}nnerberg, Alessandro Furlan, et~al.
\newblock Rna velocity of single cells.
\newblock \emph{Nature}, 560\penalty0 (7719):\penalty0 494--498, 2018.

\bibitem[Fesser and Weber(2024)]{fesser2024effective}
Lukas Fesser and Melanie Weber.
\newblock Effective structural encodings via local curvature profiles.
\newblock In \emph{The Twelfth International Conference on Learning Representations}, 2024.
\newblock URL \url{https://openreview.net/forum?id=GIUjLsDP4Z}.

\bibitem[Potamias et~al.(2009)Potamias, Bonchi, Castillo, and Gionis]{potamias2009fast}
Michalis Potamias, Francesco Bonchi, Carlos Castillo, and Aristides Gionis.
\newblock Fast shortest path distance estimation in large networks.
\newblock In \emph{Proceedings of the 18th ACM conference on Information and knowledge management}, pages 867--876, 2009.

\bibitem[Riondato and Kornaropoulos(2014)]{riondato2014fast}
Matteo Riondato and Evgenios~M Kornaropoulos.
\newblock Fast approximation of betweenness centrality through sampling.
\newblock In \emph{Proceedings of the 7th ACM international conference on Web search and data mining}, pages 413--422, 2014.

\bibitem[Sreejith et~al.(2016)Sreejith, Mohanraj, Jost, Saucan, and Samal]{Sreejith_2016}
R~P Sreejith, Karthikeyan Mohanraj, Jürgen Jost, Emil Saucan, and Areejit Samal.
\newblock Forman curvature for complex networks.
\newblock \emph{Journal of Statistical Mechanics: Theory and Experiment}, 2016\penalty0 (6):\penalty0 063206, June 2016.
\newblock ISSN 1742-5468.
\newblock \doi{10.1088/1742-5468/2016/06/063206}.
\newblock URL \url{http://dx.doi.org/10.1088/1742-5468/2016/06/063206}.

\bibitem[Samal et~al.(2018)Samal, Sreejith, Gu, Liu, Saucan, and Jost]{samal2018comparative}
Areejit Samal, RP~Sreejith, Jiao Gu, Shiping Liu, Emil Saucan, and J{\"u}rgen Jost.
\newblock Comparative analysis of two discretizations of ricci curvature for complex networks.
\newblock \emph{Scientific reports}, 8\penalty0 (1):\penalty0 8650, 2018.

\bibitem[Narayan et~al.(2021)Narayan, Berger, and Cho]{narayan2021assessing}
Ashwin Narayan, Bonnie Berger, and Hyunghoon Cho.
\newblock Assessing single-cell transcriptomic variability through density-preserving data visualization.
\newblock \emph{Nature biotechnology}, 39\penalty0 (6):\penalty0 765--774, 2021.

\bibitem[Tsun(2020)]{tsun2020probability}
A~Tsun.
\newblock Probability \& statistics with applications to computing, 2020.

\bibitem[Li(2010)]{li2010concise}
Shengqiao Li.
\newblock Concise formulas for the area and volume of a hyperspherical cap.
\newblock \emph{Asian Journal of Mathematics \& Statistics}, 4\penalty0 (1):\penalty0 66--70, 2010.

\bibitem[{\relax DLMF}()]{NIST:DLMF}
{\relax DLMF}.
\newblock {\it NIST Digital Library of Mathematical Functions}.
\newblock \url{https://dlmf.nist.gov/}, Release 1.2.4 of 2025-03-15.
\newblock URL \url{https://dlmf.nist.gov/}.
\newblock F.~W.~J. Olver, A.~B. {Olde Daalhuis}, D.~W. Lozier, B.~I. Schneider, R.~F. Boisvert, C.~W. Clark, B.~R. Miller, B.~V. Saunders, H.~S. Cohl, and M.~A. McClain, eds.

\end{thebibliography}

\appendix

\section{Computational and memory complexity} \label{sec: comp and memory complexity}

\textbf{Computational complexity}. To analyze the computational complexity, we note that when using $k$-NN graphs in \Cref{alg: EmbedOR metric}, the number of edges scales as $\mathcal{O}(kN)$. According to \cite{fesser2024effective}, computing the ORC for all edges in a graph with $|E|$ edges scales as $\mathcal{O}(|E|k^3)$. All together, this yields a complexity of $\mathcal{O}(Nk^4)$ for the ORC computation stage in \Cref{alg: EmbedOR metric}. To compute the pairwise distance matrix, a call to an APSP algorithm is required, for which an optimized implementation will run in $\mathcal{O}(N^2k+N^2\log N)$. To compute the $\sigma_i$ values we use the subroutine from the scikit-learn tSNE implementation, for which they report a runtime linear in $N$. The SNE stage will have computational complexity of $\mathcal{O}(T)$, where $T$ is the number of iterations chosen by the user. Combining these results, we conclude that the runtime of \Cref{alg: EmbedOR} is approximately $\mathcal{O}(N^2(k+\log N) + Nk^4 +T)$.

\begin{figure}[H]
    \centering
    \includegraphics[width=0.9\linewidth]{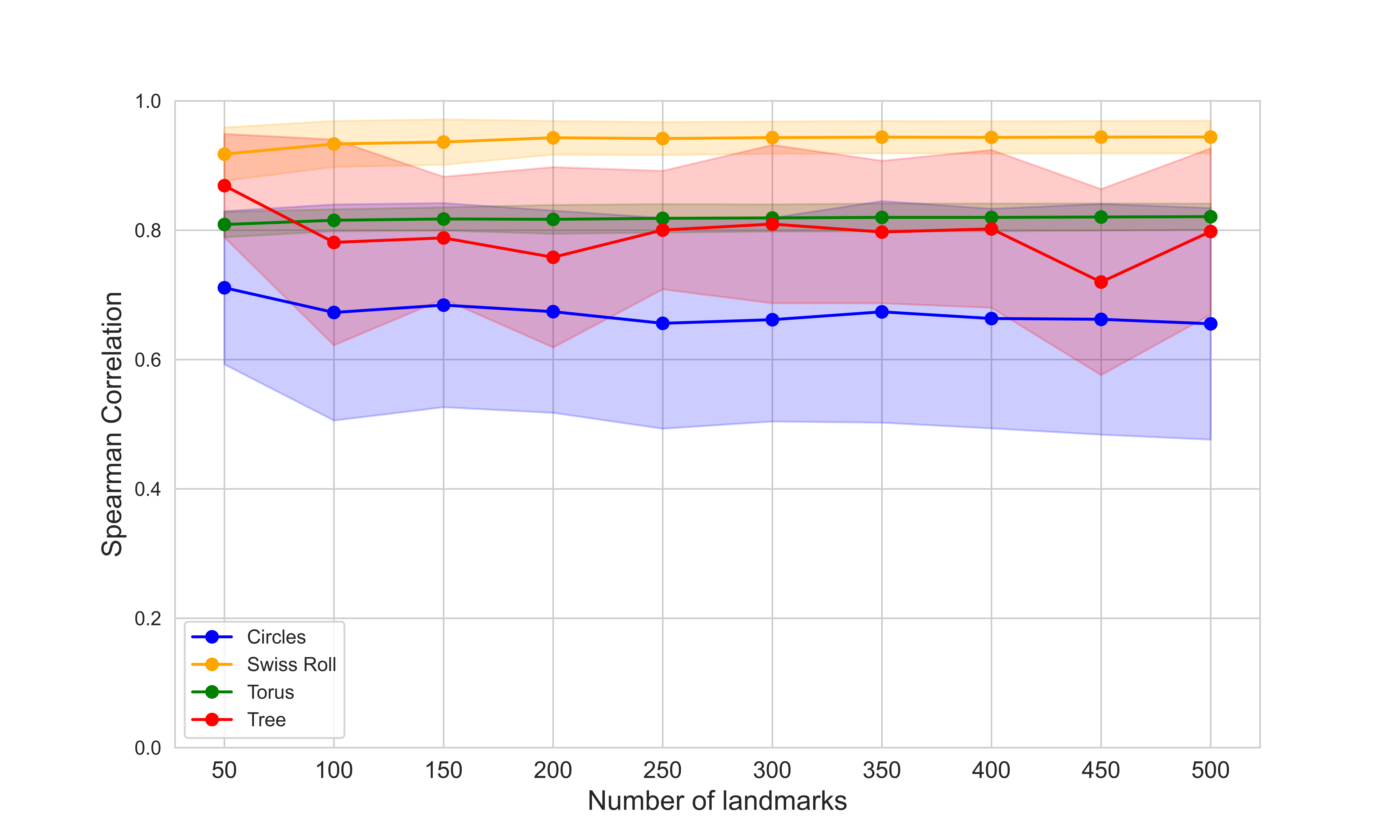}
    \caption{Spearman correlation coefficient between embedded distances and geodesic distances as a function of landmark set sizes for various synthetic datasets of size $5000$. }\label{fig: landmark}
\end{figure}

\begin{figure}[H]
    \centering
    \includegraphics[width=0.4\linewidth]{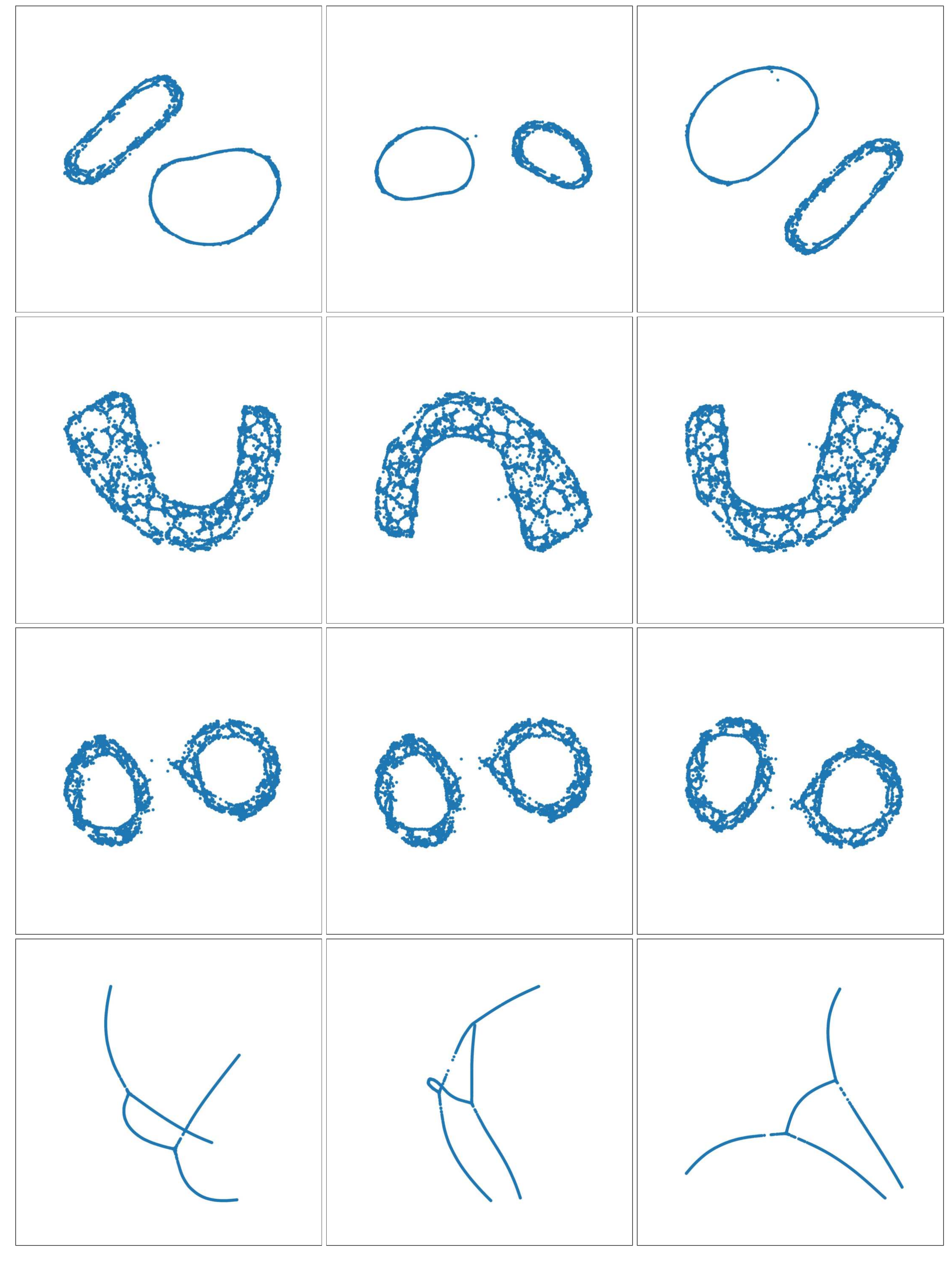}
    \caption{EmbedOR Embeddings of the concentric circles (top), Swiss roll (middle-top), chained torii (middle-bottom) and tree (bottom) datasets using exact APSP (left), landmark APSP with $250$ landmarks (center), and landmark APSP with $50$ landmarks (right). Landmark points are highlighted in red.}\label{fig: landmark embeddings}
\end{figure}

Stochastic Neighbor Embedding algorithms like UMAP and tSNE have been shown to be able to scale to extremely large datasets. We note that the scaling properties of APSP algorithms may bottleneck the ability of EmbedOR to achieve the scaling of UMAP and tSNE; to address this, we provide evidence that a landmark-based approximate APSP approach could be viable to achieve better scaling with EmbedOR. In particular, we employ the landmark-based approximation proposed in \cite{potamias2009fast}, which we describe here.
\begin{enumerate}
    \item (Landmark selection): We select $l \ll N$ landmark points that have the maximum estimated \textit{betweenness centrality}: a number denoting the fraction of shortest paths that travel through that vertex. To compute the betweenness centrality estimates, we use the implementation from \cite{riondato2014fast}. In our setting this computation has time complexity $\mathcal{O}(rNk + rN\log N)$, where $r$ is a factor that depends on the desired accuracy of estimate of betweenness, and the probability of that accuracy being satisfied. We remark that landmarks can also be selected at random - more involved selection schemes like the one we have discussed are likely to improve the accuracy of the landmark-based APSP estimates.
    \item (APSP approximation): Denote $u_1, \dots, u_l \in V$ as the selected landmark nodes. The shortest path between any pair $(x,y) \in V \times V$ is constructed as $\hat{\Delta}^{\mathcal{E}}(x,y) := \max_{j}|d_{G,w}(x,u_j) - d_{G,w}(y,u_j)|$, where $d_{G,w}$ denotes the weighted shortest-path distance through $G$ with respect to the weights described in \Cref{alg: EmbedOR metric}. Observe that, by the triangle inequality we have that $\hat{\Delta}^{\mathcal{E}}(x,y) \leq \Delta^{\mathcal{E}}(x,y).$ Computing $d_{G,w}(\cdot, u_j)$ using Dijkstra's algorithm with a Fibonacci heap has time complexity $\mathcal{O}(Nk + N\log N)$, and thus doing it for all $j \in [l]$ yields a total complexity of $\mathcal{O}(lNk + lN\log N)$. 
\end{enumerate} 
Overall, the computational complexity of this landmark-based APSP approximation scales as $\mathcal{O}\big((r+l)Nk + (l+r)N\log N\big)$. We note that in general $r$ should be independent of $N$ (or can be avoided all together with a naive choice of landmark selection), and $l$ is typically chosen to scale logarithmically with $N$ \cite{potamias2009fast}. This means such an approximation manages to reduce the computation complexity of EmbedOR to sub-quadratic in $N$. We evaluate the feasibility of using this landmark-APSP approximation by replicating the experiments from \Cref{table: geodesic distances} using a variety of landmark set sizes. We provide these experiments in \Cref{fig: landmark}  and \Cref{fig: landmark embeddings} - qualitatively we find that using landmark set sizes anywhere from $1-10\%$ of the size of the dataset has a minimal effect on the resulting embedding. This is backed up quantitatively as well, as for 3 of the 4 synthetic datasets, we see that the Spearman correlation with the ground truth geodesic distance does not drop at all from the correlation using the exact APSP computation (see \Cref{table: geodesic distances}). 


\begin{figure}[H]
    \centering
    \includegraphics[width=0.9\linewidth]{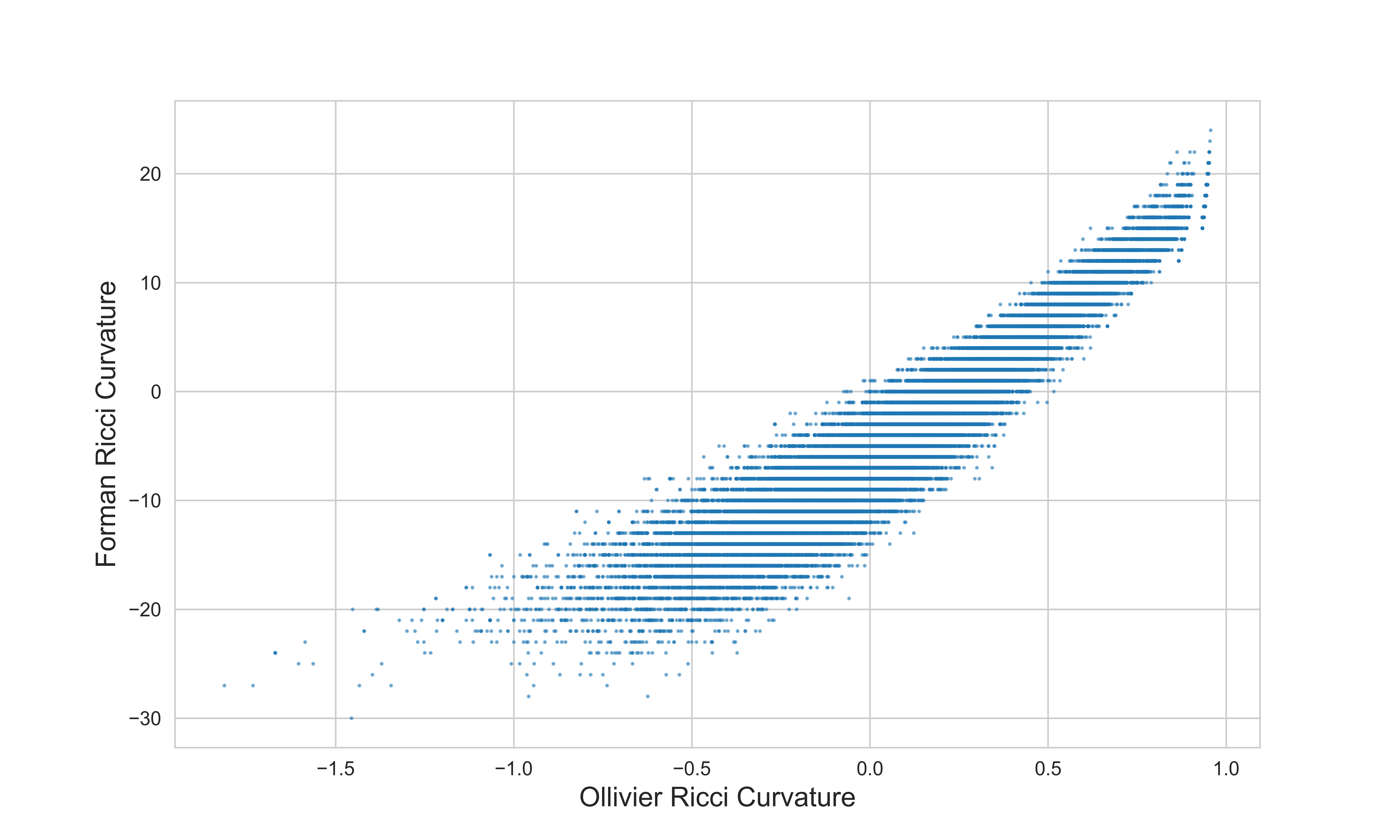}
    \caption{ORC versus FRC scatter plot for the concentric circles dataset, for which we observe a Spearman correlation coefficient of $0.958$ and a Pearson correlation coefficient of $0.952$.}
    \label{fig: frc orc correlation}
\end{figure}

In our implementation, however, we find that one of the primary computational bottlenecks is the ORC computation stage, despite the fact that its theoretical complexity does not impose bottlenecks. We attribute this to the fact that the engineering infrastructure around computing ORC has not been as highly optimized as procedures like computing the APSP of a graph. To this end, we provide some experimental evidence that indicates that, for large scale applications, one can employ Augmented Forman-Ricci Curvature (FRC) \cite{Sreejith_2016, samal2018comparative} in place of ORC and retain comparable results. We choose FRC as it empirically scales much better than ORC, despite the fact that its theoretical complexity shares a linear dependence on $N$ for $k$-NN graphs. \Cref{fig: frc orc correlation} shows the correlation of the two notions of graph curvature on the concentric circles dataset, where we observe a strong linear relationship.

\begin{figure}[H]
    \centering
    \includegraphics[width=0.7\linewidth]{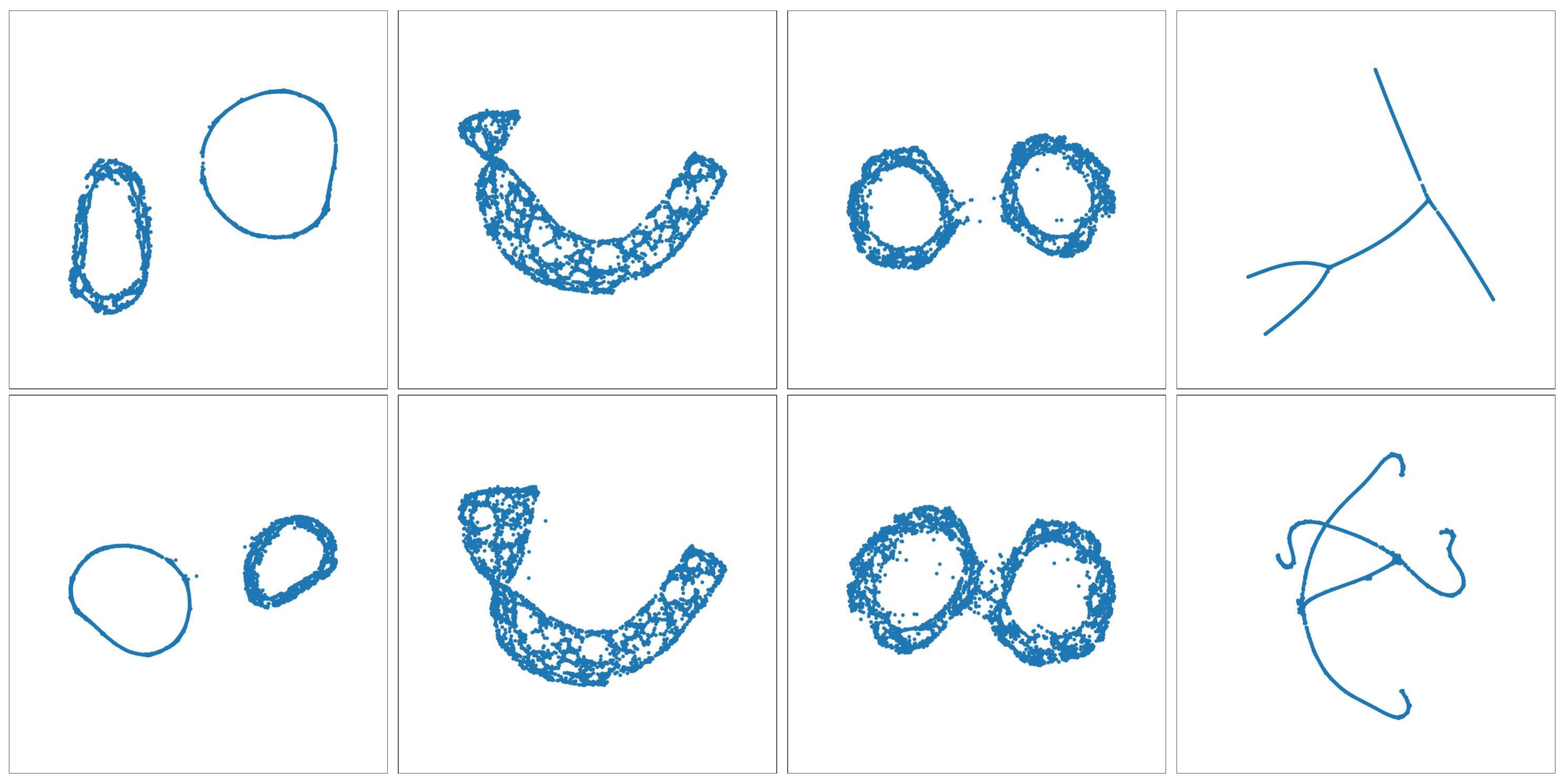}
    \caption{EmbedOR embeddings of $5000$ points sampled from the concentric circles (left), Swiss roll (middle left), chained torii (middle right) and tree (right) datasets using ORC and FRC based EmbedOR.}
    \label{fig: frc}
\end{figure}

This strong correlation supports the idea that one can substitute FRC for ORC in \Cref{alg: EmbedOR} to obtain speedups for large datasets. Fortunately, we find a minimal drop in visualization quality for three of four synthetic datasets, as indicated in \Cref{fig: frc} and \Cref{table: frc}. 

\begin{table}[h]
\small
\centering
 \begin{tabular}{>{\centering\arraybackslash}p{0.15\textwidth}  >{\centering\arraybackslash}p{0.15\textwidth}  >{\centering\arraybackslash}p{0.15\textwidth} >{\centering\arraybackslash}p{0.15\textwidth}   } 
  \toprule
   Concentric Circles & Swiss Roll & Torii & Tree \\
  \midrule \hline
  $0.729 \pm 0.164$ & $0.953 \pm 0.010$ & $0.749 \pm 0.054$ & $0.570 \pm 0.081$ \\ 
 \bottomrule
 \end{tabular}
 \caption{Spearman correlation coefficient between embedded distances and geodesic distances when using FRC instead of ORC.}
 \label{table: frc}
\end{table}

\begin{figure}[H]
    \centering
    \includegraphics[width=0.9\linewidth]{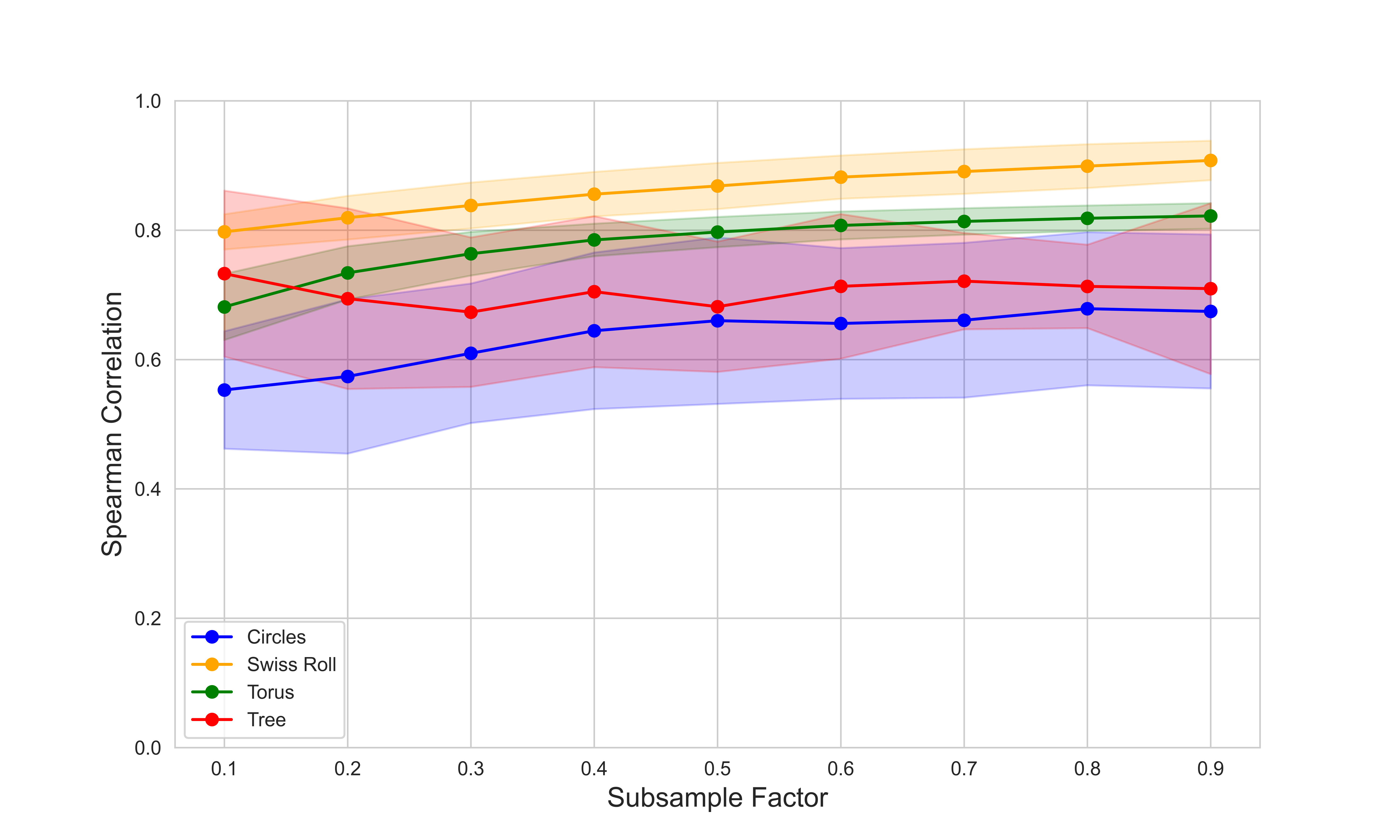}
    \caption{Spearman correlation coefficient between embedded distances and geodesic distances as a function of subsampling level for various synthetic datasets. }\label{fig: subsample}
\end{figure}

\begin{figure}[H]
    \centering
    \includegraphics[width=0.4\linewidth]{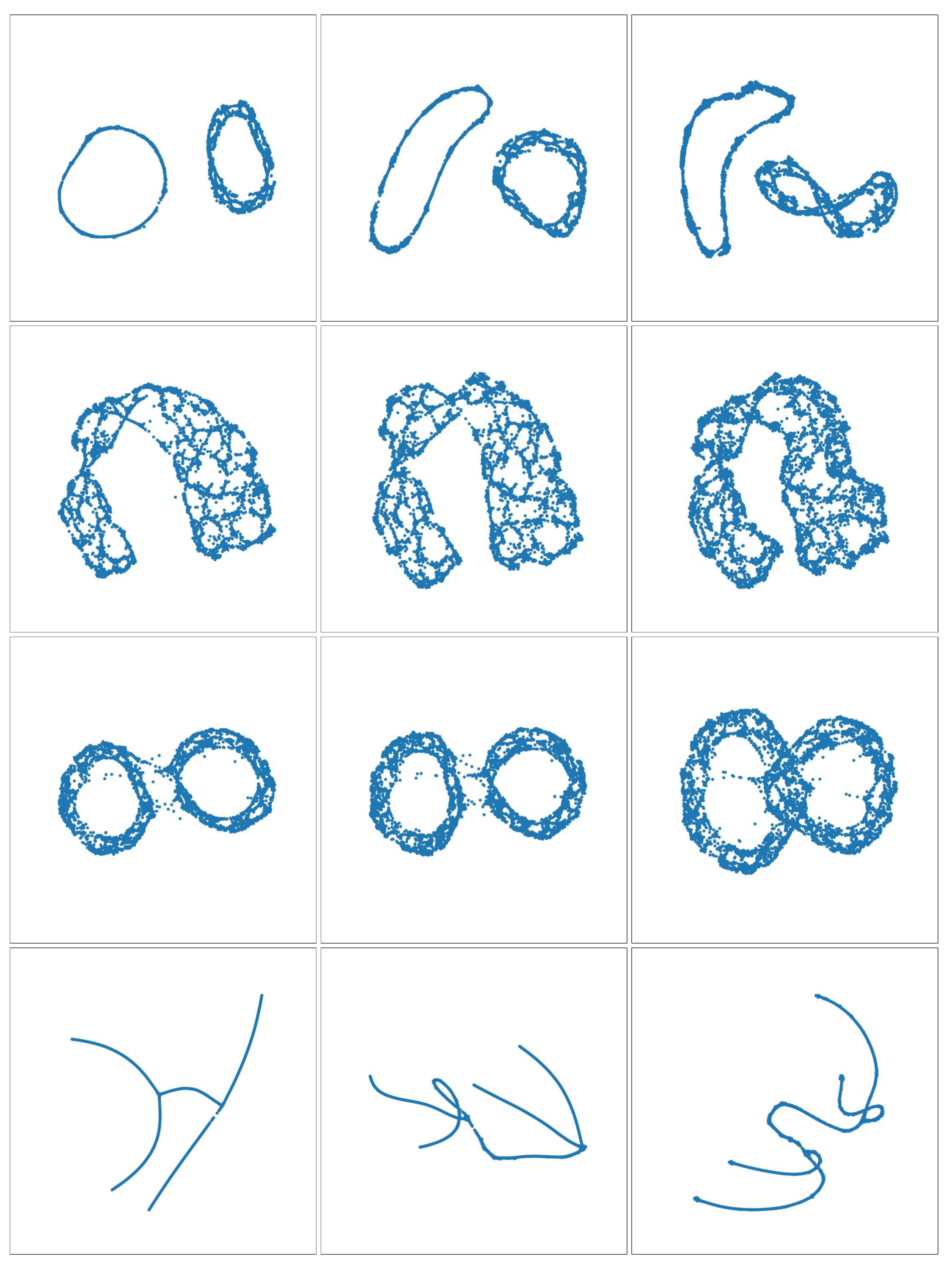}
    \caption{EmbedOR Embeddings of the concentric circles (top), Swiss roll (middle-top), chained torii (middle-bottom) and tree (bottom) datasets using all pairwise interactions (left), a $50\%$ subsample (middle), and a $10\%$ subsample (right). }\label{fig: subsampled embedding}
\end{figure}

\noindent \textbf{Memory Complexity.} The need for $\mathcal{O}(N^2)$ pairwise affinities and repulsions during the SGD portion of \Cref{alg: EmbedOR} renders its memory complexity $\mathcal{O}(N^2)$. We note that this memory complexity is identical to that of tSNE when dense affinities are used. We note that popular implementations of tSNE do not by default use dense affinities in order to achieve better scaling \cite{linderman2019fast, scikit-learn}. To assess whether or not similar ideas can be applied to EmbedOR, we replicate the experiments from \Cref{table: geodesic distances} with subsampling (with results provided in \Cref{fig: subsample} and \Cref{fig: subsampled embedding}). We find that subsampling indeed degrades performance, but at a non-aggressive rate. The geodesic distance correlations fall to around that of the average UMAP and tSNE performance across hyperparameter settings (see \Cref{table: geodesic distances umap ablation} and \Cref{table: geodesic distances tsne ablation}). Overall, we believe that these results indicate that EmbedOR has the potential to avoid the scaling bottlenecks imposed by the O($N^2$) pairwise interactions being considered. We remark that, in the experiments described above, we use a uniform distribution to subsample the interactions; we believe that a more thorough investigation into sampling distributions that leverage information about the metric $\Delta^{\mathcal{E}}$ might mitigate performance degradation when subsampling. We think that this is a fruitful direction for future study.

\begin{figure}[H]
    \centering
    \includegraphics[width=0.9\linewidth]{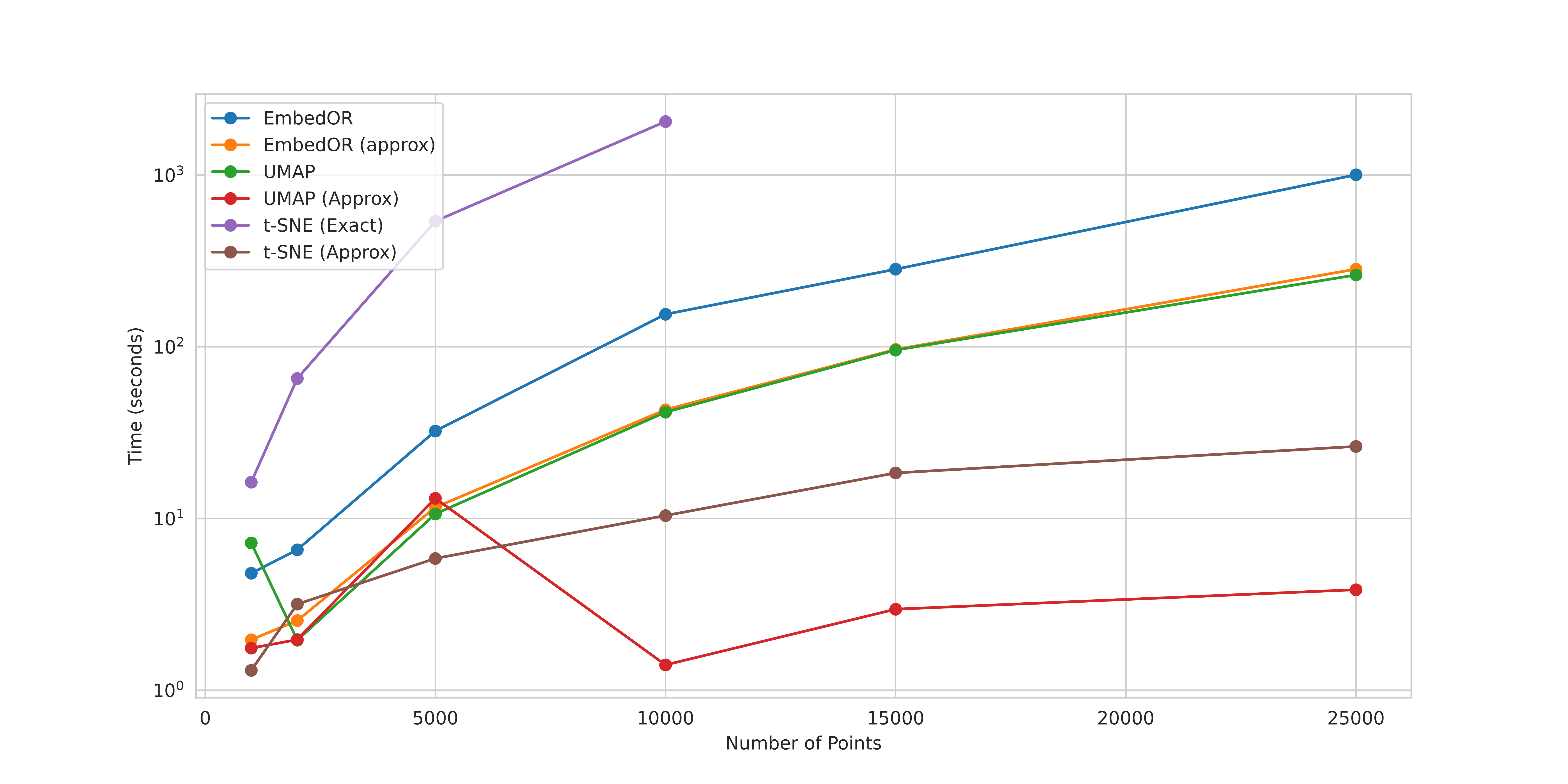}
    \caption{Runtime of EmbedOR, EmbedOR with approximations (APSP approximation with $50$ landmarks, $20\%$ subsampling of pairwise interactions and FRC), UMAP without approximations, standard UMAP, exact tSNE and standard tSNE as a function of dataset size. For this experiment, all algorithms were run on a 16-core machine with 187 GB of RAM. }\label{fig: runtime}
\end{figure}

Finally, we include empirical runtimes for EmbedOR, UMAP and tSNE in \Cref{fig: runtime}. For each method we include the runtime of the exact algorithm, and the approximate algorithm. For EmbedOR with approximations, we use $50$ landmarks, a $20\%$ subsample of pairwise interactions, and FRC instead of ORC. For UMAP without approximation, we comment out a portion of their implementation that employs additional approximations for large datasets.

\section{Additional Experiments}

\subsection{Statistical testing}\label{sec: statistical tests desc}

To assess the significance of the results of \Cref{table: low distance edge distortions} in the main manuscript, we conduct statistical testing of the following form: define $d_{ij}^{\text{ALG}} \overset{\text{def}}{=} \|y_i^{\text{ALG}} - y_i^{\text{ALG}}\|_2$ where $y_i^{\text{ALG}}$ are embeddings produced by ALG. Further define the empirical statistics
\[\bar{\mu}_{\text{ALG}} = \frac{1}{|E|}\sum_{(i,j) \in E}d_{ij}^{\text{ALG}}\]
and 
\[\bar{\sigma}^2_{\text{ALG}} = \frac{1}{|E|-1}\sum_{(i,j) \in E}\Big(d_{ij}^{\text{ALG}}-\bar{\mu}_{\text{ALG}}\Big)^2\]
where $E$ is the set of edges in the data's nearest neighbor graph. Finally, define the \textit{empirically} $z$-scored embedded distances,
\[\tilde{d}_{ij}^{\text{ALG}} = \frac{d_{ij}^{\text{ALG}} - \bar{\mu}_{\text{ALG}}}{\bar{\sigma}_{\text{ALG}}}.\]
For each dataset we run two permutation tests with $10,000$ resamples with the null hypotheses (1) $\mu^{\text{UMAP}}_{\beta} = \mu^{\text{EmbedOR}}_{\beta}$ and (2) $\mu^{\text{tSNE}}_{\beta} = \mu^{\text{EmbedOR}}_{\beta}$, and alternate hypotheses (1) $\mu^{\text{UMAP}}_{\beta} > \mu^{\text{EmbedOR}}_{\beta}$ and (2) $\mu^{\text{tSNE}}_{\beta} > \mu^{\text{EmbedOR}}_{\beta}$ for $\beta =0.33$. We also run all tests with a confidence level $\alpha = 0.01$ to ensure significance. Here, we define
\begin{equation}\label{eq: statistical test distribution mean}
    \mu^{\text{ALG}}_{\beta} = \mathbb{E}\Big[\tilde{d}_{ij}^{\,\text{ALG}} \, \Big|\, (i,j) \in E_{\beta}\Big]
\end{equation}
and $E_{\beta}$ is defined to be the set of the smallest $(100\beta )\%$ of edges in $E$ as measured by $\Delta^{\mathcal{E}}$. Put simply, this statistical test captures whether EmbedOR expands low $\Delta^{\mathcal{E}}$ edges less  (relative to the entire embedding) than UMAP and tSNE. To emphasize an important point: we use $z$-scored distances as scale is not preserved by EmbedOR, UMAP or tSNE. Now that we have explained our testing procedure, we provide the results of such tests in \Cref{table: statistical tests}.

\begin{table}[H]
\small
\centering
 \begin{tabular}{l | >{\centering\arraybackslash}p{0.30\textwidth}  >{\centering\arraybackslash}p{0.3\textwidth}  >{\centering\arraybackslash}p{0.3\textwidth} } 
  \toprule  
  & \makecell{$H_0: \mu_{0.33}^{\text{UMAP}} = \mu_{0.33}^{\text{EmbedOR}}$ \\ $H_1: \mu_{0.33}^{\text{UMAP}} > \mu_{0.33}^{\text{EmbedOR}}$} 
  & \makecell{$H_0: \mu_{0.33}^{\text{tSNE}} = \mu_{0.33}^{\text{EmbedOR}}$  \\ $H_1: \mu_{0.33}^{\text{tSNE}} > \mu_{0.33}^{\text{EmbedOR}}$}  \\ \midrule 
 \hline 
 MNIST  & \rule{0pt}{3ex} \Checkmark & \XSolidBrush \\ 
 fMNIST & \Checkmark & \Checkmark \\ 
 DT & \Checkmark & \Checkmark \\
 RC & \Checkmark & \Checkmark \\
 CBO & \Checkmark & \Checkmark \\
 \bottomrule
 \end{tabular}
 \caption{Permutation test using $10,000$ resamples and $\alpha = 0.01$ to evaluate statistical significance of results presented in \Cref{table: low distance edge distortions}, with dataset acronyms defined there. We note that $\mu_{0.33}^{\text{ALG}}$ is defined in \cref{eq: statistical test distribution mean}. A (\Checkmark) denotes a rejection of the null hypothesis, while a (\XSolidBrush) denotes failure to reject the null.}
 \label{table: statistical tests}
\end{table}

Now we provide statistical tests that address the scenario \textit{converse} to that explored in \Cref{table: low distance edge distortions} and above. To recall, these tables explore the degree to which EmbedOR, UMAP and tSNE \textit{expand} edges assigned a low value under the metric $\Delta^{\mathcal{E}}$. In this section, we seek to understand the degree to which EmbedOR, UMAP, and tSNE contract edges that have a larger $\Delta^{\mathcal{E}}$. To do this, we look at the difference in the average $\Delta^{\mathcal{E}}$ values of edges with the $33\%$ smallest embedded distance. These results are shown in \Cref{table: low distance edge distortions (2)}, with accompanying tests of statistical significance in \Cref{table: statistical tests (2)}. Similar to \cref{eq: statistical test distribution mean}, we define
\begin{equation}\label{eq: statistical test distribution mean (2)}
    \mu_{\beta, \bullet}^{\text{ALG}} = \mathbb{E}\Big[\Delta^{\mathcal{E}}(x_i, x_j) \, \Big | \, (y_i, y_j) \in E_{\beta}^{\text{ALG}}\Big] 
\end{equation}
where $E^{\text{ALG}}_{\beta}$ is the set of the smallest $(100\beta)\%$ ALG-embedded edges in $E$. When taken together, \Cref{table: statistical tests} and \Cref{table: statistical tests (2)} indicate that 
\begin{enumerate}
    \item We can be confident that UMAP tends to expand low $\Delta^\mathcal{E}$ edges more than EmbedOR, and we can be confident that UMAP tends to contract higher $\Delta^\mathcal{E}$ edges more than EmbedOR. 
    \item We can be confident that tSNE tends to expand low $\Delta^\mathcal{E}$ edges more than EmbedOR, but we \textit{cannot} be confident that tSNE tends to contract higher $\Delta^\mathcal{E}$ edges more than EmbedOR. 
\end{enumerate}

\begin{table}[H]
\small
\centering
 \begin{tabular}{l | >{\centering\arraybackslash}p{0.125\textwidth}  >{\centering\arraybackslash}p{0.125\textwidth}  >{\centering\arraybackslash}p{0.125\textwidth} >{\centering\arraybackslash}p{0.125\textwidth} >{\centering\arraybackslash}p{0.125\textwidth} } 
  \toprule  & MNIST &  fMNIST & \makecell{DT} & RC & \makecell{CBO} \\ \midrule 
 \hline
 EmbedOR (ours)  & $\mathbf{-0.33} \pm 1.02$ & $\underline{-0.44}\pm 1.04$ &  $\mathbf{-0.65}\pm1.00$ & $\mathbf{-0.18} \pm 1.06$ & $-0.47 \pm 0.99$\\ 
 UMAP \cite{umap}  & $-0.26 \pm 1.02$ & $-0.39\pm1.04$ & $-0.55\pm 1.02$ & $-0.15 \pm 1.06$ & $-\underline{0.47} \pm 1.01$\\ 
 tSNE \cite{tsne} & $\underline{-0.33} \pm 1.02$ & $\mathbf{-0.45}\pm 1.04$ & $\underline{-0.63} \pm 1.00$ & $\underline{-0.17} \pm 1.07$ & $\mathbf{-0.56} \pm 1.00$\\ 
 \bottomrule
 \end{tabular}
 \caption{Mean and standard deviation of $z$-scored edge lengths under $\Delta^{\mathcal{E}}$ among the $33\%$ lowest distance edges in the embedding, reported for stochastic neighbor embedding algorithms. Bold text indicates the best performance, while underlined text indicates the second best.}
 \label{table: low distance edge distortions (2)}
\end{table}

\begin{table}[H]
\small
\centering
 \begin{tabular}{l | >{\centering\arraybackslash}p{0.3\textwidth}  >{\centering\arraybackslash}p{0.3\textwidth}  >{\centering\arraybackslash}p{0.3\textwidth} } 
  \toprule  
  & \makecell{$H_0: \mu_{0.33, \bullet}^{\text{UMAP}} = \mu_{0.33, \bullet}^{\text{EmbedOR}}$ \\ $H_1: \mu_{0.33, \bullet}^{\text{UMAP}} > \mu_{0.33, \bullet}^{\text{EmbedOR}}$} 
  & \makecell{$H_0: \mu_{0.33, \bullet}^{\text{tSNE}} = \mu_{0.33, \bullet}^{\text{EmbedOR}}$  \\ $H_1: \mu_{0.33, \bullet}^{\text{tSNE}} > \mu_{0.33, \bullet}^{\text{EmbedOR}}$}  \\ \midrule 
 \hline 
 MNIST  & \rule{0pt}{3ex} \Checkmark & \XSolidBrush \\ 
 fMNIST & \Checkmark & \XSolidBrush \\ 
 DT & \Checkmark & \Checkmark \\
 RC & \Checkmark & \XSolidBrush \\
 CBO & \XSolidBrush & \XSolidBrush \\
 \bottomrule
 \end{tabular}
 \caption{Results of permutation testing using significance level $\alpha = 0.01$ and $10,000$ resamples to evaluate statistical significance of results presented in \Cref{table: low distance edge distortions (2)}, with dataset acronyms defined there. A (\Checkmark) denotes a rejection of the null hypothesis, while a (\XSolidBrush) denotes a failure to reject the null.}
 \label{table: statistical tests (2)}
\end{table}

\subsection{Additional Visualizations}
To save space in the paper, we have deferred visualizations of real-world datasets produced by Laplacian Eigenmaps \cite{belkin2003laplacian} and Isomap \cite{tenenbaum2000global} to this section.

\begin{figure}[H] 
    \centering
    \includegraphics[width=\linewidth]{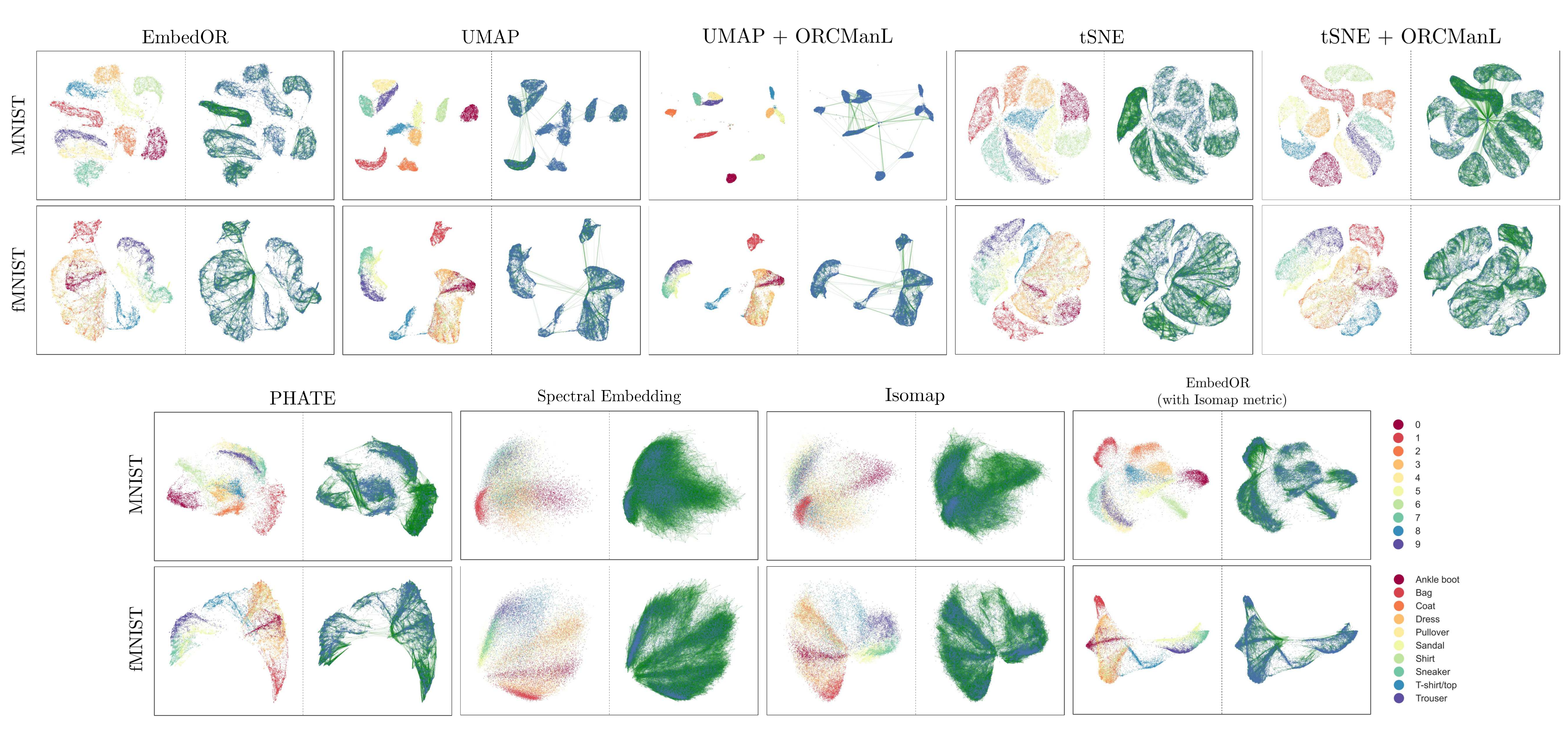}
    \caption{Embeddings produced by various non-linear dimension reduction techniques (using default parameters) of $25,000$ datapoints from the MNIST \cite{6296535} and Fashion-MNIST \cite{xiao2017fashionmnistnovelimagedataset} datasets. Visualizations with class annotations are shown on the left of each pane, while visualizations annotated with edges that have the $33\%$ smallest distances under $\Delta^{\mathcal{E}}$ are shown on the right of each pane.} \label{fig: (f)mnist extra}
\end{figure}

\begin{figure}[H]
    \centering
    \includegraphics[width=\linewidth]{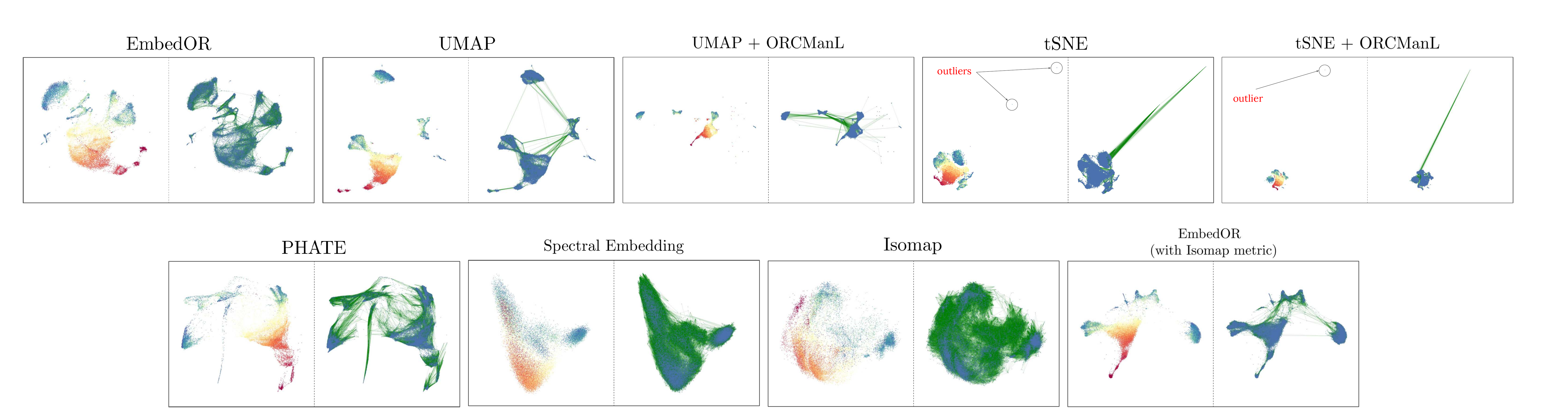}
    \caption{Embeddings produced by various non-linear dimension reduction techniques (using default parameters) of $25,000$ datapoints from a dataset of induced pluripotent stem cells. Visualizations with time annotation (indicated by color) are shown on the left of each pane, while visualizations annotated with edges that have the $33\%$ smallest distances under $\Delta^{\mathcal{E}}$ are shown on the right of each pane.}
    \label{fig: iPSCs extra}
\end{figure}

\begin{figure}[H]
    \centering
    \includegraphics[width=\linewidth]{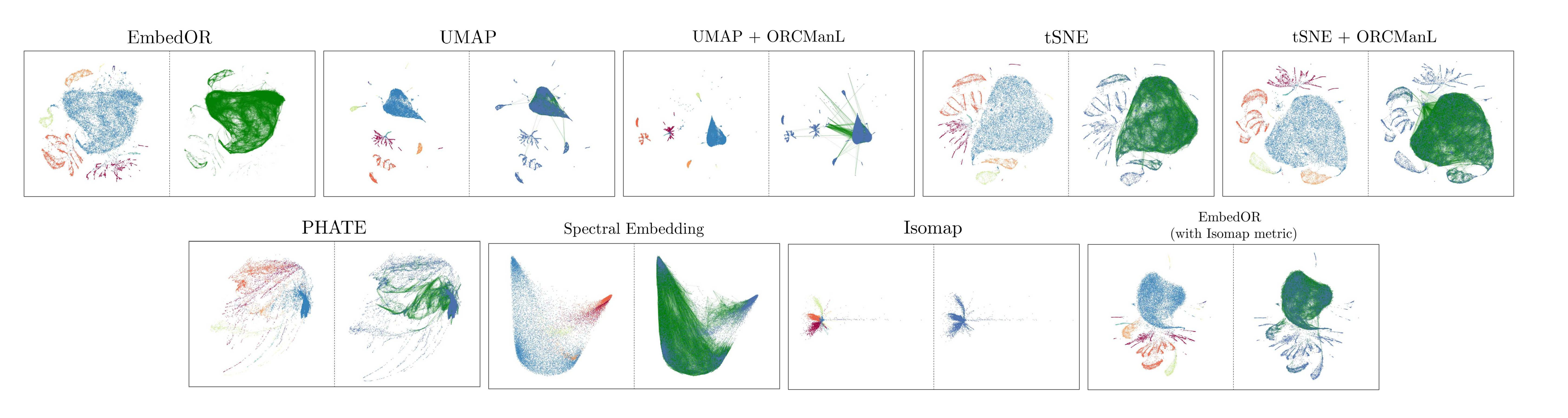}
    \caption{Embeddings produced by various non-linear dimension reduction techniques (using default parameters) of a dataset of $25,000$ retinal cells. Visualizations with cell type annotations (indicated by color) are shown on the left of each pane, while visualizations annotated with edges that have the $33\%$ smallest distances under $\Delta^{\mathcal{E}}$ are shown on the right of each pane.}
    \label{fig: retinal extra}
\end{figure}

\begin{figure}[h]
    \centering
    \includegraphics[width=\linewidth]{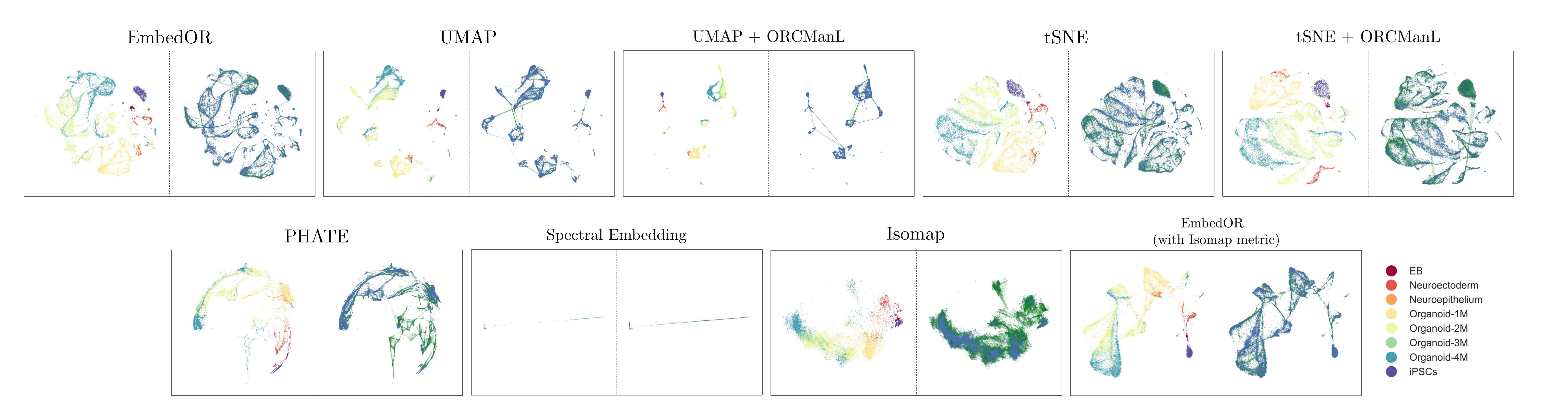}
    \caption{Embeddings produced by various non-linear dimension reduction techniques (using default parameters) of $25,000$ datapoints from a dataset of cells sampled from chimpanzee brain organoids. Visualizations with time annotation (indicated by color) are shown on the left of each pane, while visualizations annotated with edges that have the $33\%$ smallest distances under $\Delta^{\mathcal{E}}$ are shown on the right of each pane.}
    \label{fig: chimp extra}
\end{figure}

\subsection{Scale sensitivity}
\begin{figure}[h]
    \centering
    \includegraphics[width=0.9\linewidth]{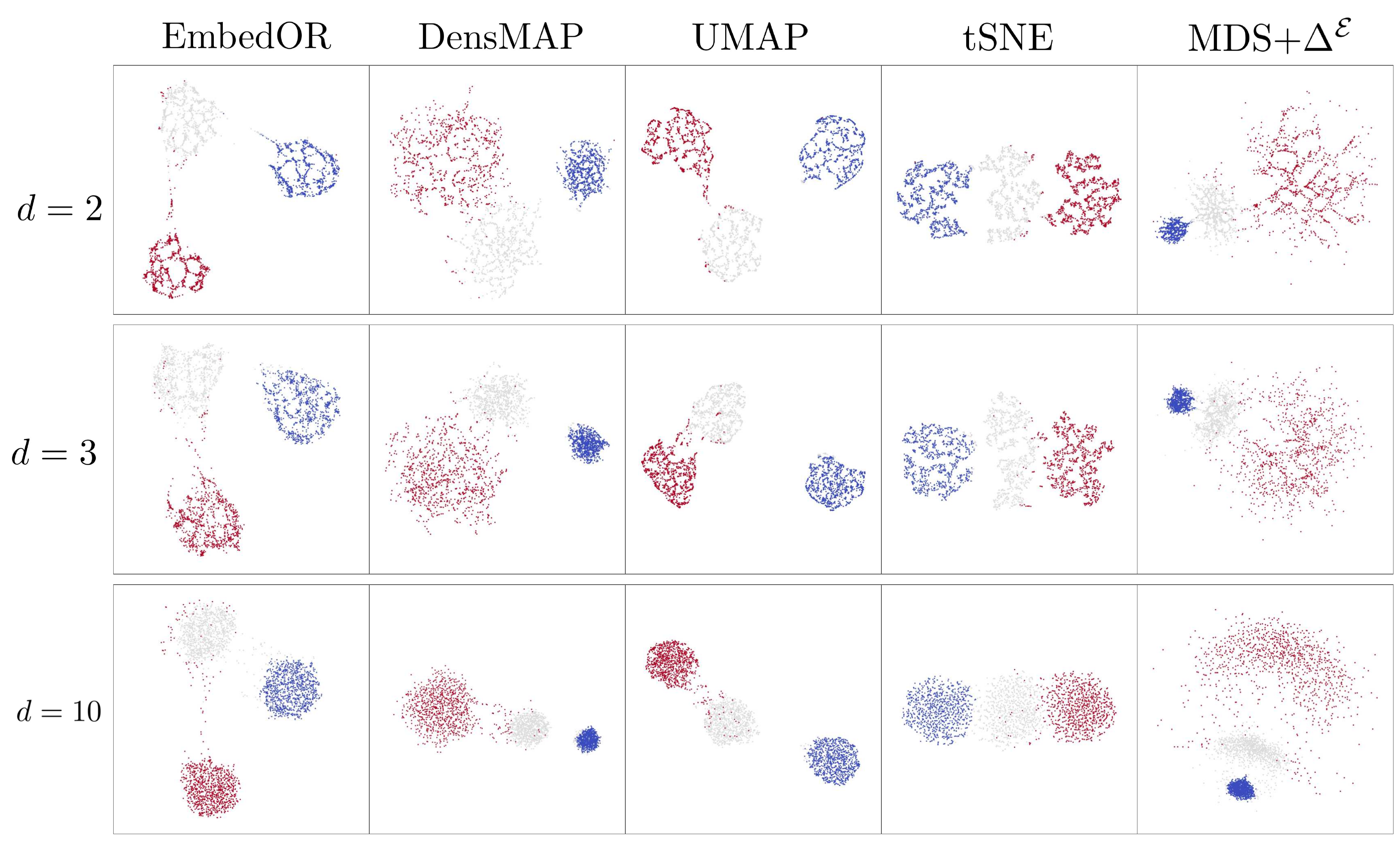}
    \caption{Embeddings of Gaussian mixture dataset for varying dimension. }\label{fig: density}
\end{figure}

In this section, we illustrate the effect of the presence of varying density on EmbedOR embeddings, UMAP, tSNE and densMAP. We note that, among these methods, densMAP is the only one that optimizes for preserving relative densities \cite{narayan2021assessing}. We also include metric MDS embeddings of the metric $\Delta^{\mathcal{E}}$, to show that \textit{unlike} EmbedOR itself, the EmbedOR metric does capture variations in scale. We evaluate on a synthetic dataset consisting of a mixture of 3 isotropic Gaussians, each with $\sigma = 0.1, 0.2$, and $0.5$, respectively. The mean of each Gaussian is positioned at the origin, $2/3$ of a unit along the first axis, and $2$ units along the first axis, respectively. We show the embeddings for varying dimension $d$ in \Cref{fig: density}.  

As one might expect, we see that EmbedOR, UMAP and tSNE all exhibit similar insensitivity to variations in density. This stems from the fact that all three methods employ a similar normalization of the metric, so that all points have a similar number of effective nearest neighbors in the embedding. DensMAP, on the other hand, preserves the variation in density of the original data; again, this is rather unsurprising, as the method is directly optimized for preserving relative scale. Finally, we find that Multidimensional Scaling (MDS) applied to the EmbedOR metric $\Delta^{\mathcal{E}}$ does in fact preserve the density of the original mixture components. This stems from the fact that the EmbedOR metric approximates geodesic distances up to a scaling factor as the ORC of each edge along the path approaches $+1$ (a condition we would expect to hold within the same mixture component as $N \rightarrow \infty$). We note that these results suggest that if the user intends to produce embeddings that preserve scale instead of the underlying connected components, a method like DensMAP should be used instead of EmbedOR. 

\subsection{EmbedOR Parameter Ablations} \label{sec: parameter ablations}

\begin{figure}[H]
    \centering
    \includegraphics[width=\linewidth]{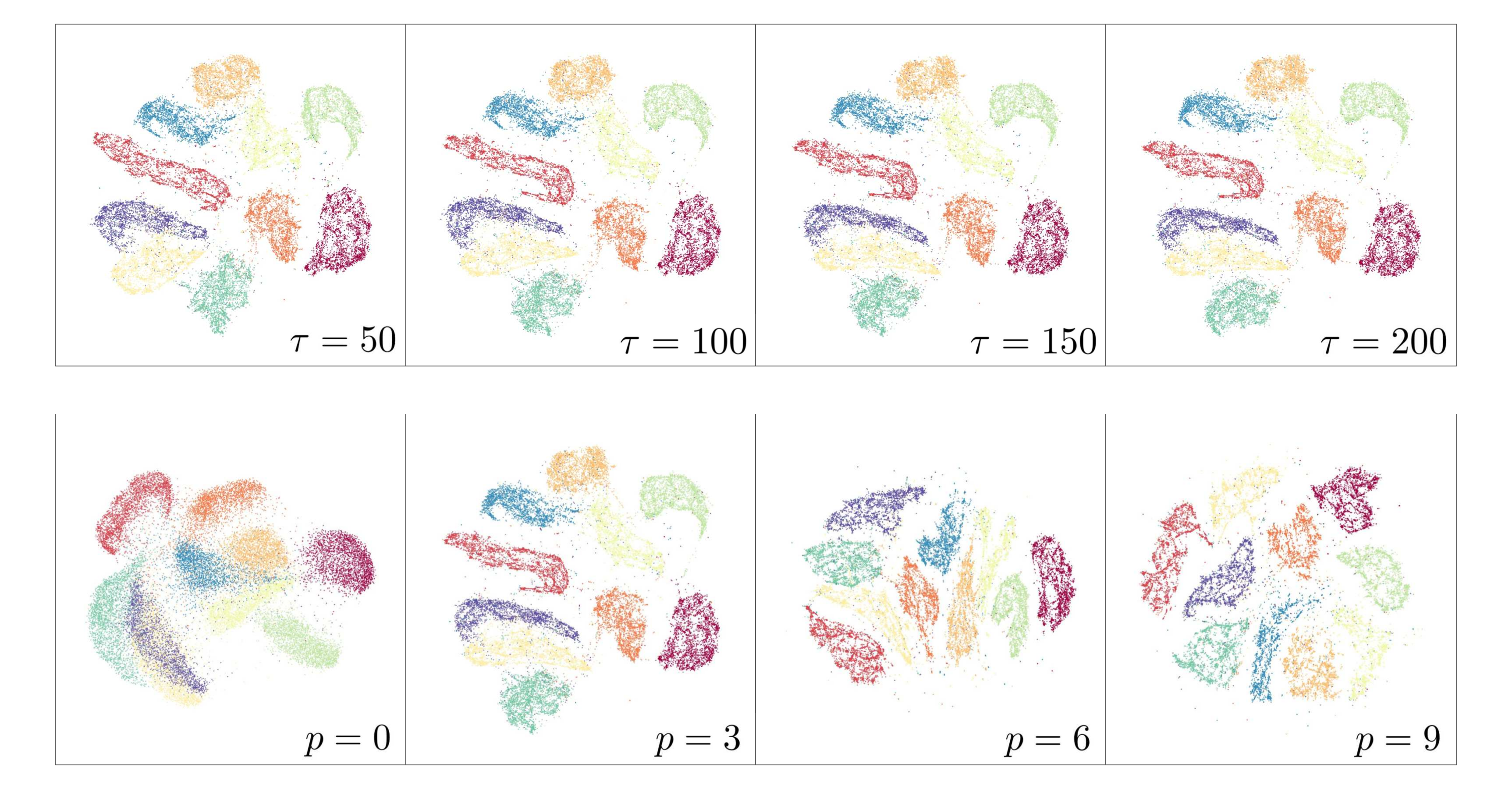}
    \caption{Parameter ablations for \Cref{alg: EmbedOR} applied to $25,000$ samples from the MNIST digits dataset.}\label{fig: parameter ablations}
\end{figure}

\Cref{fig: parameter ablations} displays embeddings produced by EmbedOR \cref{alg: EmbedOR} while varying parameters $p$ (repulsion strength) and $\tau$ (perplexity). Fortunately, we find that EmbedOR is extremely stable with respect to changes in the perplexity argument; no substantial changes are identifiable while ranging from $\tau =50$ to $\tau = 200$. As expected, for $p = 0$ we recover the embedding produced by EmbedOR with the Isomap metric, as shown in \Cref{fig: (f)mnist} of the main text. For larger values of $p$ we find that the embeddings are relatively stable, with the only changes arising from the presence of \say{clumping} appearing on very small scales.

\subsection{Baseline Parameter Ablations} \label{sec: baseline parameter ablations}

UMAP and tSNE are known to be very sensitive to hyperparameter choices; to ensure the significance of the results presented in the main manuscript, we perform hyperparameter ablations of UMAP and tSNE and repeat experiments that compare either of these methods with EmbedOR. 

\begin{table*}[h]
\footnotesize
\centering
 \begin{tabular}{l | >{\centering\arraybackslash}p{0.125\textwidth}  >{\centering\arraybackslash}p{0.125\textwidth}  >{\centering\arraybackslash}p{0.125\textwidth} >{\centering\arraybackslash}p{0.125\textwidth} >{\centering\arraybackslash}p{0.125\textwidth} >{\centering\arraybackslash}p{0.125\textwidth} >{\centering\arraybackslash}p{0.125\textwidth} >{\centering\arraybackslash}p{0.125\textwidth} } 
 \toprule & \multicolumn{6}{c}{min\_dist} \\
  \toprule  & $0.01$ &  $0.05$ & $0.1$ & $0.2$ & $0.5$ & $1.0$  \\ 
\midrule 
 \hline
 Circles  & $0.475 \pm 0.026$ & $0.474 \pm 0.025$ & $0.459 \pm 0.024$ & $0.443 \pm 0.031$ & $0.398 \pm 0.017$ & $0.382 \pm 0.005$  \\ 
 Swiss Roll & $0.744 \pm 0.109$ & $0.744 \pm 0.117$ & $0.750 \pm 0.121$ & $0.740 \pm 0.124$ & $0.690 \pm 0.142$ & $0.595 \pm 0.133$  \\ 
 Tori  & $0.682 \pm 0.031$ & $0.676 \pm 0.036$ & $0.654 \pm 0.033$ & $0.618 \pm 0.047$ & $0.529 \pm 0.018$ & $0.478 \pm 0.010$ \\ 
 Tree  & $0.802 \pm 0.058$ & $0.803 \pm 0.055$ & $0.721 \pm 0.079$ & $0.673 \pm 0.113$ & $0.676 \pm 0.106$ & $0.659 \pm 0.153$  \\ 
 \bottomrule
 \end{tabular} \\
\vspace{5mm}
 \footnotesize
 \begin{tabular}{l | >{\centering\arraybackslash}p{0.125\textwidth}  >{\centering\arraybackslash}p{0.125\textwidth}  >{\centering\arraybackslash}p{0.125\textwidth} >{\centering\arraybackslash}p{0.125\textwidth} >{\centering\arraybackslash}p{0.125\textwidth} >{\centering\arraybackslash}p{0.125\textwidth} >{\centering\arraybackslash}p{0.125\textwidth} >{\centering\arraybackslash}p{0.125\textwidth} >{\centering\arraybackslash}p{0.125\textwidth}  } 
 \toprule & \multicolumn{6}{c}{negative\_sample\_rate} \\
  \toprule  & $1$ &  $2$ & $3$ & $4$ & $5$ & $10$ \\ 
\midrule 
 \hline
 Circles  & $0.455 \pm 0.036$ & $0.464 \pm 0.037$ & $0.465 \pm 0.037$ & $0.464 \pm 0.03$ & $0.461 \pm 0.030$ & $0.461 \pm 0.037$  \\ 
 Swiss Roll  & $0.838\pm 0.130$ & $0.823 \pm 0.113$ & $0.790 \pm 0.113$ & $0.761 \pm 0.123$ & $0.764 \pm 0.117$ & $0.694 \pm 0.119$  \\ 
 Tori  & $0.637 \pm 0.017$ & $0.645 \pm 0.031$ & $0.647 \pm 0.034$ & $0.648 \pm 0.032$ & $0.647 \pm 0.032$ & $0.642 \pm 0.045$  \\ 
 Tree  & $0.921 \pm 0.009$ & $0.870 \pm 0.021$ & $0.836 \pm 0.042$ & $0.729 \pm 0.067$ & $0.760 \pm 0.056$ & $0.679 \pm 0.114$ \\ 
 \bottomrule
 \end{tabular}
 \caption{Spearman correlation coefficient between UMAP embedding distances and estimated geodesic distance for the synthetic manifolds over $10$ random samples of the dataset. Note that the embeddings were computed on noisy data, while the ground-truth geodesic distances were estimated on the noiseless data. This table provides a parameter ablation of the experiments from \Cref{table: geodesic distances} for UMAP hyperparameters min\_dist and negative\_sample\_rate. For datasets with more than 1 connected component, ground-truth distances between points in differing connected components were set to a large constant value.}
 \label{table: geodesic distances umap ablation}
\end{table*}

\begin{table*}[h]
\scriptsize
\centering
 \begin{tabular}{l | >{\centering\arraybackslash}p{0.1\textwidth}  >{\centering\arraybackslash}p{0.1\textwidth}  >{\centering\arraybackslash}p{0.1\textwidth} >{\centering\arraybackslash}p{0.1\textwidth} >{\centering\arraybackslash}p{0.1\textwidth} >{\centering\arraybackslash}p{0.1\textwidth} >{\centering\arraybackslash}p{0.1\textwidth} >{\centering\arraybackslash}p{0.1\textwidth} >{\centering\arraybackslash}p{0.1\textwidth} } 
 \toprule & \multicolumn{7}{c}{perplexity} \\
  \toprule  & $5$ &  $10$ & $15$ & $30$ & $40$ & $50$ & $100$ \\ 
\midrule 
 \hline
 Circles  & $0.397 \pm 0.001$ & $0.405 \pm 0.002$ & $0.415 \pm 0.003$ & $0.429 \pm 0.005$ & $0.436 \pm 0.004$ & $0.443 \pm 0.006$ & $0.466 \pm 0.014$   \\ 
 Swiss Roll & $0.419 \pm 0.011$ & $0.495 \pm 0.016$ & $0.576 \pm 0.016$ & $0.702 \pm 0.010$ & $0.741 \pm 0.019$ & $0.777 \pm 0.014$ & $0.518 \pm 0.106$  \\ 
 Tori  & $0.432 \pm 0.006$ & $0.472 \pm 0.018$ & $0.504 \pm 0.022$ & $0.617 \pm 0.077$ & $0.612 \pm 0.068$ & $0.573 \pm 0.027$ & $0.592 \pm 0.013$   \\ 
 Tree & $0.684 \pm 0.027$ & $0.682 \pm 0.019$ & $0.701 \pm 0.022$ & $0.740 \pm 0.013$ & $0.758 \pm 0.013$ & $0.771 \pm 0.012$ & $0.839 \pm 0.005$  \\ 
 \bottomrule
 \end{tabular} \\
\vspace{5mm}
 \footnotesize
 \begin{tabular}{l | >{\centering\arraybackslash}p{0.125\textwidth}  >{\centering\arraybackslash}p{0.125\textwidth}  >{\centering\arraybackslash}p{0.125\textwidth} >{\centering\arraybackslash}p{0.125\textwidth} >{\centering\arraybackslash}p{0.125\textwidth} >{\centering\arraybackslash}p{0.125\textwidth} >{\centering\arraybackslash}p{0.125\textwidth} >{\centering\arraybackslash}p{0.125\textwidth} } 
 \toprule & \multicolumn{6}{c}{early\_exaggeration} \\
  \toprule  & $2$ &  $4$ & $8$ & $12$ & $16$ & $24$  \\ 
\midrule 
 \hline
 Circles  & $0.435 \pm 0.004$ & $0.432 \pm 0.004$ & $0.429 \pm 0.004$ & $0.429 \pm 0.005$ & $0.428 \pm 0.004$ & $0.420 \pm 0.004$  \\ 
 Swiss Roll & $0.510 \pm 0.036$ & $0.631 \pm 0.037$ & $0.706 \pm 0.012$ & $0.702 \pm 0.010$ & $0.697 \pm 0.010$ & $0.673 \pm 0.017$   \\ 
 Tori  & $0.543 \pm 0.026$ & $0.553 \pm 0.016$ & $0.637 \pm 0.075$ & $0.617 \pm 0.077$ & $0.537 \pm 0.025$ & $0.512 \pm 0.019$   \\ 
 Tree & $0.761 \pm 0.012$ & $0.740 \pm 0.017$ & $0.743 \pm 0.015$ & $0.740 \pm 0.013$ & $0.732 \pm 0.015$ & $0.730 \pm 0.010$ \\ 
 \bottomrule
 \end{tabular}
 \caption{Spearman correlation coefficient between tSNE embedding distances and estimated geodesic distance for the synthetic manifolds over $10$ random samples of the dataset. Note that the embeddings were computed on noisy data, while the ground-truth geodesic distances were estimated on the noiseless data. This table provides a parameter ablation of the experiments from \Cref{table: geodesic distances} for tSNE hyperparameters perplexity and early\_exaggeration. For datasets with more than 1 connected component, ground-truth distances between points in differing connected components were set to a large constant value.}
 \label{table: geodesic distances tsne ablation}
\end{table*}

\begin{figure}[H]
    \centering
    \includegraphics[width=\linewidth]{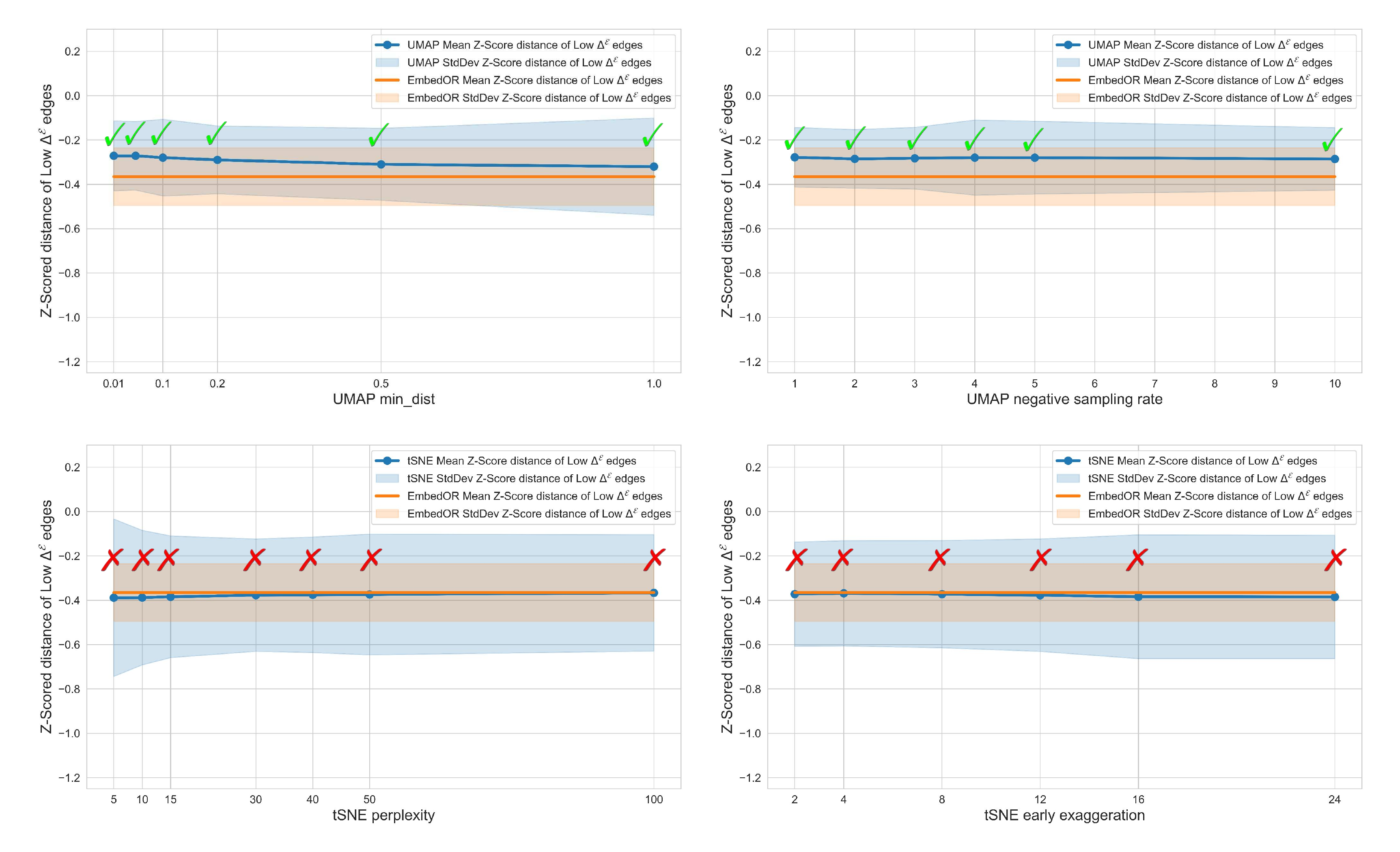}
    \caption{Parameter ablation for the MNIST entry of \Cref{table: low distance edge distortions}. Note that a checkmark (\Checkmark) denotes a rejection of the null hypothesis that $\mu_{0.33}^{\text{EmbedOR}} = \mu_{0.33}^{\text{ALG}}$ in favor of the alternate hypothesis $\mu_{0.33}^{\text{EmbedOR}} < \mu_{0.33}^{\text{ALG}}$ (under the statistical testing framework described by \Cref{table: statistical tests} and \cref{eq: statistical test distribution mean}), while a (\XSolidBrush) denotes a failure to reject the null.}
    \label{fig: umap tsne mnist ablation}
\end{figure}

\begin{figure}[H]
    \centering
    \includegraphics[width=\linewidth]{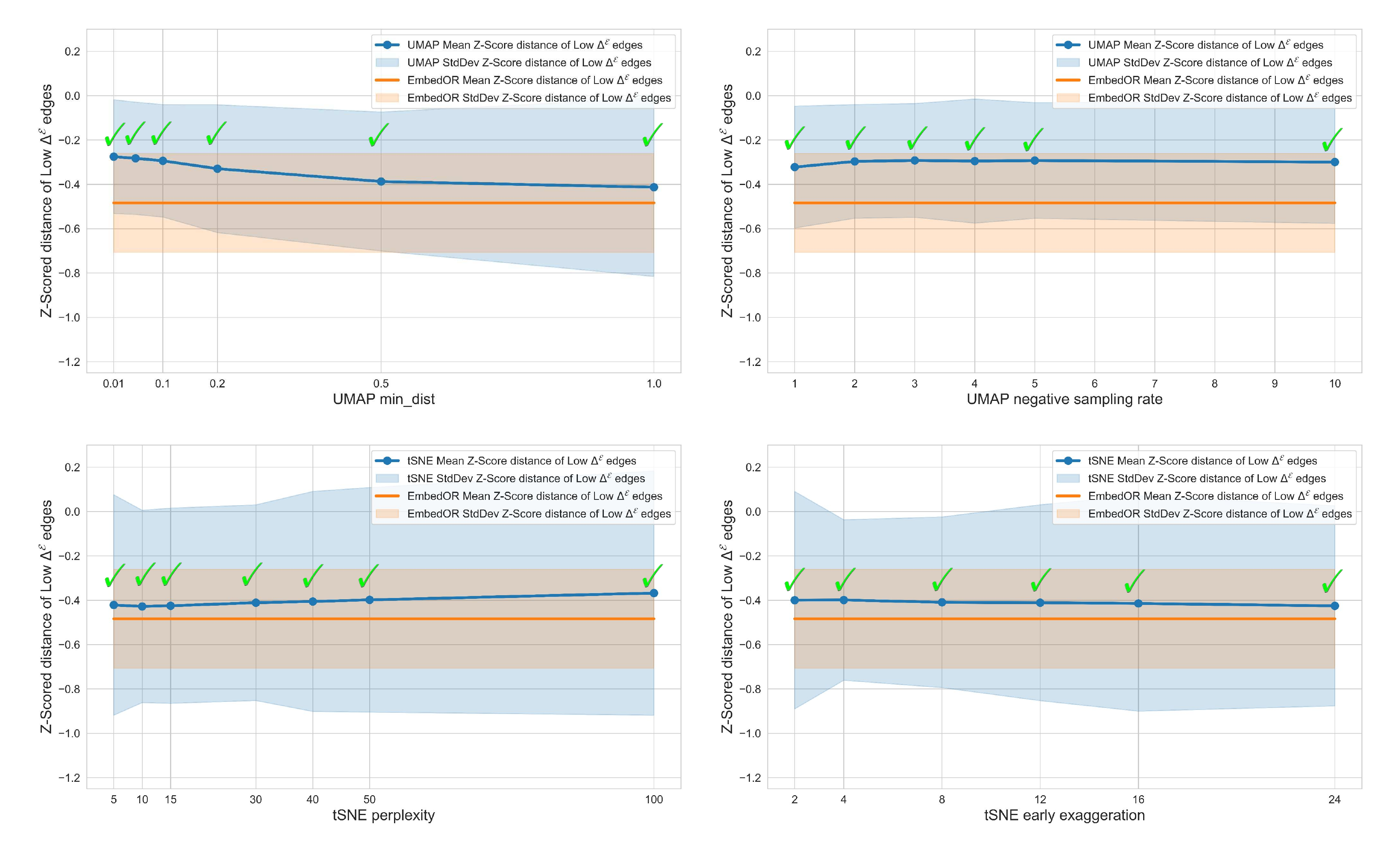}
    \caption{Parameter ablation for the fMNIST entry of \Cref{table: low distance edge distortions}. Note that a checkmark (\Checkmark) denotes a rejection of the null hypothesis that $\mu_{0.33}^{\text{EmbedOR}} = \mu_{0.33}^{\text{ALG}}$ in favor of the alternate hypothesis $\mu_{0.33}^{\text{EmbedOR}} < \mu_{0.33}^{\text{ALG}}$ (under the statistical testing framework described by \Cref{table: statistical tests} and \cref{eq: statistical test distribution mean}), while a (\XSolidBrush) denotes a failure to reject the null. } 
    \label{fig: umap tsne fmnist ablation}
\end{figure}

\begin{figure}[H]
    \centering
    \includegraphics[width=\linewidth]{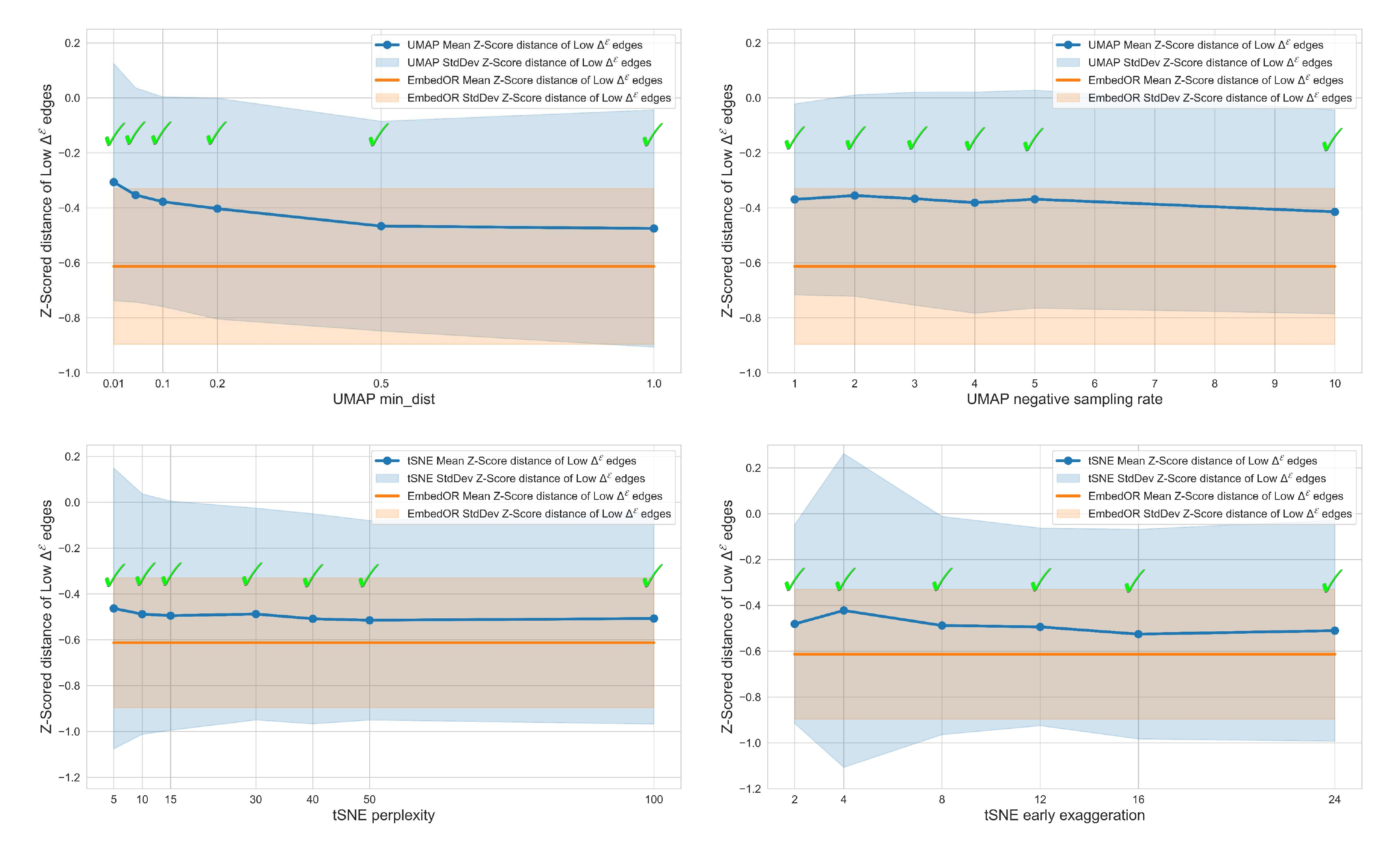}
    \caption{Parameter ablation for the developmental trajectories entry of \Cref{table: low distance edge distortions}. Note that a checkmark (\Checkmark) denotes a rejection of the null hypothesis that $\mu_{0.33}^{\text{EmbedOR}} = \mu_{0.33}^{\text{ALG}}$ in favor of the alternate hypothesis $\mu_{0.33}^{\text{EmbedOR}} < \mu_{0.33}^{\text{ALG}}$ (under the statistical testing framework described by \Cref{table: statistical tests} and \cref{eq: statistical test distribution mean}), while a (\XSolidBrush) denotes a failure to reject the null.}
    \label{fig: umap tsne developmental ablation}
\end{figure}

\begin{figure}[H]
    \centering
    \includegraphics[width=\linewidth]{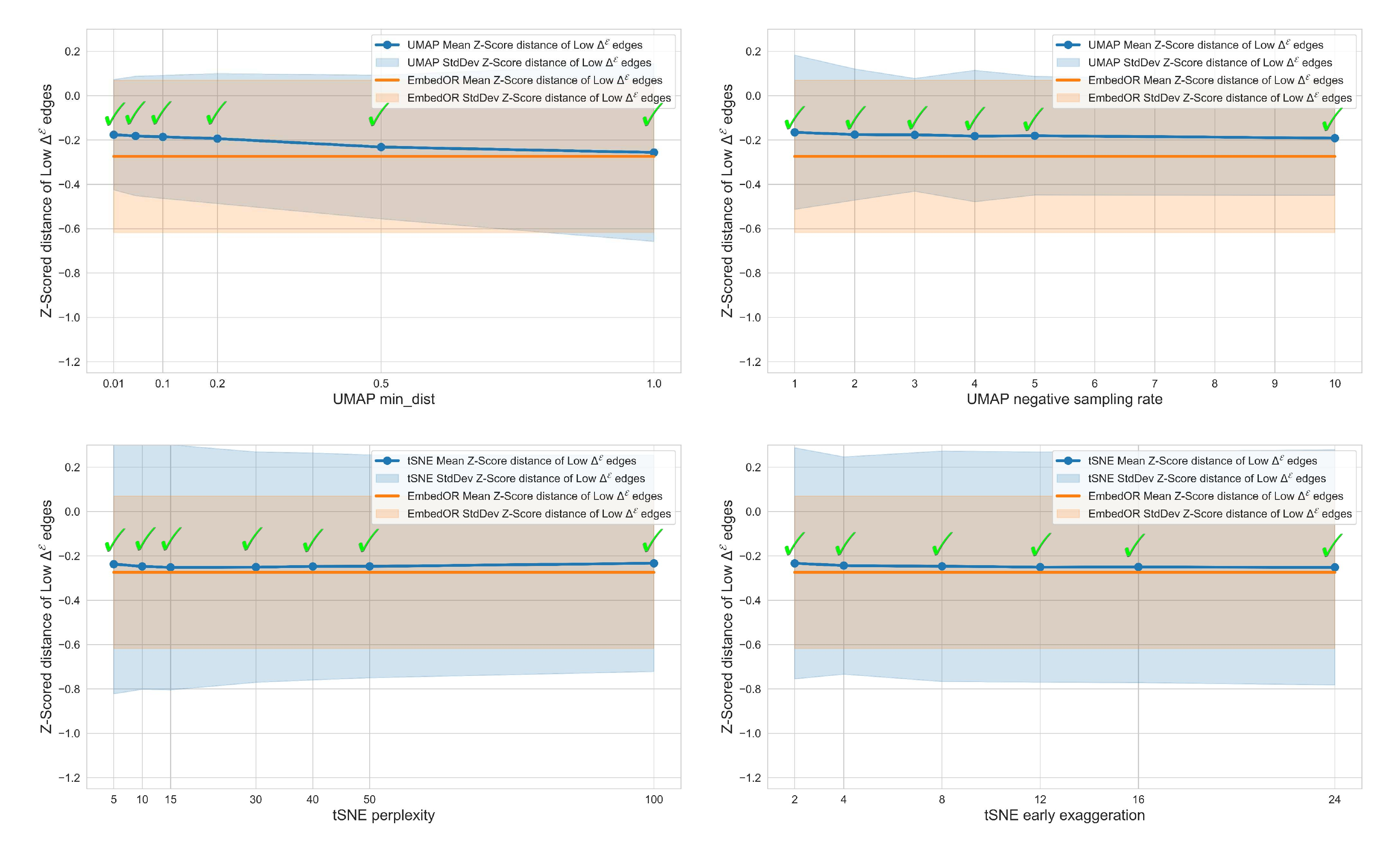}
    \caption{Parameter ablation for the retinal cells entry of \Cref{table: low distance edge distortions}. Note that a checkmark (\Checkmark) denotes a rejection of the null hypothesis that $\mu_{0.33}^{\text{EmbedOR}} = \mu_{0.33}^{\text{ALG}}$ in favor of the alternate hypothesis $\mu_{0.33}^{\text{EmbedOR}} < \mu_{0.33}^{\text{ALG}}$ (under the statistical testing framework described by \Cref{table: statistical tests} and \cref{eq: statistical test distribution mean}), while a (\XSolidBrush) denotes a failure to reject the null. }
    \label{fig: umap tsne macosko ablation}
\end{figure}

\begin{figure}[H]
    \centering
    \includegraphics[width=\linewidth]{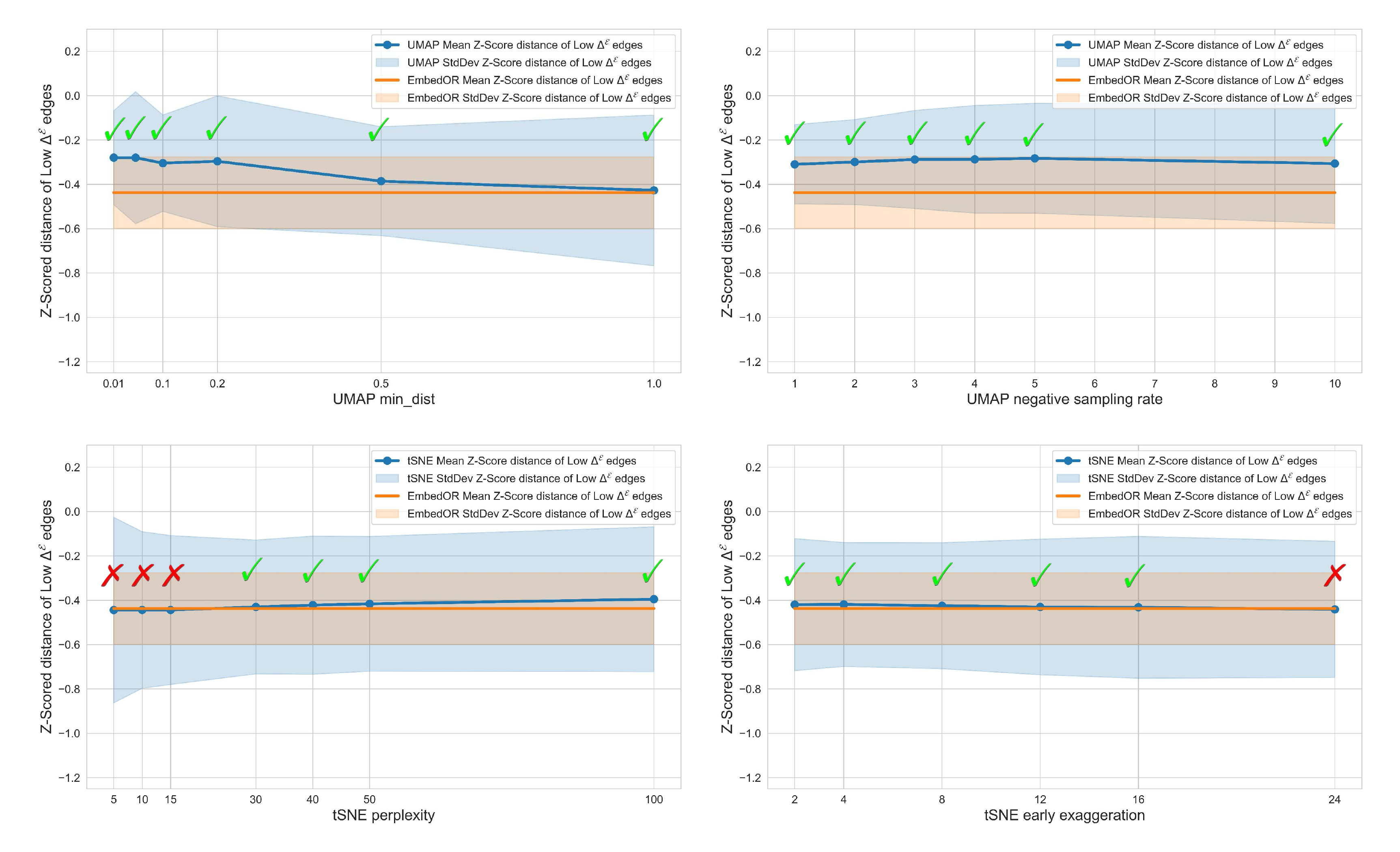}
    \caption{Parameter ablation for the chimp brain organoids entry of \Cref{table: low distance edge distortions}. Note that a checkmark (\Checkmark) denotes a rejection of the null hypothesis that $\mu_{0.33}^{\text{EmbedOR}} = \mu_{0.33}^{\text{ALG}}$ in favor of the alternate hypothesis $\mu_{0.33}^{\text{EmbedOR}} < \mu_{0.33}^{\text{ALG}}$ (under the statistical testing framework described by \Cref{table: statistical tests} and \cref{eq: statistical test distribution mean}), while a (\XSolidBrush) denotes a failure to reject the null.}
    \label{fig: umap tsne chimp ablation}
\end{figure}

\begin{figure}[H]
    \centering
    \includegraphics[width=\linewidth]{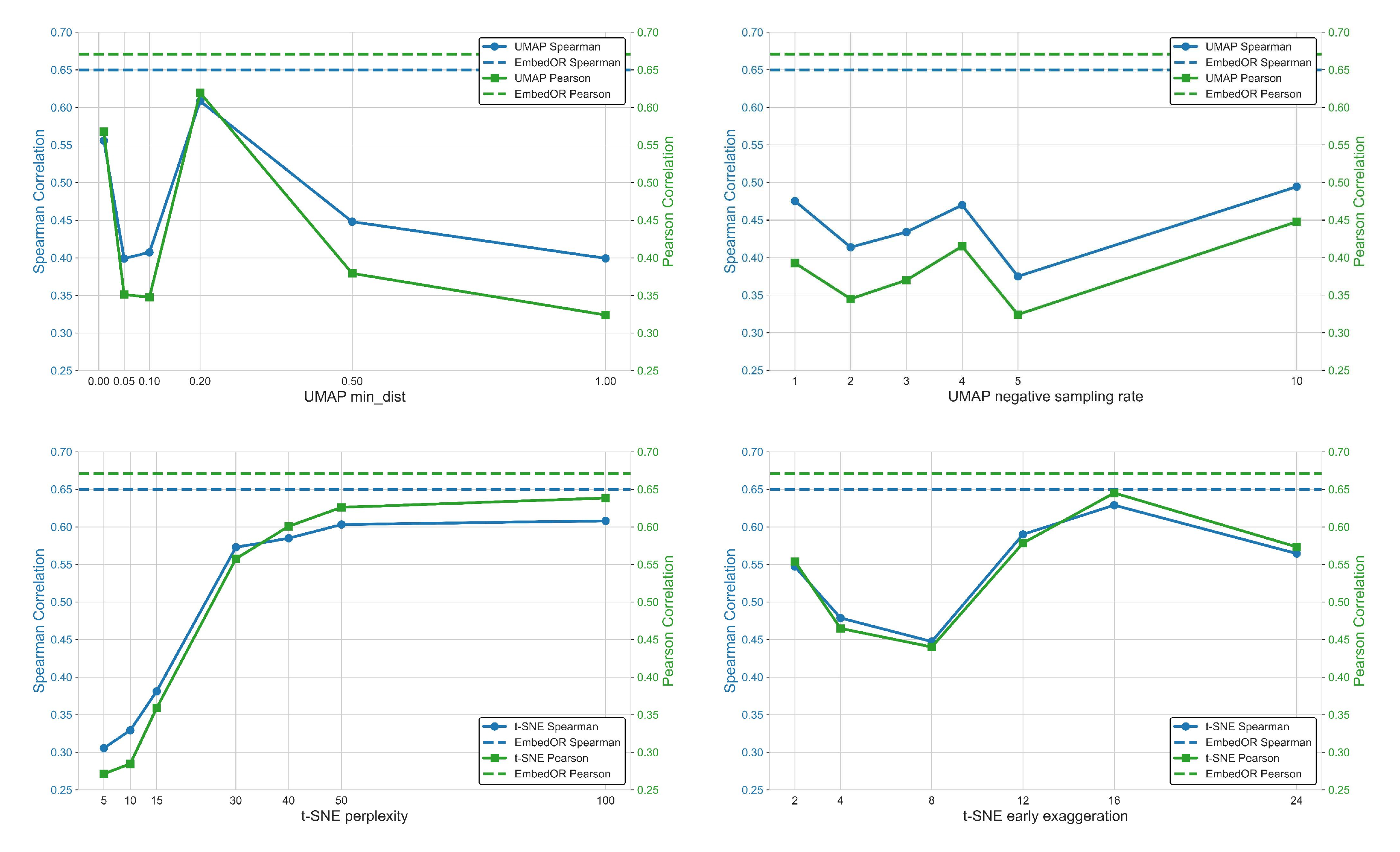}
    \caption{Parameter ablation for experiment that computes the Spearman and Pearson correlation coefficients between embedded distances and temporal distances for the iPSC developmental trajectories dataset (\Cref{table: temporal correlation})}
    \label{fig: umap tsne developmental temporal ablation}
\end{figure}

\subsection{Noise model}\label{sec: noise model experiments}

\begin{figure}[H]
    \centering
    \includegraphics[width=\linewidth]{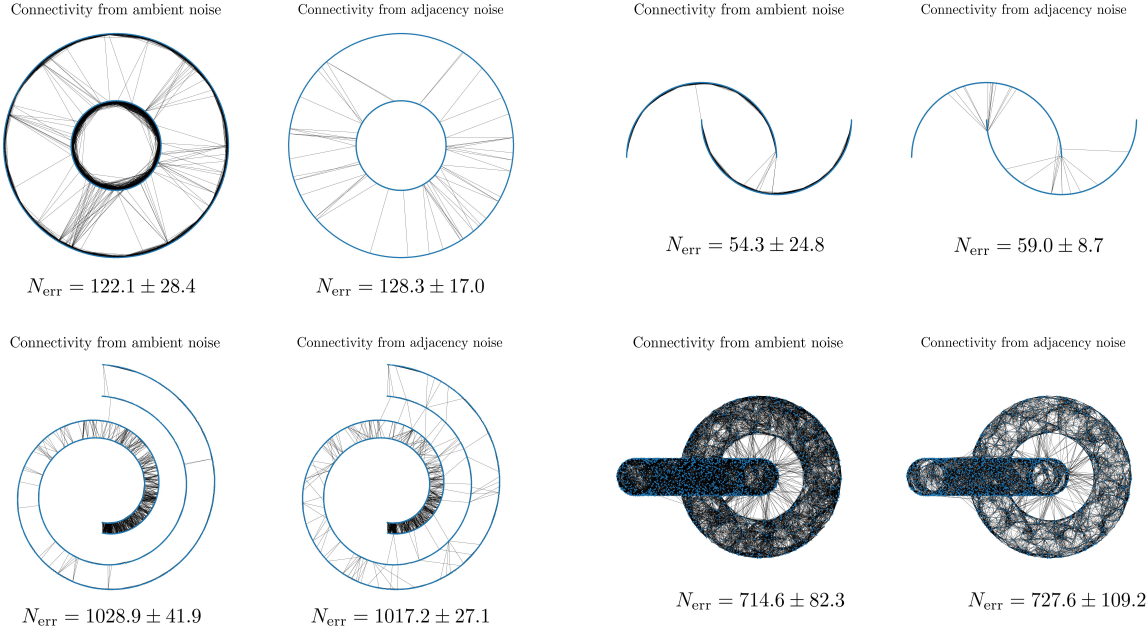}
    \caption{Connectivity of nearest neighbor graphs built from data perturbed with ambient noise, and with the adjacency noise model described in \Cref{sec: noise model}. Graphs are visualized with noiseless point clouds for ease of comparison. The quantity $N_{\text{err}}$ denotes the number of edges in the visualization where the underlying connected component of one endpoint does not match that of the other. Furthermore, for all experiments shown, the model parameters for the adjacency-based model were chosen such that they do \textit{not} violate the assumptions of \Cref{thm: dissimilar pairs} ($s_{\max} < 3/8e$).} 
    \label{fig: noise model}
\end{figure}

\Cref{fig: noise model} compares nearest neighbor graphs generated
from point clouds with ambient perturbations (additive isotropic Gaussian noise) and graphs produced by the noise model described in \Cref{sec: noise model}.
For ease of comparison, we project the perturbed points back onto the manifold for visualization.
Empirically, we see no systematic difference in the connectivity structure between differing underlying connected components of the generated graphs. Furthermore, we observe that the average number of cluster-bridging edges averaged across 50 trials is extremely similar between both models. While we consider this to be evidence that our proposed noise model adheres well to the standard ambient perturbation model, we leave an extensive theoretical analysis of this claim to future work.

\begin{figure}[H]
    \centering
    \includegraphics[width=\linewidth]{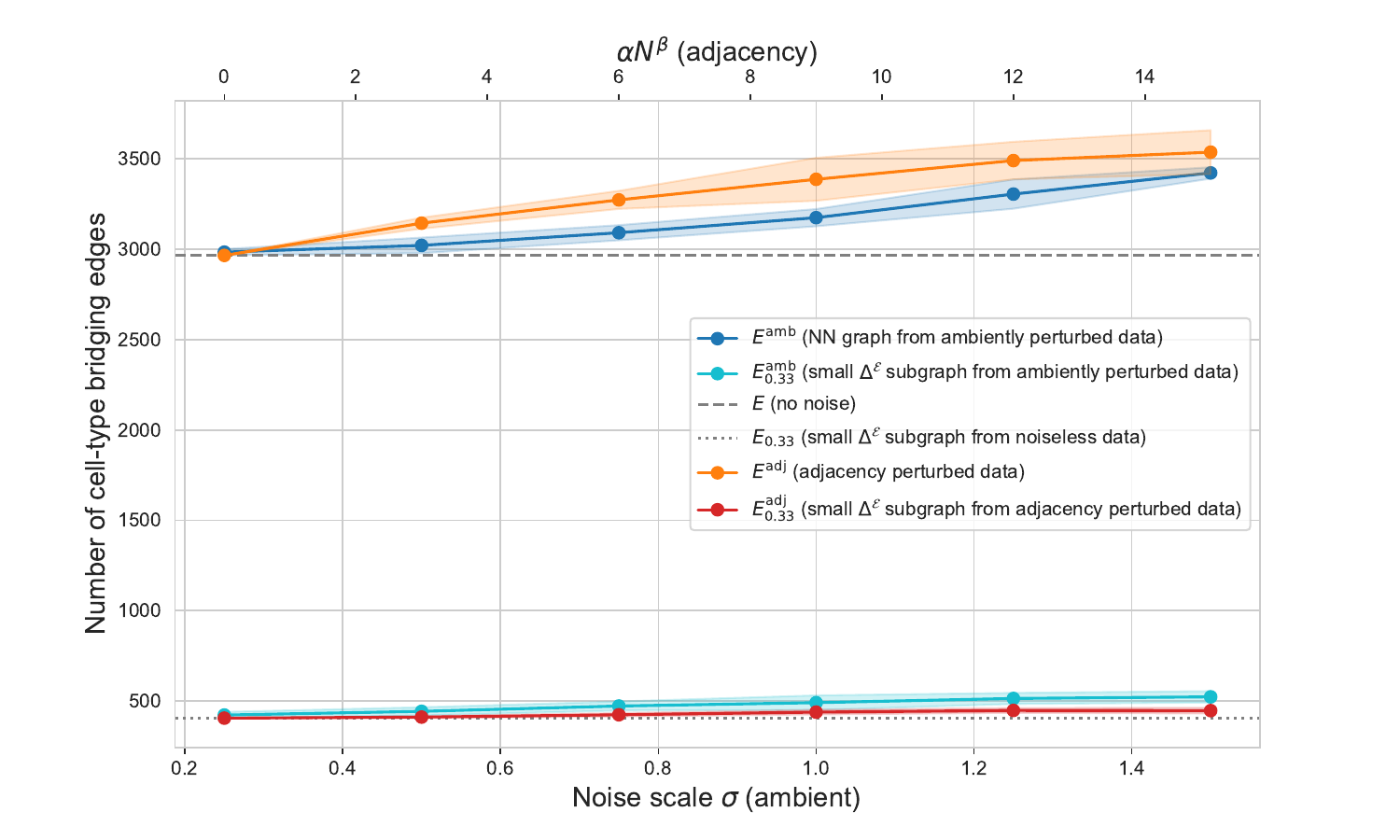}
    \caption{Plots of the mean$\pm$std-dev number of cell-type bridging edges in nearest neighbor graphs built from perturbed scRNAseq data as a function of noise intensity, under two different models of noise. The first model (denoted \textit{ambient}) perturbs the data by adding isotropic Gaussian noise. The second model (denoted \textit{adjacent}) directly perturbs the adjacency matrix of the nearest neighbor graph, as described in \Cref{sec: noise model}.}
    \label{fig: pbmc noise model}
\end{figure}

\begin{figure}[H]
    \centering
    \includegraphics[width=\linewidth]{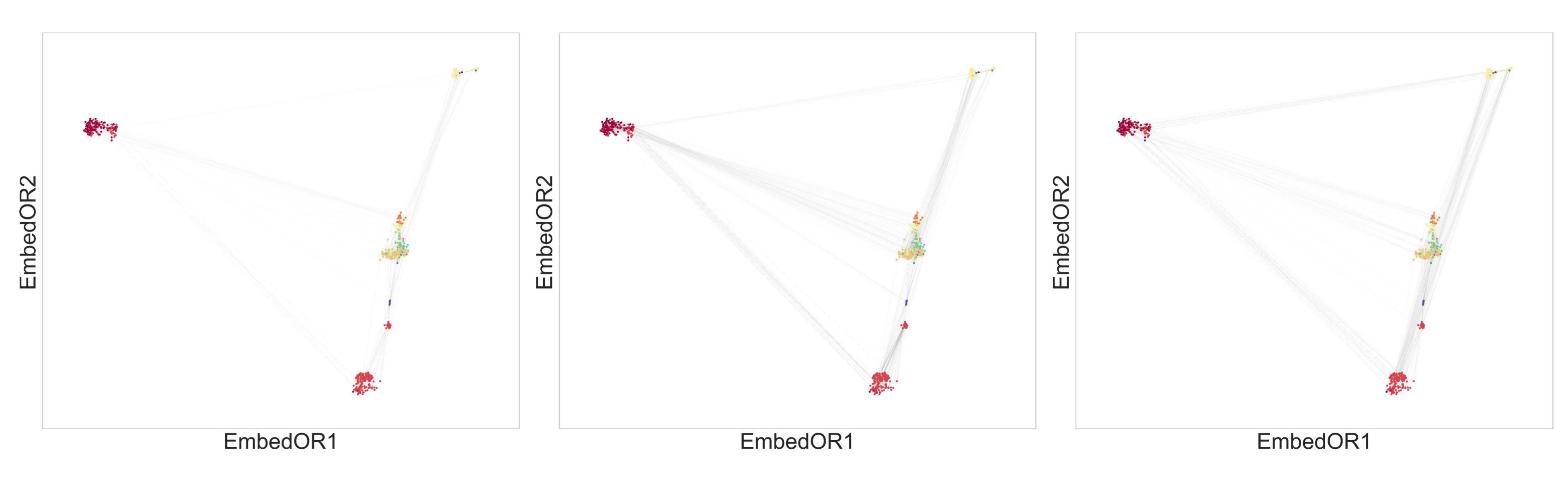}
    \caption{Visualization of nearest neighbor graph of unperturbed peripheral blood mononuclear cell data (left), nearest neighbor graph of the data perturbed with additive Gaussian perturbations (middle), and adjacency perturbations (right) under the model described in \Cref{sec: noise model}.}
    \label{fig: pbmc noise visualization}
\end{figure}

\begin{figure}[H]
    \centering
    \includegraphics[width=\linewidth]{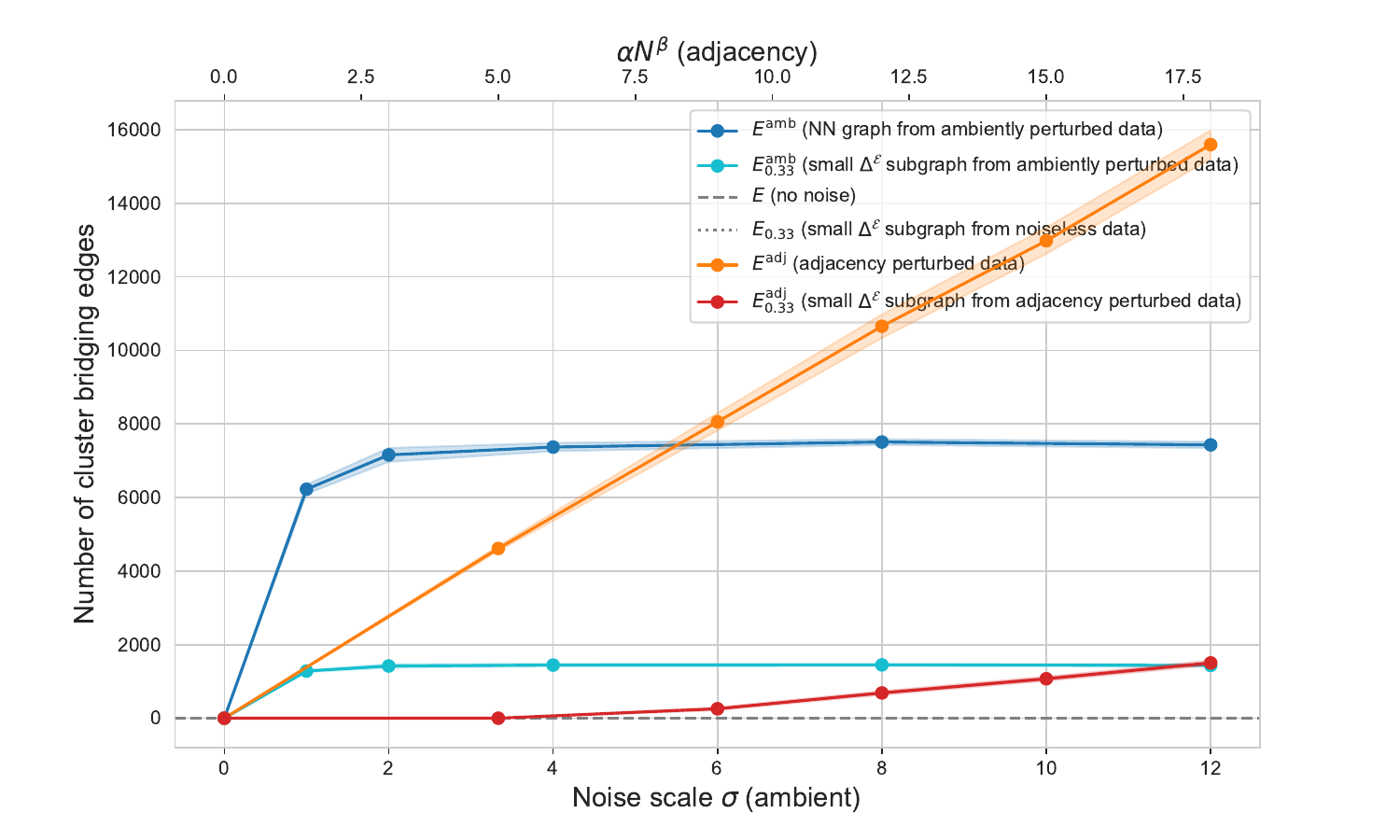}
    \caption{Plots of the mean$\pm$std-dev number of cell-type bridging edges in nearest neighbor graphs built from perturbed moons dataset as a function of noise intensity, under two different models of noise. The first model (denoted \textit{ambient}) perturbs the data by adding isotropic Gaussian noise. The second model (denoted \textit{adjacent}) directly perturbs the adjacency matrix of the nearest neighbor graph, as described in \Cref{sec: noise model}.}
    \label{fig: circles noise model}
\end{figure}

We also provide an experiment that demonstrates the similarity between the noise models when used to corrupt real data in \Cref{fig: pbmc noise model}. In particular, we take a dataset of peripheral blood mononuclear cells from 10XGenomics and we perturb it with (1) ambient perturbations and (2) the adjacency noise described in \Cref{sec: noise model}. We then plot the number of cell-type bridging edges as a function of the noise intensity for each model in \Cref{fig: pbmc noise model}, and we visualize perturbed graphs in \Cref{fig: pbmc noise visualization}. In \Cref{fig: pbmc noise model} we observe a similar trend in the number of cell-type bridging edges among nearest neighbor graphs built from both perturbations. We also plot the number of cell-type bridging edges in the subgraph of the $k$-NN graph with edges that have a small distance with respect to $\Delta^{\mathcal{E}}$. Again, we see a similar trend in the number of cell-type bridging edges; with that said, we do find that EmbedOR performs slightly better under our adjacency perturbation model than with ambient perturbations. This suggests that the correlations in noisy added edges that are not entirely captured by our model may have a non-negligible effect. This is also supported by \Cref{fig: circles noise model}, where we repeated the same experiment on the synthetic moons dataset. There, we find that the EmbedOR metric $\Delta^{\mathcal{E}}$ is again slightly more robust to adjacency perturbations than ambient perturbations. 

\section{Derivations}

\subsection{The objective function}\label{sec: objective function derivation}
We formulate our optimization problem as follows,
\begin{align*}
    Y^* &= \arg \min_{Y} \, \sum_{j < i}\biggl[p_{ij}\log\biggl(\frac{p_{ij}}{f_{ij}}\biggr) + \lambda(1-p_{ij})\log\biggl(\frac{1-p_{ij}}{1-f_{ij}}\biggr) \biggr]\,,
\end{align*}
where $f_{ij} = (1 + \norm{y_i - y_j}^2)^{-1}$ and $y_i, y_j$ are the embedded coordinates of points $x_i, x_j$, respectively.
We choose 
\[\lambda = \frac{1}{N^2}\]
for our implementation. This approximately balances the coefficients in the objective so that neither dominates the objective function. It empirically produces the best results on synthetic data, and also happens to be on the order of the inverse of the number of interaction pairs. Now expanding the logarithms, we get
\begin{equation}
        Y^* = \arg \min_{Y} \, \sum_{j < i} \Biggl( \Bigg[p_{ij}\log(p_{ij}) + \frac{1}{N^2}(1-p_{ij})\log(1-p_{ij})\Bigg] - \Bigg[p_{ij}\log(f_{ij}) + \frac{1}{N^2}(1-p_{ij})\log(1-f_{ij})\Bigg] \Biggr).
\end{equation}
The first term in the parenthesis has no dependence on $Y$, so we can drop it,
\begin{align*}
    Y^* &= \arg \min_{Y} \, - \sum_{j < i} \biggl( p_{ij}\log(f_{ij}) + \frac{1}{N^2}(1-p_{ij})\log(1-f_{ij}) \biggr) \\
    &= \arg \max_{Y} \sum_{j < i} p_{ij}\log(f_{ij}) + \frac{1}{N^2}\sum_{j < i} (1-p_{ij})\log(1-f_{ij}) \\
    \begin{split}
        &= \arg \max_{Y} \Biggl(\sum_{j' < i'}p_{i'j'}\Biggr)\sum_{j < i} \underbrace{\frac{p_{ij}}{\sum_{j' < i'}p_{i'j'}}}_{\overset{\text{def}}{=} p^+_{ij}}\log(f_{ij}) \\
        &\hspace{30mm}+ \Biggl(\sum_{j' < i'}(1-p_{i'j'})\Biggr)\Bigg(\frac{1}{N^2}\Biggr) \sum_{j < i} \underbrace{\frac{(1-p_{ij})}{\sum_{j' < i'}(1-p_{i'j'})}}_{\overset{\text{def}}{=} p^-_{ij}}\log(1-f_{ij}).
    \end{split}
\end{align*}
Observe that $p^+_{ij}$ and $p^-_{ij}$ define valid probability measures over all pairwise interactions. Thus, we can rewrite the objective as follows,
\begin{align*}
    Y^* &= \arg \max_{Y} \Biggl(\sum_{j' < i'}p_{i'j'}\Biggr)\sum_{j < i} p^+_{ij}\log(f_{ij}) +\bigg(\sum_{j' < i'}(1-p_{i'j'})\bigg)\Bigg(\frac{1}{N^2}\Biggr) \sum_{j < i} p^-_{ij}\log(1-f_{ij}) \\
    &= \arg \max_{Y} \, \Biggl(\sum_{j' < i'}p_{i'j'}\Biggr) \mathbb{E}_{(i,j) \sim p^+} \Bigl[\log(f_{ij})\Bigr]  + \bigg(\sum_{j' < i'}(1-p_{i'j'})\bigg)\Bigg(\frac{1}{N^2}\Biggr) \mathbb{E}_{(i,j) \sim p^-}\Bigl[\log(1-f_{ij})\Bigr] \\
    &= \arg \max_{Y} \,  \mathbb{E}_{(i,j) \sim p^+}\Bigl[\log(f_{ij})\Bigr] + \Bigg(\frac{\sum_{j' < i'}(1-p_{i'j'})}{\sum_{j' < i'}p_{i'j'}}\Bigg)\Bigg(\frac{1}{N^2}\Biggr) \mathbb{E}_{(i,j) \sim p^-}\Bigl[\log(1-f_{ij})\Bigr].
\end{align*}
Defining $Z \overset{\text{def}}{=} \sum_{j < i}p_{ij}$ we can say
\begin{align*}
    \arg \max_{Y} \,  \underbrace{\mathbb{E}_{(i,j) \sim p^+}\Bigl[\log(f_{ij})\Bigr]}_{\mathclap{\text{attractive term}}} + \Bigg(\frac{\frac{N^2-N}{2}-Z}{Z}\Bigg)\Bigg(\frac{1}{N^2}\Biggr) \underbrace{\mathbb{E}_{(i,j) \sim p^-}\Bigl[\log(1-f_{ij})\Bigr]}_{\mathclap{\text{repulsive term}}}.
\end{align*}
\subsection{The gradient of the objective}
Let 
\[\tilde{\mathcal{L}}(Y) = \mathbb{E}_{(i,j) \sim p^+}\Bigl[\log(f_{ij})\Bigr] + \Bigg(\frac{\frac{N^2-N}{2}-Z}{Z}\Bigg)\Bigg(\frac{1}{N^2}\Biggr) \mathbb{E}_{(i,j) \sim p^-}\Bigl[\log(1-f_{ij})\Bigr] \]
where the $f_{ij}$ terms are functions of $Y$. Taking the gradient with respect to one of the low-dimensional points $y_m$, we obtain
\begin{align*}
    \nabla_{y_m} \tilde{\mathcal{L}}(Y) &= \nabla_{y_m} \biggl(\mathbb{E}_{(i,j) \sim p^+}\Bigl[\log(f_{ij})\Bigr] +\Bigg(\frac{\frac{N^2-N}{2}-Z}{Z}\Bigg)\Bigg(\frac{1}{N^2}\Biggr) \mathbb{E}_{(i,j) \sim p^-}\Bigl[\log(1-f_{ij})\Bigr]\biggr) \\
    &=\mathbb{E}_{(i,j) \sim p^+} \Bigl[\nabla_{y_m}\log(f_{ij})\Big] + \Bigg(\frac{\frac{N^2-N}{2}-Z}{Z}\Bigg)\Bigg(\frac{1}{N^2}\Biggr) \mathbb{E}_{(i,j) \sim p^-}\Big[\nabla_{y_m}\log(1-f_{ij})\Big].
\end{align*}
Writing the gradient in this way makes it clear that we can apply stochastic gradient descent by sampling positive and negative pairs according to the distributions $p^+$ and $p^-$.

\section{Theoretical Results and Proofs}

The theoretical work of this paper amounts to showing that under mild assumptions about the underlying manifold structure, a specific instantiation of the Stochastic Neighbor Embedding framework operating on the distance metric $\Delta^{\mathcal{E}}$ created from noisy samples produces data visualizations that reveal the connected components of the underlying manifold with high probability. This boils down to showing that we can control $\Delta^{\mathcal{E}}$ from above for points in the same connected component (\Cref{thm: similar pairs}), and we can control $\Delta^{\mathcal{E}}$ from below for pairs of points in different components (\Cref{thm: dissimilar pairs}). Then, we show that this property allows one to more easily meet the requirements of Theorem 3.1 of \cite{pmlr-v75-arora18a}, which then guarantees cluster visualization via Stochastic Neighbor Embeddings. This last part is shown in \Cref{thm: tsne convergence}. 

\subsection{Setting and assumptions} 

We'll begin by restating the setting and our assumptions below. We suppose that our data $X = \{x_1, \dots, x_N\}$ is drawn uniformly from a $d$-dimensional compact submanifold $\man \subset \mathbb{R}^D$ consisting of $k$ connected components. Note that we'll use the words connected component and cluster interchangeably. Equivalently, $\man$ can be represented as the union of $n_C$ nonintersecting submanifolds $\man_1, \dots, \man_{n_C}$ of the same dimension. We'll now define the clustering function $C: \man \rightarrow [n_C]$ which maps points in $\man$ to their corresponding cluster label $\in [n_C]$. Furthermore, we'll define $C_i = \{p \in \man \, | \, C(p) = i\}$ to be the set of all points in the $i$-th cluster. We also assume that underlying clusters are well separated with respect to the nearest neighbor graph parameter $\epsilon$ (\cref{assumption: cluster separation} in the main text).


\subsubsection{The noise model}\label{sec: noise model} While it is standard to model noise as ambient perturbations to $X$, in this work we depart from this convention. Instead, for our theoretical analysis, the noise will be injected into the adjacency matrix of the nearest neighbor graph built from the \textit{noiseless} data $X$. We find that this approach drastically simplifies theoretical analysis, and also empirically produces nearest neighbor graphs that structurally adhere to graphs generated via perturbing the data with ambient noise. Our experiments to support this claim are provided in \Cref{sec: noise model experiments}.

The perturbation to the adjacency matrix of the nearest neighbor graph of $X$ consists of two stages. First, for each node $x_i$ a Bernoulli random variable $\nu_i \sim \text{Bernoulli}(s_i)$ will indicate whether node $x_i$ will have \textit{any} added incident edges. The parameter $s_i$, which controls how likely this is for a node $x_i$, will be related to the distance to the nearest connected component of $\man$ such that $x_i$ is not contained in that component,
\begin{equation} \label{eq: noise model}
    s_{i} = p_{\max}\cdot\exp\biggl(-\frac{1}{\sigma^2}\inf_{x\in \man}\Bigl\{\|x-x_i\|_2^2 \, \Big| \,C(x_i)\neq C(x) \Bigr\}\biggl)
\end{equation}
where $0\leq p_{\max} \leq 1$ and $\sigma > 0.$ Intuitively, the closer a point is to a different connected component, the more likely it is to have incident edges added. For the theoretical analysis that follows, we'll define $s_{\max} = \max_{p \in \man} s_p$, which is necessarily bounded by $p_{\max}$. The second stage stochastically adds edges between points that have $\nu_i=1$. Namely, for each $x_i$ such that $\nu_i = 1$, we'll draw $\alpha N^{\beta}$ points from some suitable distribution over other points with $\nu_j =1$ to connect to point $x_i$. We assume that $\alpha < \infty$ and $0 < \beta < 1$. Such a distribution could be, for instance, 
\begin{equation}
    \mathbb{P}\Big[(x_i,x_j) \, \Big| \,\nu_i = 1, \nu_j = 1\Big] = \frac{e^{-\|x_i-x_j\|_2^2/\sigma^2}}{\sum_{x_{l} \text{ s.t. } \nu_l=1}e^{-\|x_i-x_l\|_2^2/\sigma^2}}.
\end{equation}
where edges are sampled with replacement, and thrown out if duplicated. This distribution is the one we opt for in our experiments in \Cref{sec: noise model experiments}. Again, we see that points that are closer together are more likely to be connected. 

\subsubsection{Limitations of the noise model} \label{sec: noise limitations}
We believe that, for an appropriate selection of model parameters, the noise model we present is a reasonable model for many data-generating processes for which visualization techniques have been applied. We have provided experimental evidence on synthetic data for this claim in \Cref{sec: noise model experiments}. That being said, it has some key limitations that are worth discussing. In particular, the model restricts the support of the sampling distribution to compact submanifolds of Euclidean space, a phenomenon that does not always hold in practice. A mixture of full-dimensional Gaussian distributions, for example, are not well captured by our model - this stems from the fact that the supports of the mixture components are non-compact and are not restricted to a submanifold of the space. Another example of a poorly modeled phenomenon is any nearest neighbor graph that has regions that locally look like a \say{line-graph} - these would be highly improbable under our model unless $d=1$. However, this is a scenario we might expect to encounter in practice. Finally, we note that a key limitation lies in the fact that the number of cluster-bridging edges incident to any vertex grows sub-linearly with $N$. This arises from a technical detail embedded in \Cref{lemma: similar pairs} - it allows us to avoid the computation of intractable constants. 

\subsection{Theorems}\label{sec: proof of theorems}

\subsubsection{\Cref{thm: similar pairs}} The key result of this theorem is the upper bound on $\Delta^{\mathcal{E}}$, as the lower bound arises deterministically from repeated application of the triangle inequality. To show the upper bound, we adopt the following proof strategy: we approximate the geodesic path connecting $a$ and $b$ with a path through the $\epsilon$-nearest-neighbor-graph built from $X$ with resolution increasing in $N$. This increase in resolution implies that the length of any edge along this path decreases with $N$. One can then show that this shrinking of edge length results in more neighborhood overlap, which results in more positive ORC $\kappa$. As the curvature of each edge along the path increases, the energy function will decrease to align with the Euclidean distance of that edge. Thus, the distance as measured by $\Delta^{\mathcal{E}}$ will converge to be upper bounded by the manifold geodesic distance. This discretization argument is applied for a single pair in \Cref{lemma: similar pairs}, and expanded to all pairs in \Cref{thm: similar pairs}. Before we dive into the proof of \Cref{thm: similar pairs}, we state \Cref{lemma: similar pairs}.

\begin{restatable}[Similarities]{lemma}{singlepairsimilarities}
\label{lemma: similar pairs}
    Suppose $X$ consists of $N$ i.i.d. samples from a uniform distribution over $\man$, and let $\Delta^{\mathcal{E}}$ be the metric created using \Cref{alg: EmbedOR} from $X$'s nearest neighbor graph corrupted with noisy connections as described in \Cref{eq: noise model}. Then for any $a,b \in X \times X$ we have \begin{equation}\label{eq: similarities lemma statement}
    \mathbb{P}_{X \sim \text{Uniform}(\man)}\biggl[ \Delta^{\mathcal{E}}(a,b) \leq d_{\man}(a,b)\biggr] \geq   1 - \mathcal{O}(N^{1/2 + 1/2d})e^{-\Omega(N^{1/2})}.\end{equation}
\end{restatable}
\noindent As mentioned, \Cref{lemma: similar pairs} establishes the desired bound for a single pair of points. Now we can prove \Cref{thm: similar pairs}.

\textit{Proof of \Cref{thm: similar pairs}.} Let's define $P_{\text{same}} = \{(a,b) \in X\times X \, |\,C(a) = C(b)\}.$ Rewriting the theorem statement, we have,
\begin{equation} \label{eq: all pairs similarities probability}
    \mathbb{P}\biggl[ \Delta^{\mathcal{E}}(a,b) \leq d_{\man}(a,b) \text{ $\forall$ $a,b \in X\times X$ s.t. $C(a) = C(b)$}\biggr] = \mathbb{P}\biggl[ \Delta^{\mathcal{E}}(a,b) \leq d_{\man}(a,b) \text{ $\forall$ $(a,b) \in P_{\text{same}}$}\biggr].
\end{equation}
Now we'll decompose the term on the right,
\begin{align*}
    \mathbb{P}\biggl[ \Delta^{\mathcal{E}}(a,b) \leq d_{\man}(a,b) \text{ $\forall$ $(a,b) \in P_{\text{same}}$}\,\biggr] & = 1 - \mathbb{P}\biggl[ \Delta^{\mathcal{E}}(a,b) > d_{\man}(a,b) \text{ for some $(a,b) \in P_{\text{same}}$}\biggr] \\
    & = 1 - \mathbb{P}\Biggl[\bigcup_{(a,b) \in P_{\text{same}}} \Delta^{\mathcal{E}}(a,b) > d_{\man}(a,b) \Biggr].
\end{align*}
Now if we define $D(a,b): X\times X \rightarrow\mathbb{R}$ such that
\[D(a,b) = \begin{cases}
    \Delta^{\mathcal{E}}(a,b) & (a,b) \in P_{\text{same}} \\
    0 & \text{otherwise} 
\end{cases}\]
then we have
\begin{align*}
    \mathbb{P}\biggl[ \Delta^{\mathcal{E}}(a,b) \leq d_{\man}(a,b) \text{ $\forall$ $(a,b) \in P_{\text{same}}$}\biggr] &\geq 1 - \mathbb{P}\Biggl[\bigcup_{(a,b) \in P_{\text{same}}} D(a,b) > d_{\man}(a,b) \Biggr] \\
    &\geq 1 - \mathbb{P}\Biggl[\bigcup_{(a,b) \in X \times X} D(a,b) > d_{\man}(a,b) \Biggr] \\
    &\geq 1 - \sum_{(a,b) \in X \times X}\mathbb{P}\Bigl[ D(a,b) > d_{\man}(a,b)\Bigr] \\
    &\geq 1 - N^2 \max_{(a,b) \in X \times X} \mathbb{P}\Bigl[ D(a,b) > d_{\man}(a,b)\Bigr] \\
    &= 1 - N^2 \max_{(a,b) \in X \times X} \bigg(1-\mathbb{P}\Bigl[ \Delta^{\mathcal{E}}(a,b) \leq d_{\man}(a,b)\Bigr]\bigg).
\end{align*}
We can deterministically lower bound the right-hand side by applying \Cref{lemma: similar pairs}. From this, we get \begin{align*}
    \mathbb{P}\biggl[ \Delta^{\mathcal{E}}(a,b) \leq d_{\man}(a,b) \text{ $\forall$ $(a,b) \in P_{\text{same}}$}\biggr] &\geq 1 - N^2\mathcal{O}(N^{1/2 + 1/2d})e^{-\Omega(N^{1/2})} \\
    &= 1 - \mathcal{O}(N^{5/2+1/2d})e^{-\Omega(N^{1/2})}.
\end{align*}
The proof of the lower bound on $\Delta^{\mathcal{E}}(a,b)$ follows from application of the triangle inequality. Note that this bound is independent of the stochasticity of sampling. Namely,
\begin{align*}
    \Delta^{\mathcal{E}}(a,b) &= \inf_{\Gamma}\sum_{(x,x') \in \Gamma}\frac{1}{7}\mathcal{E}(\kappa(x,x'); p)\cdot\|x-x'\|_2 \\
    &\geq \frac{1}{7}\inf_{\Gamma}\sum_{(x,x') \in \Gamma}\mathcal{E}(1)\cdot\|x-x'\|_2 \\
    &= \frac{1}{7}\inf_{\Gamma}\sum_{(x,x') \in \Gamma}\|x-x'\|_2 \\
    &\geq \frac{1}{7}\|a-b\|_2
\end{align*}
where $\Gamma$ is a path in the graph that connects $a$ and $b$ (which exists w.h.p. for sufficiently large $N$). 
\qed{}

\subsubsection{\Cref{thm: dissimilar pairs}}

\Cref{thm: dissimilar pairs} establishes a lower bound on the distance between all pairs of points in different connected components by generalizing the result of \Cref{lemma: negative curvature}, stated here. 

\begin{restatable}[cluster-bridging edges]{lemma}{dissimilarpairs}\label{lemma: negative curvature}
    Let $X \subset \man$ be a point cloud sampled from $\man$. Furthermore, let $G=(X,E)$ be the nearest neighbor graph created from the data using $\epsilon$-radius connectivity, with noisy edge additions according to the noise scheme described in \Cref{eq: noise model}. If $s_{\max} < \frac{3}{8e}$, then we have,
    \[\mathbb{P}\Big[\kappa(a,b) < -1/2 \,\, \forall \, (a,b) \text{ such that }C(a) \neq C(b)  \, \Big| \, X  \Big] > 1-2|X|^2e^{-N_{\min}s_{\max}}\]
where $N_{\min} = \min_{p \in X}|\{(B(p, \epsilon)\setminus{{p}}) \bigcap X\}|$, which is the minimum number of neighbors of any point in $X$ \textit{before} noisy edge additions.
\end{restatable}

Observe that, unlike the bound on distances between pairs in the same connected component, the bound obtained in \Cref{thm: dissimilar pairs} is (desirably) dependent on the EmbedOR algorithm parameter $p$. 

\noindent\textit{Proof.} From \Cref{lemma: negative curvature} we can then say that 
\[\mathbb{P}\Big[\kappa(i,j) < -1/2 \,\, \forall \, (i,j) \in E \text{ such that }C(i) \neq C(j)  \, \Big| \, X  \Big] > 1-2|X|^2e^{-N_{\min}s_{\max}}\]
where $N_{\min} = \min_{p \in X}|\{(B(p, \epsilon)\setminus{{p}}) \bigcap X\}|$, which is the minimum number of neighbors of any point in $X$ \textit{before} noisy edge additions. Taking the expectation with respect to $X \sim \text{Uniform}(\man)$ we get
\[\mathbb{P}\Big[\kappa(i,j) < -1/2 \,\, \forall \, (i,j) \text{ such that }C(i) \neq C(j) \Big] > 1-2\mathbb{E}_{X  \sim \text{Uniform}(\man)}\biggl[ |X|^2e^{-N_{\min}s_{\max}}\biggr].\]
Note that $\kappa(i,j) < -1/2 \,\, \forall \, (i,j) \text{ such that }C(i) \neq C(j)$ implies $w(i,j) \geq \frac{1}{7}\norm{x_i - x_j}_2 \mathcal{E}(-1/2;p)$ for all $i,j$ such that $C(i) \neq C(j)$. This follows from the definition of $w$ in \Cref{eq: weight} in \Cref{sec: the metric} of the main text. From \Cref{eq: branch separation} we know that $\norm{x_i - x_j}_2 \geq c_{ij}$, from which we have
\[\mathbb{P}\Big[w(i,j) > \frac{c_{ij}}{7}\mathcal{E}(-1/2; p) \,\, \forall \, (i,j) \text{ such that }C(i) \neq C(j) \Big] > 1-2\mathbb{E}_{X \sim \text{Uniform}(\man)}\biggl[ |X|^2e^{-N_{\min}s_{\max}}\biggr].\]
For every pair $a,b$ that are in different connected components of $\man$, every path connecting them must traverse one edge $(i,j)$ where $C(i) \neq C(j)$. Thus,
\[\mathbb{P}\Big[\Delta^{\mathcal{E}}(a,b) > \frac{c_{ij}}{7}\mathcal{E}(-1/2; p)\text{ for all $a,b$ s.t. }C(a) \neq C(b)\biggr] > 1-2\mathbb{E}_{X \sim \text{Uniform}(\man)}\biggl[ |X|^2e^{-N_{\min}s_{\max}}\biggr].\]
Applying \Cref{lemma: limit of expectation} we obtain,
\[\mathbb{P}\Big[\Delta^{\mathcal{E}}(a,b) > \frac{c_{ij}}{7}\mathcal{E}(-1/2; p)\text{ for all $a,b$ s.t. }C(a) \neq C(b) \biggr] > 1-\mathcal{O}(N^4)e^{-\Omega(N)}.\]
The proof is concluded by observing that $c_{\man} \leq c_{ij}$. 
\qed{}

\subsubsection{\Cref{thm: tsne convergence (informal)}} Our third theorem, stated informally as \Cref{thm: tsne convergence (informal)} and formally as \Cref{thm: tsne convergence}, establishes that a specific instantiation of tSNE using the metric $\Delta^{\mathcal{E}}$ produces a cluster-preserving visualization (\Cref{def: full visualization}) of the data for any dataset whose generating process aligns with that described in \Cref{sec: noise model}. To do this, \Cref{thm: tsne convergence} shows that $\Delta^{\mathcal{E}}$ satisfies \Cref{def: separated spherical data}, and in turn satisfies the requirements of Theorem 3.1 in \cite{pmlr-v75-arora18a}.

\begin{restatable}[Well-separated, spherical data \cite{pmlr-v75-arora18a}]{defn}{gamma-sperical-separable} \label{def: separated spherical data}
Let $C_1, C_2,\dots, C_{n_C}$ be the individual clusters such that for each $m \in [n_C]$, $|C_m\cap X| \geq 0.1(|X|/n_C)$. The data $X=\{x_i\}_{i=1}^{|X|}$ is said to be $\gamma$-spherical and $\gamma$-well-separated w.r.t a metric $d_X$ if for some $b_1, b_2, \dots, b_{n_C} > 0$:
\begin{itemize}
    \item 1) $\gamma$-spherical:
    \begin{itemize}
        \item 1a) for any $m\in[n_C]$ and for any $x_i,x_j\in C_m\cap X$ ($i\neq j)$, we have $d_X(x_i,x_j)^2\geq\frac{b_m}{1+\gamma}$, and 
        \item 1b) for any $x_i\in C_m \cap X$ we have 
    $\left|\left\{x_j\in C_m\cap X\setminus \{x_i\}: d_X(x_i,x_j)^2\leq b_m\right\}\right|\geq 0.51|C_m \cap X|$.
    \end{itemize}
    \item 2) $\gamma$-well-separated: for any $m,m'\in [n_C]$ ($m\neq m')$, $x_i\in C_m\cap X$ and $x_j\in C_{m'}\cap X$ we have $d_X(x_i,x_j)^2\geq(1+\gamma\log |X|)\max\{b_m, b_{m'}\}$.
\end{itemize}
\end{restatable}
In effect, this definition requires that intra-cluster distances be concentrated around some small value, while inter-cluster distances are significantly larger than that value. According to \cite{pmlr-v75-arora18a}, these assumptions are satisfied by, for example, a mixture of Gaussians with the ambient Euclidean metric. We'll show that a broad class of datasets satisfy this definition when the EmbedOR metric $\Delta^{\mathcal{E}}$ is used. 

\begin{restatable}[t-SNE convergence]{thm}{tsne}\label{thm: tsne convergence}
    Suppose $X$ consists of $N$ i.i.d. samples from a uniform distribution over $\man$, and let $\Delta^{\mathcal{E}}$ be the metric created using \Cref{alg: EmbedOR} from $X$'s nearest neighbor graph corrupted with noisy connections as described in \Cref{eq: noise model}. Suppose that:
    \begin{enumerate}
        \item $s_{\max} < \frac{3}{8e}$ from \Cref{eq: noise model}; 
        \item the number of clusters $n_C \ll N^{1/5}$; 
        \item the t-SNE stepsize $h = 1$; 
        \item the choice of \textrm{EmbedOR} algorithm parameter $p$ satisfies \Cref{eq: choice of algorithm parameter p};
        \item the t-SNE point-wise perplexity is $\tau_i = \gamma/4 \cdot \min_{j \in [N]\setminus\{i\}}\Delta^{\mathcal{E}}(x_i,x_j)^2$ with $\gamma$ defined in \Cref{eq: choice of gamma};
        \item for all $l \in [n_C]$ we have $\text{Vol}(C_l)/\text{Vol}(\man) \gg 1/(10n_C)$;
        \item the t-SNE early exaggeration parameter $g$ is chosen so that $n_C^2\sqrt{N}\log N \ll \alpha \ll N$.
    \end{enumerate}
    Then with probability at least $0.99(1-\mathcal{O}(N^4)e^{-\Omega(N^{1/2})})$ over the choice of initialization and the configurations of $X$, t-SNE after $\mathcal{O}(\frac{\sqrt{N}}{n_C^2})$ iterations using the metric $\Delta^{\mathcal{E}}$ returns a full visualization of the data $X$. 
\end{restatable}
\noindent \textit{Proof.} We'll show that \Cref{thm: similar pairs} and \Cref{thm: dissimilar pairs} imply a metric that is $\gamma$-spherical and $\gamma$-well-separated with respect to the underlying connected components of $\man$. From \Cref{thm: similar pairs}, we know that for all $x_i, x_j$ such that $C(x_i) = C(x_j)$ we have 
\begin{align*}
    \frac{1}{7}\|x_i-x_j\|_2 \leq \Delta^{\mathcal{E}}(x_i, x_j) &\leq d_{\man}(x_i,x_j) \\
    &\leq \text{diam}(\man)
\end{align*}
with probability at least $1-\mathcal{O}(N^{5/2+1/2d})e^{-\Omega(N^{1/2})}$, where $\text{diam}(\man)$ is defined in \Cref{eq: diameter}. Thus, we can choose all $b_l$'s in \Cref{def: separated spherical data} to be $\text{diam}(\man)$ and we can choose
\begin{equation}\label{eq: choice of gamma}
    \gamma = \bigg(\frac{7}{\min_{ij}\|x_i-x_j\|_2}\bigg)^2\text{diam}(\man)-1.
\end{equation}
This choice means the data with the metric $\Delta^{\mathcal{E}}$ satisfies the $\gamma$-spherical criterion. Now we'll turn our attention to the $\gamma$-well separated criterion. From \Cref{thm: dissimilar pairs} we know that for $x_i, x_j$ such that $C(x_i) \neq C(x_j)$ we have
\[\Delta^{\mathcal{E}}(x_i,x_j) \geq \frac{c_{\man}}{7}\mathcal{E}(-1/2; p)\]
with probability at least $1-\mathcal{O}(N^4)e^{-\Omega(N)}.$ Since $\mathcal{E}(-1/2; p)$ approaches infinity as $p$ increases, we can simply choose a large enough $p$ so that
\begin{align*}
    \mathcal{E}(-1/2;p)^2 &\geq \Big(\frac{7}{c_{\man}}\Big)^2(1+\gamma\log N)\cdot\text{diam}(\man) \\
    &= \Big(\frac{7}{c_{\man}}\Big)^2\Bigg(1 +  \bigg(\bigg(\frac{7}{\min_{ij}\|x_i-x_j\|_2}\bigg)^2\text{diam}(\man)-1\bigg)\log(N)\Bigg)\text{diam}(\man).
\end{align*}
Namely, a choice of $p$ such that
\begin{equation}\label{eq: choice of algorithm parameter p}
    p\geq \frac{\log\Bigg(-1 + \frac{7}{c_{\man}}\sqrt{1 + \bigg[\Big(\big(\frac{7}{\min_{ij}\|x_i-x_j\|_2}\big)^2\text{diam}(\man)-1\Big)\log(N)\bigg]\text{diam}(\man)}\,\Bigg)}{\log\bigg(1-\frac{\log(3/4)}{\log(3/2)}\bigg)}
\end{equation}
results in satisfaction of the desired inequality. Before moving on, we will remark on a few things about this expression. 

\begin{restatable}[Choice of $p$]{rmk}{p parameter}
    Firstly, we note that this expression for $p$ is increasing in $N$, which is a potentially undesirable property. However, this property arises from the results of \cite{pmlr-v75-arora18a}, where they require increasingly well-behaved data as $N$ increases. Secondly, we note that this bound on $p$ is dependent on $c_{\man}$, which we wouldn't expect to be able to estimate in practice. With that being said, this bound establishes that for a \textit{fixed} point cloud generated from some underlying manifold $\man$, there exists a sufficiently large $p$ such that the requirements on the metric hold. This would \textit{not} be true for a more naive choice of metric, like the ambient Euclidean metric, for example. 
\end{restatable}

Applying a union bound to both conditions on $\Delta^{\mathcal{E}}$, we know that both the $\gamma$-spherical and $\gamma$-well-separated criteria hold with probability at least $1-\mathcal{O}(N^4)e^{-\Omega(N^{1/2})}$. This, in conjunction with the assumptions, allows us to apply Theorem 3.1 from \cite{pmlr-v75-arora18a} to conclude the proof. Since we have $g > n_C^2\sqrt{N}\log N$, we know that tSNE will return a valid visualization in $\mathcal{O}(\frac{\sqrt{N}}{n_C^2})$ iterations.
\begin{restatable}[Density requirement]{rmk}{density requirement}
    The density requirement in Theorem 3.1 of \cite{pmlr-v75-arora18a}, which states that each of the $n_C$ individual clusters have to have at least $0.1(N/n_C)$ points, is satisfied with high probability due to the assumption that $\text{Vol}(C_l)/\text{Vol}(\man) \gg 1/(10n_C)$. To be completely rigorous, one should bound the joint probability of this density condition and the bounds on $\Delta^{\mathcal{E}}$, instead of the marginal probability of the latter. That being said, the arguments we make are quite general and would work in such a situation. This would simply require a decomposition of $N$ into the number of points per cluster $\{N_i\}_{i \in [n_C]}$ and applying similar arguments about $\Delta^{\mathcal{E}}$ cluster-wise. Stitching it together would then require showing that $N_i > 0.1(N/n_C)$ with high probability, which follows from the assumption about the per-cluster volumes. Note that we expect these two events to be correlated; an even distribution of points among clusters should help the other results as opposed to hurting.
\end{restatable}
\qed{}

\subsection{Lemmata}

\Cref{lemma: similar pairs} establishes a result that acts as a building block to \Cref{thm: similar pairs}. Particularly, it establishes the upper bound on $\Delta^{\mathcal{E}}$ for any \textit{single} pair of points in the same connected component. 

\singlepairsimilarities*

Before diving into the technical details, we will illustrate the high-level argument of the lemma. Our argument is as follows: we approximate a shortest geodesic path through $\man$ connecting $a$ and $b$ with a path through the $\epsilon$-nearest-neighbor-graph built from $X$ whose resolution is increasing in $N$. This increase in resolution implies that the length of any edge along this path decreases with $N$. One can then show that this shrinking of edge length results in more neighborhood overlap, which results in more positive ORC $\kappa$. As the curvature of each edge along the path increases, the energy function will decrease to align with the Euclidean distance of that edge. Thus, the distance as measured by $\Delta^{\mathcal{E}}$ will converge to be upper bounded by the manifold geodesic distance.

\noindent \textit{Proof.} Let's define $\delta = \epsilon N^{-1/2d}/3$, and we'll define $n$ as follows
\[n = \bigg\lceil\frac{d_{\man}(a,b)}{\delta}\bigg\rceil.\]
Consider now the geodesic path between $a$ and $b$. We'll divide the geodesic path $\gamma$ connecting $a$ and $b$ into $n$ segments of equal length. While $a$ and $b$ are random variables, we can upper bound their geodesic distance deterministically based on the geometry of $\man$. Namely, let's define 
\begin{equation}\label{eq: diameter}
    \text{diam}(\man) = \sup_{(x,y) \in \man}\{d_{\man}(x,y) \, | \, C(x) = C(y)\}.
\end{equation}
Observe that 
\begin{equation} \label{eq: n}
n \leq \bigg\lceil\frac{\text{diam}(\man)}{\delta}\bigg\rceil = \mathcal{O}(N^{1/2d})
\end{equation}
which is a deterministic quantity. Note that these $n$ segments of $\gamma$ have length necessarily $\leq \delta$ by construction of $n$. Let the endpoints of these segments be defined by $\{x_0=a, x_1, \dots, x_{n-1}, x_n = b\}$. Note that all $x_i$'s are random variables. Let's also define $\{a_i\}_{i=1}^{n-1} \subset X$, where $a_i \in X$ is the closest point to $x_i$ with respect to geodesic distance.

\begin{restatable}[$\delta$-sampling condition]{defn}{sampling condition} \label{def: sampling condition}
The \say{$\delta$-sampling condition} holds if for all $p \in \man$ there exists a data point $a_i \in X$ such that $d_{\man}(a_i, p) < \delta$. We'll denote $A$ to be the set of all events for which the $\delta$-sampling condition holds.
\end{restatable}
\noindent From \Cref{lemma: delta sampling assumption} and \Cref{corollary: sampling condition}, we know that 
\begin{equation} \label{eq: sampling condition bound}
    \mathbb{P}[A] \geq 1-\mathcal{O}(N^{1/2})e^{-\Omega(N^{1/2})}.
\end{equation}
The $\delta$-sampling condition ensures that our set $\{a_i\}_{i=1}^{n-1} 
\subset X$ has the property that $d_{\man}(a_i, x_i) < \delta$ for all $i$. Now let's define the function
\begin{equation}
    K(a_i,a_{i+1}) = \begin{cases}
        \kappa(a_i, a_{i+1}) & \text{if }\|a_i-a_{i+1}\|_2 \leq 3\delta \\
        -2 & \text{otherwise}
    \end{cases}.
\end{equation}
By \Cref{lemma: metric bound},
\begin{align}
    \mathbb{P}\biggl[ \Delta^{\mathcal{E}}(a,b) \leq d_{\man}(a,b)\biggr] &\geq \mathbb{P}\biggl[ K(a_i, a_{i+1}) \geq 0 \text{ for all }i \in [n-1]\biggr]\end{align}
where $E$ is the (stochastic) edge set of the corrupted nearest neighbor graph of $X$. By applying a union bound, we obtain
\begin{align}
    \mathbb{P}\biggl[ \Delta^{\mathcal{E}}(a,b) \leq d_{\man}(a,b)\biggr] &\geq 1-\mathbb{P}\biggl[ K(a_i, a_{i+1}) < 0 \text{ for some }i \in [n-1]\biggr] \\
    &\geq 1- \sum_{i=1}^{n-1} \mathbb{P}[K(a_i, a_{i+1}) < 0]\,.
\end{align}
Even though $n$ is a random variable, we have the deterministic bound $n = \mathcal{O}(N^{1/2d})$. This gives us
\begin{align}
    \mathbb{P}\biggl[ \Delta^{\mathcal{E}}(a,b) \leq d_{\man}(a,b)\biggr] &\geq 1-\mathcal{O}(N^{1/2d})\cdot \max_{i} \,\mathbb{P}\biggl[K(a_i, a_{i+1}) < 0\biggr]. \label{eq: similarities probability bound in terms of K}
\end{align}
Note that we will obtain a deterministic bound for the $\max$ term (independent of the distance between $a_i$ and $a_{i+1}$), rendering the inequality above valid. Bounding the rightmost term, we see
\begin{align*}
    \mathbb{P}\biggl[K(a_i, a_{i+1}) < 0\biggr] &= \mathbb{P}\Big[K(a_i, a_{i+1}) < 0 \, \Big| \, \|a_i - a_{i+1}\|_2 \leq 3\delta\Big] \mathbb{P}\Big[\|a_i - a_{i+1}\|_2 \leq 3\delta\Big] \\
    &\hspace{20mm} + \mathbb{P}\Big[K(a_i, a_{i+1}) < 0 \, \Big| \, \|a_i - a_{i+1}\|_2 > 3\delta\Big] \mathbb{P}\Big[\|a_i - a_{i+1}\|_2 > 3\delta\Big]\\
&\leq \mathbb{P}\Big[\kappa(a_i, a_{i+1}) < 0 \, \Big| \, \|a_i - a_{i+1}\|_2 \leq 3\delta\Big] \mathbb{P}\Big[\|a_i - a_{i+1}\|_2 \leq 3\delta\Big] \\
    &\hspace{20mm} + \mathbb{P}\Big[\|a_i - a_{i+1}\|_2 > 3\delta\Big] \\
    &= 1 - \mathbb{P}\Big[\|a_i - a_{i+1}\|_2 \leq 3\delta\Big]\bigg(1 - \mathbb{P}\Big[\kappa(a_i, a_{i+1}) < 0 \, \Big| \, \|a_i - a_{i+1}\|_2 \leq 3\delta\Big]\bigg).
\end{align*}
Now we'll further upper bound this term by \textit{lower} bounding $\mathbb{P}\Big[\|a_i - a_{i+1}\|_2 \leq 3\delta\Big]$. Recall that if the $\delta$-sampling condition holds, then
\begin{align*}
    \|a_i- a_{i+1}\|_2 &\leq d_{\man}(a_i, a_{i+1})\\
     &\leq  d_{\man}(a_i, x_{i}) + d_{\man}(x_{i}, x_{i+1}) + d_{\man}(x_{i+1}, a_{i+1}) \\
    & \leq 3\delta.
\end{align*}
Thus it follows that $\mathbb{P}[A] \leq \mathbb{P}\Big[\|a_i-a_{i+1}\|_2 \leq 3\delta\Big]$, where $A$ is the event that the $\delta$-sampling condition holds, and we have
\begin{align*}
    \mathbb{P}\biggl[K(a_i, a_{i+1}) < 0\biggr] &\leq 1 - \mathbb{P}[A]\bigg(1 - \mathbb{P}\Big[\kappa(a_i, a_{i+1}) < 0 \, \Big| \, \|a_i - a_{i+1}\|_2 \leq 3\delta\Big]\bigg).
\end{align*}
Now we can plug back into \Cref{eq: similarities probability bound in terms of K} to obtain
\begin{equation}
    \mathbb{P}\biggl[ \Delta^{\mathcal{E}}(a,b) \leq d_{\man}(a,b)\biggr]
    \geq 1-\mathcal{O}(N^{1/2d}) \cdot
    \max_{i} \,\Bigg(1 - \mathbb{P}[A]\cdot\mathbb{P}\Big[\kappa(a_i, a_{i+1}) \geq 0 \, \Big| \, \|a_i - a_{i+1}\|_2 \leq 3\delta\Big]\Bigg). \label{eq: similarities probability bound not in terms of K}
\end{equation}
Now we will pursue a lower bound on the conditional probability term above. Let $N_{I_i}$ be the number of points in $I_i = \mathcal{N}_{a_{i+1}}(a_i) \, \cap \, \mathcal{N}_{a_{i}}(a_{i+1})$, and let $N_{U_i}$ be the number of points in $U_i = \big(\mathcal{N}_{a_{i+1}}(a_i) \, \cup \, \mathcal{N}_{a_{i}}(a_{i+1})\big) \, \setminus \, I_i$, where $\mathcal{N}_y(x) = \{p \in X \setminus{y} \, | \, d_{G}(p,x) = 1\}.$ We can invoke Lemma A.2 from \cite{saidi2025recovering} to say that for some $\beta \in  (0,1)$
\begin{align*}
    \mathbb{P}\Bigl[\kappa(a_i, a_{i+1}) \geq 1-4(1-\beta) \, \Big| \, \|a_i - a_{i+1}\|_2 \leq 3\delta\Bigr] &\geq \mathbb{P}\biggl[\frac{N_{I_i}}{N_{I_i} + N_{U_i}} > \beta \, \Big| \, \|a_i - a_{i+1}\|_2 \leq 3\delta\biggr] \\
    &= \mathbb{P}\biggl[(1-\beta)N_{I_i} > \beta N_{U_i} \, \bigg| \, \|a_i - a_{i+1}\|_2 \leq 3\delta\biggr] \\
    &= \mathbb{P}\biggl[N_{I_i} > \frac{\beta}{1-\beta} N_{U_i} \, \bigg| \, \|a_i - a_{i+1}\|_2 \leq 3\delta\biggr].
\end{align*}
If we choose $\beta = 3/4$, we have
\begin{align}
    \mathbb{P}\Bigl[\kappa(a_i, a_{i+1}) \geq 0 \, \Big| \, \|a_i - a_{i+1}\|_2 \leq 3\delta\Bigr]&\geq \mathbb{P}\Bigl[N_{I_i} > 3 N_{U_i} \, \Big| \, \|a_i - a_{i+1}\|_2 \leq 3\delta\Bigr]. \label{eq: similarities probability bound}
\end{align}
Thus \Cref{eq: similarities probability bound not in terms of K} becomes 
\begin{equation}
    \mathbb{P}\biggl[ \Delta^{\mathcal{E}}(a,b) \leq d_{\man}(a,b)\biggr]
    \geq 1-\mathcal{O}(N^{1/2d}) \cdot
    \max_{i} \,\Bigg(1 - \mathbb{P}[A]\cdot\mathbb{P}\Bigl[N_{I_i} > 3 N_{U_i} \, \Big| \, \|a_i - a_{i+1}\|_2 \leq 3\delta\Bigr]\Bigg). \label{eq: similarities probability bound not in terms of K (2)}
\end{equation}
Now we'll analyze this term on the right. Let's define $V_{I_i} = \text{Vol}(I_i)$ and $V_{U_i} = \text{Vol}(U_i)$. If we consider locally Euclidean intrinsic geometry we can invoke \Cref{lemma: eball overlap} and \Cref{corollary: monotonicity of ball volumes} to obtain bounds for $V_{U_i}$ and $V_{I_i}$ in terms of $\delta$ as follows,
\begin{align}
    V_{U_i} < 2\eta_d\epsilon^d \biggl(1- \mathcal{I}_{[1 - (3\delta/2\epsilon)^2]}\Bigl(\frac{d+1}{2}, \frac{1}{2}\Bigr)\biggr)
\end{align}
and 
\begin{align}
    V_{I_i} >  \eta_d\epsilon^d \mathcal{I}_{[1 - (3\delta/2\epsilon)^2]}\Bigl(\frac{d+1}{2}, \frac{1}{2}\Bigr).
\end{align}
Note that $\eta_d$ is the volume of a $d$-dimensional Euclidean ball. Now we'll define
\begin{align}
    p_i = \frac{\mathcal{I}_{[1 - (3\delta/2\epsilon)^2]}\Bigl(\frac{d+1}{2}, \frac{1}{2}\Bigr)}{2- \mathcal{I}_{[1 - (3\delta/2\epsilon)^2]}\Bigl(\frac{d+1}{2}, \frac{1}{2}\Bigr)} \label{eq: p_i}
\end{align}
which simply represents a lower bound on the ratio $V_{I_i}/(V_{I_i}+V_{U_i})$. Note that \Cref{prop: p_i} indicates that $p_i = 1- \mathcal{O}(N^{-1/2d}).$ Now let's decompose $N_{I_i}$ and $N_{U_i}$ as follows. Observe that $N_{I_i}$ counts both (1) points that are within $\epsilon$ of $a_i$ and $a_{i+1}$, and (2) points that become neighbors of both $a_i, a_{i+1}$ only after the noisy edge additions described in \Cref{sec: noise model}. The same holds for the points contributing to the term $N_{U_i}$. Thus, let's decompose these quantities as follows
\begin{align*}
    N_{I_i} &= N_{I_i}^{\text{orig}} + N_{I_i}^{\text{noise}} \\
    N_{U_i} &= N_{U_i}^{\text{orig}} + N_{U_i}^{\text{noise}}.
\end{align*}
Let's further decompose 
\begin{align*}
    N_{i}^{\text{orig}} &= N_{I_i}^{\text{orig}} + N_{U_i}^{\text{orig}} \\
    N_{i}^{\text{noise}} &= N_{I_i}^{\text{noise}} + N_{U_i}^{\text{noise}}
\end{align*}
which represent the total number of original points incident to $a_i$, $a_{i+1}$ and the total number of noisy incident additions to $a_i$, $a_{i+1}$ respectively. From \Cref{sec: noise model} we know that the number of incident edge additions for any point is bounded by $\alpha N^{\beta}$ for some $\beta > 0$ and $0 < \gamma < 1$. Thus
\begin{align*}
    0 \leq N_{I_i}^{\text{noise}} \leq \alpha N^{\beta} \\
    0 \leq N_{U_i}^{\text{noise}} \leq 2\alpha N^{\beta}. 
\end{align*}
Now we can say
\begin{align*}
    \mathbb{P}\Bigl[N_{I_i} > 3 N_{U_i} \, \Big| \, \|a_i - a_{i+1}\|_2 & \leq 3\delta\Bigr]\\
    &= \mathbb{P}\Bigl[N_{I_i}^{\text{orig}} + N_{I_i}^{\text{noise}} > 3 (N_{U_i}^{\text{orig}} + N_{U_i}^{\text{noise}}) \, \Big| \, \|a_i - a_{i+1}\|_2 \leq 3\delta\Bigr] \\
    &\geq \mathbb{P}\Bigl[N_{I_i}^{\text{orig}} > 3 (N_{U_i}^{\text{orig}} + 2\alpha N^{\beta}) \, \Big| \, \|a_i - a_{i+1}\|_2 \leq 3\delta\Bigr] \\
    &= \mathbb{E}_{N_{i}^{\text{orig}}}\bigg[\mathbb{P}\Bigl[N_{I_i}^{\text{orig}} > 3 (N_{U_i}^{\text{orig}} + 2\alpha N^{\beta}) \, \Big| \, \|a_i - a_{i+1}\|_2 \leq 3\delta, N_i^{\text{orig}}\Big]\bigg] \\
    &= \mathbb{E}_{N_{i}^{\text{orig}}}\bigg[\mathbb{P}\Bigl[N_{I_i}^{\text{orig}} > 3 (N_{i}^{\text{orig}} - N_{I_i}^{\text{orig}} + 2\alpha N^{\beta}) \, \Big| \, \|a_i - a_{i+1}\|_2 \leq 3\delta, N_i^{\text{orig}}\Big]\bigg] \\
    &= \mathbb{E}_{N_{i}^{\text{orig}}}\bigg[\mathbb{P}\Bigl[N_{I_i}^{\text{orig}} > \frac{3}{4} (N_{i}^{\text{orig}} + 2\alpha N^{\beta}) \, \Big| \, \|a_i - a_{i+1}\|_2 \leq 3\delta, N_i^{\text{orig}}\Big]\bigg].
\end{align*}
If we define $Z_i \sim \text{Binomial}(N_i^{\text{orig}}, p_i)$ with $p_i$ defined in \Cref{eq: p_i}, then we can say 
\begin{align*}
    \mathbb{P}\Bigl[N_{I_i} > 3 N_{U_i} \, \Big| \, \|a_i - a_{i+1}\|_2 \leq 3\delta\Bigr] &\geq \mathbb{E}_{N_{i}^{\text{orig}}}\bigg[\mathbb{P}\Bigl[Z_i > \frac{3}{4} (N_{i}^{\text{orig}} + 2\alpha N^{\beta}) \, \Big| \, \|a_i - a_{i+1}\|_2 \leq 3\delta, N_i^{\text{orig}}\Big]\bigg].
\end{align*}
Now let $j_* = \Big\lceil\frac{3\alpha N^{\beta}/2}{p_i - 3/4}\Big\rceil$. Since $p_i = 1 -\mathcal{O}(N^{-1/2d})$ we know that for sufficiently large $N$, the integer $j_*$ is necessarily positive and finite. We'll lower bound the right side as follows
\begin{equation}
    \mathbb{P}\Bigl[N_{I_i} > 3 N_{U_i} \, \Big| \, \|a_i - a_{i+1}\|_2 \leq 3\delta\Bigr] \geq \sum_{j > j_*}\mathbb{P}\Big[N_{i}^{\text{orig}} = j\Big]\cdot \mathbb{P}\Bigl[Z_i > \frac{3}{4} (N_{i}^{\text{orig}} + 2\alpha N^{\beta}) \, \Big| \, \|a_i - a_{i+1}\|_2 \leq 3\delta, N_i^{\text{orig}}=j\Big]. \label{eq: finite sum expectation bound}
\end{equation}
Let $\mu = \mathbb{E}\Big[Z_i|N_{i}^{\text{orig}}\Big] = N_{i}^{\text{orig}}p_i$ and let $c = \frac{3}{4} (N_{i}^{\text{orig}} + 2\alpha N^{\beta})$. Applying a Chernoff bound for the binomial random variable \cite{tsun2020probability}, we get
\begin{align*}
    \mathbb{P}\Bigl[Z_i > \frac{3}{4} (N_{i}^{\text{orig}} + 2\alpha N^{\gamma}) \, \Big| \, \|a_i - a_{i+1}\|_2 \leq 3\delta, N_i^{\text{orig}}=j\Big] &\geq 1-\exp\bigg\{-\frac{(1-c/\mu)^2\mu}{2}\bigg\}\\
    &= 1-\exp\bigg\{-\frac{(\mu-c)^2}{2\mu}\bigg\}
\end{align*}
for $\mu > c$. Note that our restriction of $N_{i}^{\text{orig}} > j_*$ in \Cref{eq: finite sum expectation bound} implies that $\mu > c$, rendering our bound valid. Plugging back in the definitions for $\mu$ and $c$ we get
\begin{align*}
    \mathbb{P}\biggl[Z_i > \frac{3}{4} (N_{i}^{\text{orig}} + 2\alpha N^{\beta}) \, \Big| \, \|a_i - &a_{i+1}\|_2 \leq 3\delta, N_i^{\text{orig}}=j\biggr] \\
    &\geq 1-\exp\bigg\{-\frac{(N_{i}^{\text{orig}}p_i - \frac{3}{4}(N_{i}^{\text{orig}} + 2\alpha N^{\beta}))^2}{2N_{i}^{\text{orig}}p_i}\bigg\} \\ &= 1-\exp\bigg\{-\frac{\big(N_{i}^{\text{orig}}(p_i-3/4) - 3\alpha N^{\beta}/2\big)^2}{2N_{i}^{\text{orig}}p_i}\bigg\}\\
    &= 1-\exp\bigg\{-\frac{(N_{i}^{\text{orig}})^2(p_i-3/4)^2 - 3\alpha N^{\beta} N_{i}^{\text{orig}}(p_i-3/4) + (3\alpha N^{\beta}/2)^2}{2N_{i}^{\text{orig}}p_i}\bigg\}\\
    &\geq 1-\exp\bigg\{-\frac{(N_{i}^{\text{orig}})^2(p_i-3/4)^2 - 3\alpha N^{\beta} N_{i}^{\text{orig}}(p_i-3/4)}{2N_{i}^{\text{orig}}p_i}\bigg\}\\
    &= 1-\exp\bigg\{-\frac{N_{i}^{\text{orig}}(p_i-3/4)^2 - 3\alpha N^{\beta} (p_i-3/4)}{2p_i}\bigg\}.
\end{align*}
We'll continue to lower bound,
\begin{align*}
    \mathbb{P}\biggl[Z_i > \frac{3}{4} (N_{i}^{\text{orig}} + 2\alpha N^{\beta}) \, \Big| \, \|a_i - a_{i+1}\|_2 \leq 3\delta, N_i^{\text{orig}}=j\biggr] &\geq  1-\exp\bigg\{-\frac{N_{i}^{\text{orig}}(p_i-3/4)^2 - 3\alpha N^{\beta} (p_i-3/4)}{2p_i}\bigg\} \\
    &= 1-\exp\bigg\{-\frac{N_{i}^{\text{orig}}(p_i-3/4)^2}{2p_i} + \frac{3\alpha N^{\beta} (p_i-3/4)}{2p_i}\bigg\} \\
    &\geq 1-\exp\bigg\{-\frac{N_{i}^{\text{orig}}(p_i-3/4)^2}{2p_i} + \frac{3}{2}\alpha N^{\beta}\bigg\} \\
    &\geq 1-\exp\bigg\{-\frac{N_{i}^{\text{orig}}}{2}(p_i-3/4)^2 + \frac{3}{2}\alpha N^{\beta}\bigg\} \\
    &= 1 - e^{-\frac{N_{i}^{\text{orig}}}{2}(p_i-3/4)^2}\cdot e^{\frac{3}{2}\alpha N^{\beta}}.
\end{align*}
Plugging back into \Cref{eq: finite sum expectation bound}, we obtain
\begin{align}
     \mathbb{P}\Bigl[N_{I_i} > 3 N_{U_i} \, \Big| \, \|a_i - a_{i+1}\|_2 \leq 3\delta\Bigr]&\geq \sum_{j > j_*}\mathbb{P}\Big[N_{i}^{\text{orig}} = j\Big]\cdot \Big(1 - e^{-\frac{N_{i}^{\text{orig}}}{2}(p_i-3/4)^2}\cdot e^{\frac{3}{2}\alpha N^{\beta}}\Big) \notag \\
     &= \sum_{j = 0}^{\infty}\mathbb{P}\Big[N_{i}^{\text{orig}} = j\Big]\cdot \Big(1 - e^{-\frac{N_{i}^{\text{orig}}}{2}(p_i-3/4)^2}\cdot e^{\frac{3}{2}\alpha N^{\beta}}\Big) \notag \\
     &\hspace{20mm}- \sum_{j \leq j_*}\mathbb{P}\Big[N_{i}^{\text{orig}} = j\Big]\cdot \Big(1 - e^{-\frac{N_{i}^{\text{orig}}}{2}(p_i-3/4)^2}\cdot e^{\frac{3}{2}\alpha N^{\beta}}\Big) \notag \\
     &\geq \mathbb{E}_{N_i^{\text{orig}}}\Big[1 - e^{-\frac{N_{i}^{\text{orig}}}{2}(p_i-3/4)^2}\cdot e^{\frac{3}{2}\alpha N^{\beta}}\Big] - \mathbb{P}\Big[N_{i}^{\text{orig}} \leq j_*\Big] \notag \\
     &= 1-e^{\frac{3}{2}\alpha N^{\beta}}\cdot\mathbb{E}_{N_i^{\text{orig}}}\Big[e^{-\frac{N_{i}^{\text{orig}}}{2}(p_i-3/4)^2}\Big] - \mathbb{P}\Big[N_{i}^{\text{orig}} \leq j_*\Big]. \label{eq:bound1}
\end{align}
Now we'll bound the last two terms above. Observe that $N_{i}^{\text{orig}} \sim \text{Binomial}\Big(N-2, \frac{V_{I_i}+V_{U_i}}{\text{Vol}(\man)}\Big)$. Let's define $\lambda = \mathbb{E}[N_i^{\text{orig}}]=(N-2)\frac{V_{I_i}+V_{U_i}}{\text{Vol}(\man)}$. Note that if $\alpha$ satisfies the following inequality
\begin{equation} \label{eq: alpha bound}
\alpha < \frac{2(p_i-3/4)}{3N^{\beta}}\bigg(\frac{(N-2)\eta_d\epsilon^d}{\text{Vol}(\man)} - 1\bigg)
\end{equation}
then we have $\lambda > 3j_*$. Note that the right hand side asymptotes to $\infty$ with rate $\Omega(N^{1-\beta})$. Thus \Cref{eq: alpha bound} is satisfied for sufficiently large $N$. In this case, we can apply a Chernoff bound to say
\begin{align*}
    \mathbb{P}[N_i^{\text{orig}} \leq j_*] &\leq \exp\bigg(-\frac{(\lambda-j_*)^2}{2\lambda}\bigg) \\
    &\leq \exp\bigg(-\frac{\lambda}{2}+j_*\bigg) \\
    &\leq \exp\bigg(-\frac{\lambda}{6}\bigg) \\
    &=  e^{-\Omega(N)}.
\end{align*}
Plugging back into \Cref{eq:bound1}, we have
\begin{align}\label{eq:bound2}
     \mathbb{P}\Bigl[N_{I_i} > 3 N_{U_i} \, \Big| \, \|a_i - a_{i+1}\|_2 \leq 3\delta\Bigr]&\geq 1-e^{\frac{3}{2}\alpha N^{\beta}}\cdot\mathbb{E}_{N_i^{\text{orig}}}\Big[e^{-\frac{N_{i}^{\text{orig}}}{2}(p_i-3/4)^2}\Big] - e^{-\Omega(N)}.
\end{align}
Now we'll bound the expectation. Observe that 
\begin{align*}
    \mathbb{E}_{N_i^{\text{orig}}}\Big[e^{-\frac{N_{i}^{\text{orig}}}{2}(p_i-3/4)^2}\Big] &= \mathbb{E}_{N_i^{\text{orig}}}\Big[e^{-N_{i}^{\text{orig}}\Omega(1)}\Big].
\end{align*}
Since $N_i^{\text{orig}}$ is distributed as a binomial random variable, we can use the MGF to say 
\begin{align*}
    \mathbb{E}_{N_i^{\text{orig}}}\Big[e^{-N_i^{\text{orig}}\mathcal{O}(1)}\Big]  &= \bigg(1-\frac{V_{U_i}+V_{I_i}}{\text{Vol}(\man)} + \frac{V_{U_i}+V_{I_i}}{\text{Vol}(\man)}e^{-\Omega(1)}\bigg)^{N-2} \\
    &= \bigg(1-\frac{V_{U_i}+V_{I_i}}{\text{Vol}(\man)} + \frac{V_{U_i}+V_{I_i}}{\text{Vol}(\man)}e^{-\Omega(1)}\bigg)^{\Omega(N)} \\
    &=C^{\Omega(N)}
\end{align*}
for some $C < 1$. Therefore we have
\begin{align*}
    \mathbb{E}_{N_i^{\text{orig}}}\Big[e^{-N_i^{\text{orig}}\mathcal{O}(1)}\Big]  &= C_*^{-\Omega(N)}
\end{align*}
for some $C_* > 1$. Manipulating further, we obtain
\begin{align*}
    \mathbb{E}_{N_i^{\text{orig}}}\Big[e^{-N_i^{\text{orig}}\mathcal{O}(1)}\Big]  &=  e^{-\Omega(N) \ln(C_*)} \\
    &= e^{-\Omega(N)}.
\end{align*}
Plugging into \Cref{eq:bound2}, we have
\begin{align*}
     \mathbb{P}\Bigl[N_{I_i} > 3 N_{U_i} \, \Big| \, \|a_i - a_{i+1}\|_2 \leq 3\delta\Bigr]&\geq 1-e^{\frac{3}{2}\alpha N^{\beta}}\cdot e^{-\Omega(N)} - e^{-\Omega(N)} \\
     &= 1 - 2e^{-\Omega(N)}
\end{align*}
since $\beta < 1$. Plugging this and \Cref{eq: sampling condition bound} into \Cref{eq: similarities probability bound not in terms of K (2)} we obtain 
\begin{align*}
    \mathbb{P}\biggl[ \Delta^{\mathcal{E}}(a,b) \leq d_{\man}(a,b)\biggr]
    &\geq1-\mathcal{O}(N^{1/2d}) \cdot
    \Bigg(1 - \Big(1-\mathcal{O}(N^{1/2})e^{-\Omega(N^{1/2})}\Big)\cdot\Big(1 - 2e^{-\Omega(N)}\Big)\Bigg) \\
    &= 1 - \mathcal{O}(N^{1/2+1/2d})e^{-\Omega(N^{1/2})}.
\end{align*}
\qed{}

\begin{restatable}[$\Delta^{\mathcal{E}}$ bound]{lemma}{metric bound}
\label{lemma: metric bound}
    Let $G$ be a graph containing a connected path $\{a_i\}_{i\in[n]} \subset \man$, and let $\{x_i\}_{i \in [n]} \subset \man$ be a set  of points that divides the geodesic path between $a_0$ and $a_n$ into $n$ equal-length segments. Suppose that $a_0 = x_0$, $a_n = x_n$ and for all other $i$ we have $d_{\man}(a_i, x_i) < \delta$. Also suppose $\delta$ and $n$ are defined as in \Cref{lemma: similar pairs}. If $\kappa(a_i, a_{i+1}) \geq 0$ for all $i$ then we have,
    \[\Delta^{\mathcal{E}}(a_0, a_n) \leq d_{\man}(a_0, a_n).\]
\end{restatable}
\noindent \textit{Proof.} By the definition of $\Delta^{\mathcal{E}}$ we have,
\begin{align*}
    \Delta^{\mathcal{E}}(a_0,a_n) &\leq \sum_{i=0}^{n-1}w(a_i,a_{i+1}) \\
    &= \frac{1}{7}\sum_{i=0}^{n-1}\mathcal{E}(\kappa(a_i,a_{i+1}); p)\|a_i-a_{i+1}\|_2.
\end{align*}
Now, since $\mathcal{E}$ is monotonically decreasing in $\kappa$ we can say
\[\Delta^{\mathcal{E}}(a,b) \leq \frac{1}{7}\mathcal{E}(0;p)\sum_{i=0}^{n-1}\|a_i-a_{i+1}\|_2.\] 
Note that
\begin{align*}
    \|a_i-a_{i+1}\|_2 &\leq d_{\man}(a_i, a_{i+1}) \\
    &\leq d_{\man}(a_i, x_{i}) + d_{\man}(x_{i}, x_{i+1}) + d_{\man}(x_{i+1}, a_{i+1})\\
    &< d_{\man}(x_{i}, x_{i+1}) + 2\delta
\end{align*}
where $\{x_i\}_i$ represent the endpoints of the $n$ segments of the geodesic path between $a_0$ and $a_n$ as defined in \Cref{lemma: similar pairs}. Summing over all segments, we see that 
\begin{align*}
    \sum_{i=0}^{n-1}\|a_i-a_{i+1}\|_2 &< 2n\delta + \sum_{i=0}^{n-1}d_{\man}(x_{i}, x_{i+1}) \\
    &= 3 d_{\man}(a_0,a_n) + 2\delta
\end{align*}
since $n\delta < d_{\man}(a_0, a_n) + \delta$ by definition of $\delta$ and since $\delta \ll d_{\man}(a_0, a_n)$. Thus, from the fact that $\mathcal{E}(0; p) = 2$ it follows that
\[\Delta^{\mathcal{E}}(a_0,a_n) \leq d_{\man}(a_0,a_n).\]
\qed{}

\begin{restatable}[$\delta$-sampling assumption \cite{bernstein2000graph}]{lemma}{delta sampling assumption}
\label{lemma: delta sampling assumption}
    Suppose $X$ consists of $N$ i.i.d. samples from a uniform distribution over $\man$. Assuming a small $\delta$ relative to the curvature of geodesics in $\man$ and ignoring boundary effects, then \begin{equation}\label{eq: sampling condition} \mathbb{P}\bigl[A\bigr] \geq 1-\frac{\text{Vol}(\man)}{\eta_d (\delta/4)^d}\bigg(1-\frac{\eta_d (\delta/2)^d}{\text{Vol}(\man)}\bigg)^N
    \end{equation}
    where $A$ denotes the set of all events where the $\delta$-sampling condition holds, as defined in \Cref{def: sampling condition}, and $\eta_d$ is the volume of a $d$-dimensional Euclidean ball.
\end{restatable}

\noindent\textit{Proof.} The proof will be an adaptation of the Sampling Lemma from \cite{bernstein2000graph}. We'll cover $\man$ with a finite family of geodesic balls of radius $\delta/2$. The sequence of centers $p_i$ are chosen such that
\[p_i \notin \bigcup_{j=1}^iB(p_j, \delta/2).\]
When this is no longer possible, we'll terminate. This process will result family of at most $\frac{\text{Vol}(\man)}{\eta_d (\delta/4)^d}$ elements, as each of the smaller balls $B(p_i, \delta/4)$ are necessarily disjoint. It follows that every $p \in \man$ belongs to some $B_i(p_i, \delta/2)$. The number of points in a given $B_i(p_i, \delta/2)$, call it $n_i$, is distributed as 
\[n_i \sim \text{Binomial}\bigg(N, \frac{\eta_d (\delta/2)^d}{\text{Vol}(\man)}\bigg)\]
and therefore,
\begin{align*}
    \mathbb{P}[B(p_i,\delta/2) \text{ empty}] &= \bigg(1-\frac{\eta_d (\delta/2)^d}{\text{Vol}(\man)}\bigg)^N
\end{align*}
where $\eta_d$ is the volume of a unit $d$-dimensional Euclidean ball; this stems from the assumption of a suitably small $\delta$. Now we can say
\begin{align*}
    \mathbb{P}\bigl[\text{no $B(p_i, \delta/2)$ is empty}\bigr] &= 1 - \mathbb{P}\Bigl[\text{there exists an empty }B(p_i, \delta/2)\Bigr]\\
    &\geq 1 - \sum_{i} \bigg(1-\frac{\eta_d (\delta/2)^d}{\text{Vol}(\man)}\bigg)^N \\
    &\geq 1-\frac{\text{Vol}(\man)}{\eta_d (\delta/4)^d}\bigg(1-\frac{\eta_d (\delta/2)^d}{\text{Vol}(\man)}\bigg)^N.
\end{align*}
\qed{}

\begin{restatable}{corollary}{sampling condition simplified}
\label{corollary: sampling condition}
    For $\delta = \Omega(N^{-1/2d})$ and $n \leq N^{1/2d}\cdot \text{diam}(\man)$ as defined in \Cref{lemma: similar pairs}, the bound in equation \Cref{eq: sampling condition} can be simplified to
    \begin{equation}\label{eq: sampling condition simplified} \mathbb{P}\bigl[A\bigr] \geq 1-\mathcal{O}(N^{1/2})e^{-\Omega(N^{1/2})}.\end{equation}
\end{restatable}
\noindent\textit{Proof.} First note that $\frac{\text{Vol}(\man)}{\eta_d (\delta/4)^d} = \mathcal{O}(N^{1/2})$. Now let's define 
\[f(N) = (1-\Omega(N^{-1/2}))^N.\]
We'll manipulate this to obtain an exponential upper bound as follows,
\begin{align*}
    \ln f(N) &= N\ln (1-\Omega(N^{-1/2})) \\
    &\leq -N\cdot\Omega(N^{-1/2}) \\
    &= -\Omega(N^{1/2})
\end{align*}
since $\ln(1-x) \leq -x$. Thus,
\begin{align*}
    f(N) &\leq e^{-\Omega(N^{1/2})}.
\end{align*}
\qed{}

\begin{restatable}{lemma}{eball overlap}
\label{lemma: eball overlap}
    Suppose we have two $d$-dimensional Euclidean $\epsilon$-balls $B_1$ and $B_2$ whose centers $p_1$ and $p_2$ are at distance $3\delta < \epsilon$. Let \[I = B_1 \, \cap B_2\] and \[U = \bigl(B_1 \, \cup B_2\bigr) \setminus\bigl(B_1 \, \cap B_2\bigr).\] Then we have \begin{equation} \label{eq: intersection volume}
    \text{Vol}(I) = \eta_d\epsilon^d \mathcal{I}_{[1 - (3\delta/2\epsilon)^2]}\Bigl(\frac{d+1}{2}, \frac{1}{2}\Bigr)
    \end{equation}
    and 
    \begin{equation} \label{eq: union volume} 
    \text{Vol}(U) = 2\eta_d\epsilon^d \biggl(1 - \mathcal{I}_{[1 - (3\delta/2\epsilon)^2]}\Bigl(\frac{d+1}{2}, \frac{1}{2}\Bigr)\biggr)
    \end{equation}
    where $\mathcal{I}$ is the regularized beta function and $\eta_d$ is the volume of the Euclidean unit ball of dimension $d$. 
\end{restatable}

\noindent\textit{Proof.} The balls must overlap since their centers are at distance less than $\epsilon$. Thus, the volume of $I$ must be twice the volume of a (hyper)spherical cap with radius $\epsilon$ and height $\epsilon - 3\delta/2.$ Using the derivation from \cite{li2010concise}, we can say
\[\text{Vol}(I) = \eta_d\epsilon^d \mathcal{I}_{[1 - (3\delta/2\epsilon)^2]}\Bigl(\frac{d+1}{2}, \frac{1}{2}\Bigr)\]
where $\mathcal{I}_x(a,b)$ is the regularized incomplete beta function and $\eta_d$ is the volume of the $d$-dimensional Euclidean unit ball. To obtain the volume of $U$, observe that
\[\text{Vol}(U) = 2\eta_d\epsilon^d - 2\text{Vol}(I)\]
which follows from simple set algebra. Plugging in, we find
\[\text{Vol}(U) = 2\eta_d\epsilon^d \biggl(1 - \mathcal{I}_{[1 - (3\delta/2\epsilon)^2]}\Bigl(\frac{d+1}{2}, \frac{1}{2}\Bigr)\biggr).\]
\qed{}

\begin{restatable}{corollary}{monotonicity of ball volumes}
\label{corollary: monotonicity of ball volumes}
    \Cref{eq: intersection volume} is decreasing in $\delta$, while \Cref{eq: union volume} is increasing in $\delta$.
\end{restatable}

\noindent\textit{Proof.} Since the regularized beta function $\mathcal{I}_x(\cdot,\cdot)$ is the cumulative distribution function for the beta distribution it is necessarily increasing in $x$. Therefore $\mathcal{I}_{[1 - (3\delta/2\epsilon)^2]}(\cdot,\cdot)$ must be decreasing in $\delta$. This renders \Cref{eq: intersection volume} decreasing in $\delta$. On the other hand, this renders \Cref{eq: union volume} increasing in $\delta$.

\qed{}

\dissimilarpairs*

\noindent\textit{Proof.} Observe that the edge set $E$ can be written as $E = E_{\text{orig}} \, \cup E_{\text{noisy}}$, where $E_{\text{orig}}$ connects all points with pairwise distance $\leq \epsilon$, and $E_{\text{noisy}}$ connects random pairs according to the noise model described in \Cref{sec: noise model}. By \Cref{assumption: cluster separation}, we know that all connected component bridging edges are contained in $E_{\text{noisy}}$, while \textit{none} are contained in $E_{\text{orig}}$. This allows us to say,
\begin{align*}
    \mathbb{P}\Big[\kappa(a,b) < -1/2 \,\, \forall \, (a,b) \in E \text{ s.t. } & C(a) \neq C(b) \, \Big| \, X \Big] \\ &= \mathbb{P}\Big[\kappa(a,b) < -1/2 \,\, \forall \, (a,b) \in E_{\text{noisy}}  \text{ s.t. }C(a) \neq C(b) \Big| \, X  \Big].
\end{align*}
Note that the configuration of points $X$ is fixed, and the stochasticity stems from the random edge addition process from \Cref{eq: noise model}; this means that $E_{\text{noisy}}$ is a non-deterministic set. Furthermore, $\kappa(a,b)$ is deterministic given $E_{\text{noisy}}$. To analyze this situation we'll manipulate the term above as follows,
\begin{align*}
    \mathbb{P}\Big[\kappa(a,b) < -1/2 \,\, \forall \, (a,b) \in E_{\text{noisy}} &  \text{ s.t. }C(a) \neq C(b)\, \Big| \, X  \Big] \\ &= 1 - \mathbb{P}\Big[\kappa(a,b) \geq -1/2 \,\, \text{for any} \, (a,b) \in E_{\text{noisy}}    \text{ s.t. }C(a) \neq C(b)\, \Big| \, X  \Big].
\end{align*}
Now, let's define $P_{\text{bridge}} = \{(x_i,x_j) \in X\times X \,|\, C(x_i) \neq C(x_j) \}$ and define $K^{E}: P_{\text{bridge}}\rightarrow[-2,1]$ as follows,
\begin{equation}\label{eq: Kappa}
    K^{E}(a,b) = \begin{cases}
    \kappa(a,b) & (a,b) \in E \\
    -1 & \text{otherwise}
\end{cases}
\end{equation}
which allows us to say,
\begin{align*}
    \mathbb{P}\Big[\kappa(a,b) < -1/2 \,\, \forall \, (a,b) \in E_{\text{noisy}} & \text{ s.t. } C(a) \neq C(b) \, \Big| \, X  \Big] \\ &= 1 - \mathbb{P}\Big[\kappa(a,b) \geq -1/2 \,\, \text{for any} \, (a,b) \in E_{\text{noisy}}    \text{ s.t. }C(a) \neq C(b)\, \Big| \, X  \Big] \\
    &= 1 - \mathbb{P}\Big[K^{E_{\text{noisy}}}(a,b) \geq -1/2 \,\, \text{for any} \, (a,b) \in E_{\text{noisy}}  \text{ s.t. }C(a) \neq C(b) \, \Big| \, X  \Big] \\
    &= 1 - \mathbb{P}\Big[K^{E_{\text{noisy}}}(a,b) \geq -1/2 \,\, \text{for any} \, (a,b) \in P_{\text{bridge}}  \, \Big| \, X  \Big].
\end{align*}
Observe that the randomness in the expression above still comes from $E_{\text{noisy}}$. Now we'll continue to rewrite,
\begin{flalign*}
    \mathbb{P}\Big[\kappa(a,b) < -1/2 \,\, \forall \, (a,b) \in E_{\text{noisy}}  \, \Big| \, X  \Big] &= 1 - \mathbb{P}\Bigg[\bigcup_{(a,b) \in P_{\text{bridge}}} K^{E_{\text{noisy}}}(a,b) \geq -1/2 \, \Big| \, X  \Bigg] \\
    &\geq 1 - \sum_{(a,b) \in P_{\text{bridge}}}\mathbb{P}\Big[ K^{E_{\text{noisy}}}(a,b) \geq -1/2 \, \Big| \, X  \Big] && \text{(by union bound)} \\
    &=1 - \sum_{(a,b) \in P_{\text{bridge}}}\biggl( 1- \mathbb{P}\Big[ K^{E_{\text{noisy}}}(a,b) < -1/2 \, \Big| \, X  \Big]\biggr) \\
    &> 1 - \sum_{(a,b) \in P_{\text{bridge}}}2e^{-N_{\min}s_{\max}} && \text{(by \Cref{lemma: fixed configuration negative curvature bound})}
\end{flalign*}
where $N_{\min} = \min_{p \in X}|\{(B(p, \epsilon)\setminus{{p}}) \bigcap X\}|$, which is the minimum number of neighbors of any point in $X$ \textit{before} noisy edge additions. Since $P_{\text{bridge}} \subset X \times X$, we have

\begin{align*}
    \mathbb{P}\Big[\kappa(a,b) < -1/2 \,\, \forall \, (a,b) \in E_{\text{noisy}}  \, \Big| \, X  \Big] &\geq 1-2|X|^2e^{-N_{\min}s_{\max}}.
\end{align*}
\qed{}

\begin{restatable}[]{lemma}{fixed configuration negative curvature bound}
\label{lemma: fixed configuration negative curvature bound}
    Let $X \subset \man$ be a point cloud sampled from $\man$. Furthermore, let $G=(X,E)$ be the nearest neighbor graph created from the data using $\epsilon$-radius connectivity, with noisy edge additions according to the noise scheme described in \Cref{eq: noise model}. Suppose we have $a, b \in X$ where $C(a) \neq C(b)$. Then if $s_{\max} \leq \frac{3}{8e}$ we have
    \[\mathbb{P}\Bigl[K^{E}(a,b) < -1/2  \, \Big| \, X \Bigr] > 1 - 2e^{-N_{\min}s_{\max}}\]
    where $K^E$ is defined in \Cref{eq: Kappa} and $N_{\min} = \min_{p \in X}|\{(B(p, \epsilon)\setminus{{p}}) \bigcap X\}|$, which is the minimum number of neighbors of any point in $X$ \textit{before} noisy edge additions.
\end{restatable}

\noindent\textit{Proof.} Breaking up the left-hand side of the expression above results in
\begin{align*}
    \mathbb{P}\Bigl[K^{E}(a,b) < -1/2  \, \Big| \, X \Bigr] &= \mathbb{P}\Bigl[K^{E}(a,b) < -1/2, (a,b)\in E  \, \Big| \, X \Bigr] + \mathbb{P}\Bigl[K^{E}(a,b) < -1/2, (a,b)\notin E  \, \Big| \, X \Bigr] \\
    \begin{split}
        &= \mathbb{P}\Bigl[K^{E}(a,b) < -1/2  \, \Big| \, X, (a,b)\in E \Bigr] \mathbb{P}\Bigl[(a,b)\in E \, \Big| \, X\Bigr] 
        \\ &\hspace{20mm}+ \underbrace{\mathbb{P}\Bigl[K^{E}(a,b) < -1/2  \, \Big| \, X, (a,b)\notin E \Bigr]}_{1} \mathbb{P}\Bigl[(a,b)\notin E \, \Big| \, X\Bigr].
    \end{split}
\end{align*}
Defining $p = \mathbb{P}\Bigl[(a,b)\in E \, \Big| \, X\Bigr]$ we see 
\begin{align*}
    \mathbb{P}\Bigl[K^{E}(a,b) < -1/2  \, \Big| \, X \Bigr] &= \mathbb{P}\Bigl[K^{E}(a,b) < -1/2  \, \Big| \, X, (a,b)\in E \Bigr] \cdot p + 1-p \\
    &= 1 - p\Bigg(1 -\mathbb{P}\Bigl[K^{E}(a,b) < -1/2  \, \Big| \, X, (a,b)\in E \Bigr]\Bigg) \\
    &\geq 1 - \Bigg(1 -\mathbb{P}\Bigl[K^{E}(a,b) < -1/2  \, \Big| \, X, (a,b)\in E \Bigr]\Bigg) \\
    &= \mathbb{P}\Bigl[K^{E}(a,b) < -1/2  \, \Big| \, X, (a,b)\in E \Bigr] \\
    &= \mathbb{P}\Bigl[\kappa(a,b) < -1/2  \, \Big| \, X, (a,b)\in E \Bigr].
\end{align*}
Thus, it suffices to find a lower bound for $\mathbb{P}[\kappa(a,b) < -1/2  \, | \, X, (a,b)\in E]$. Observe that any edge connecting a point in $C(a)$ and a point in $C(b)$ must arise from the adjacency noise model described in \Cref{sec: noise model} due to \Cref{assumption: cluster separation}. Let's define the random sets
\[\mathcal{N}(a) = \Bigl\{ i \in X \, \Big| \, d_G(i,a) = 1, i \neq b \Bigr\}\]
and 
\[\mathcal{N}(b) = \Bigl\{ j \in X \, \Big| \, d_G(j,b) = 1, j \neq a \Bigr\}\]
where $G$ denotes the graph \textit{after} the corruption process. Let's also define the random sets $S_a = \{x \in \mathcal{N}(a) \, | \,\nu_x =0 \}$ and $S_b = \{x \in \mathcal{N}(b) \, | \,\nu_x =0 \}$. Observe that for all $x,y \in S_a\times S_b$, we necessarily have $d_G(x,y) \geq 3$. Furthermore, let's define the ratios
\[\delta_a = \frac{|S_a|}{|\mathcal{N}(a)|}, \qquad \delta_b = \frac{|S_b|}{|\mathcal{N}(b)|}\]
and $\delta = \delta_a + \delta_b$. Observe that 
\begin{align}
    \delta &= \frac{|S_a|}{|\mathcal{N}(a)|} + \frac{|S_b|}{|\mathcal{N}(b)|} \\
    &= \frac{\sum_{i \in \mathcal{N}(a)} (1-\nu_i)}{|\mathcal{N}(a)|} + \frac{\sum_{i \in \mathcal{N}(b)} (1-\nu_i)}{|\mathcal{N}(b)|} \\
    &= 2-\frac{\sum_{i \in \mathcal{N}(a)} \nu_i}{N_a} -\frac{\sum_{i \in \mathcal{N}(b)} \nu_i}{N_b}
\end{align}
where $N_a = |\mathcal{N}(a)|$, $N_b = |\mathcal{N}(b)|$. Let's also define $N_{ab} = \big|\mathcal{N}(a)\cup\mathcal{N}(b)\big|$. Now, we'll consider the behavior of $\delta$.  Note that the number of elements of $\mathcal{N}(a)$ and $\mathcal{N}(b)$ are stochastic, rendering $N_{a}$, $N_b$ and $N_{ab}$ random variables. Furthermore, observe that if the $s_i$'s (the probabilities associated with each $\nu_i$) were all equivalent, then we would have
\[\Bigg(\sum_{i \in \mathcal{N}(a)}\nu_i \, \Bigg| \,  N_{a}\Bigg) \sim \text{Binomial}(N_{a}, s)\] 
and
\[\Bigg(\sum_{i \in \mathcal{N}(b)}\nu_i \, \Bigg| \,  N_{b}\Bigg) \sim \text{Binomial}(N_{b}, s)\] 
with $s = s_i$. This is unfortunately not the case, but if we define the random variables 
\[Z_a \, \Big| \,N_{a} \sim \text{Binomial}(N_{a}, s_{\max})\,\]
and
\[Z_b \, \Big| \,N_{b} \sim \text{Binomial}(N_{b}, s_{\max})\,\]
using $s_{\max} = \max_{p \in \man} s_p$, with $s_p$ defined in \Cref{eq: noise model}, then we can say
\[\mathbb{P}\Biggl[\sum_{i \in \mathcal{N}(a)}\nu_i < m \, \Bigg| \, N_{a}, X\Biggr] \geq \mathbb{P}[Z_a < m \, | \, N_{a}]\]
and
\[\mathbb{P}\Biggl[\sum_{i \in \mathcal{N}(b)}\nu_i < m \, \Bigg| \, N_{b}, X\Biggr] \geq \mathbb{P}[Z_b < m \, | \, N_{b}].\]
We'll analyze the inequality concerning $Z_a$ first. If we apply a Chernoff bound to the right-hand-side \cite{tsun2020probability} we obtain,
\begin{align*}
    \mathbb{P}\Biggl[\sum_{i \in \mathcal{N}(a)}\nu_i < m\, \Bigg| \, N_{a}, X\Biggr] &\geq 1 - e^{-\mathbb{E}[Z_a|N_{a}]}\biggl(\frac{\mathbb{E}[Z_a|N_{a}]\cdot e}{m}\biggr)^m \\
    &= 1 - e^{-N_{a}s_{\max}}\biggl(\frac{N_{a}s_{\max}\cdot e}{m}\biggr)^m
\end{align*}
for $m > N_{a}s_{\max}$. Choosing $m = N_as_{\max}e$ yields
\begin{align*}
    \mathbb{P}\Biggl[\sum_{i \in \mathcal{N}(a)}\nu_i < N_a s_{\max}e\, \Bigg| \, N_{a}, X\Biggr] &\geq 1 - e^{-N_{a}s_{\max}}.
\end{align*}
Applying the same argument for the inequality concerning $Z_b$ yields
\begin{align*}
    \mathbb{P}\Biggl[\sum_{i \in \mathcal{N}(b)}\nu_i < N_b s_{\max}e\, \Bigg| \, N_{b}, X\Biggr] &\geq 1 - e^{-N_{b}s_{\max}}.
\end{align*}
Now we'll apply an intersection bound to say 
\begin{align*}
    \mathbb{P}\Biggl[\sum_{i \in \mathcal{N}(a)}\nu_i < N_a s_{\max}e\,, \sum_{i \in \mathcal{N}(b)}\nu_i < N_b s_{\max}e \, \Bigg| \, N_{a}, N_{b}, X\Biggr] &\geq 1 - e^{-N_{a}s_{\max}} - e^{-N_b s_{\max}}.
\end{align*}
Bringing back the definition of $\delta$, we have
\begin{align*}
    \mathbb{P}\Biggl[\delta \geq 2 - 2s_{\max}e \, \Bigg| \, N_{a}, N_{b}, X\Biggr] &\geq 1 - e^{-N_{a}s_{\max}} - e^{-N_b s_{\max}}.
\end{align*}
Now we'll define $N_{\min} = \min_{p \in X}|\{(B(p, \epsilon)\setminus{{p}}) \bigcap X\}|$, which is the minimum number of neighbors of any point in $X$ \textit{before} noisy edge additions. We can use this to deterministically lower bound the right hand side independent of $N_a$ and $N_b$,
\begin{align*}
    \mathbb{P}\Biggl[\delta \geq 2 - 2s_{\max}e \, \Bigg| \, X\Biggr] &\geq 1 - 2e^{-N_{\min}s_{\max}}.
\end{align*}
Now if we apply \Cref{lemma: neighborhood overlap negative curvature bound}, we obtain
\begin{align*}
    1 - 2e^{-N_{\min}s_{\max}}  &\leq \mathbb{P}\Biggl[\kappa(a,b) \leq -2 + 2(2- (2 - 2s_{\max}e)) \, \Bigg| \, X\Biggr] \\
    &= \mathbb{P}\Biggl[\kappa(a,b) \leq -2 + 4(1- (1 - s_{\max}e)) \, \Bigg| \, X\Biggr] \\
    &= \mathbb{P}\Biggl[\kappa(a,b) \leq -2 + 4s_{\max}e \, \Bigg| \, X\Biggr].
\end{align*}
Now under the assumption that $s_{\max} \leq \frac{3}{8e}$ then we have $s_{\max} \leq 3/8e$. This follows from the noise model described in \Cref{eq: noise model}. In this case, we have 
\begin{align*}
    \mathbb{P}\Biggl[\kappa(a,b) \leq -1/2 \, \Bigg| \, X\Biggr] &\geq 1 - 2e^{-N_{\min}s_{\max}}.
\end{align*}
\qed{}

\begin{restatable}[Adapted from Lemma A.1 from \cite{saidi2025recovering}]{lemma}{neighborhood overlap negative curvature bound}
\label{lemma: neighborhood overlap negative curvature bound}
    Consider an edge $(a,b)$ in an unweighted graph $G=(V,E)$, and consider the edge-excluded neighborhoods of each endpoint,
    \[\mathcal{N}(a) = \Bigl\{ i \in V \, \Big| \, d_G(i,a) = 1, i \neq b \Bigr\}\]
    and 
    \[\mathcal{N}(b) = \Bigl\{ j \in V \, \Big| \, d_G(j,b) = 1, j \neq a \Bigr\}.\]
    Suppose there exist subsets $S_a \subseteq \mathcal{N}(a)$ and $S_b \subseteq \mathcal{N}(b)$ such that for all $(x,y) \in S_a \times S_b$ we have $d_G(x,y) \geq 3$. Defining
    \[\delta_a = \frac{|S_x|}{\big|\mathcal{N}(a)\big|} \qquad \delta_b = \frac{|S_x|}{\big|\mathcal{N}(b)\big|}.\]
    Then we have that 
    \[\kappa(a,b) \leq -1 + 2\big(2-(\delta_a+\delta_b)\big).\]
\end{restatable}

\noindent\textit{Proof.}  To bound the ORC we need to bound the $1$-Wasserstein distance between $\mu_a$ and $\mu_b$, where $\mu_a$ is a uniform measure on the set $\mathcal{N}(a)$ and $\mu_b$ is a uniform measure on the set $\mathcal{N}(b)$. Let's define $\hat{\mu}_a$ and $\hat{\mu}_b$ as uniform probability measures over $S_a$ and $S_b$ respectively. We can bound the $1$-Wasserstein distance between $\mu_a$ and $\mu_b$ as
\[W(\mu_a, \mu_b) \geq W(\hat{\mu}_a, \hat{\mu}_b) - W(\hat{\mu}_a, \mu_a) - W(\hat{\mu}_b, \mu_b).\]
By assumption, we know that $d_G(x,y) \geq 2$ for all $x \in S_a$ and for all $y \in S_b$. Thus, a lower bound on the first term follows, 
\[W(\hat{\mu}_a, \hat{\mu}_b) \geq 3.\]
Now, we would like to bound $W(\hat{\mu}_a, \mu_a)$ from above. There is $1/|\mathcal{N}(a)|$ mass on each node $x \in \text{supp}(\mu_a)$, while there is $1/\delta_a|\mathcal{N}(a)|$ mass on each $x' \in \text{supp}(\hat{\mu}_a)$. We can define a feasible transport plan that transports all excess mass on $x' \in \text{supp}(\hat{\mu}_a)$ to the nodes $\text{supp}(\mu_a) \setminus \text{supp}(\hat{\mu}_a)$. Since the Wasserstein distance minimizes over all possible transport plans, the Wasserstein cost for this transport plan will upper bound the true distance. Observe that the excess mass on any $x' \in \text{supp}(\hat{\mu}_a)$ is exactly 
\[\frac{1}{\delta_a|\mathcal{N}(a)|} - \frac{1}{|\mathcal{N}(b)|}\]
which means the total mass that needs to be transported is 
\[\delta_a|\mathcal{N}(a)|\Biggl(\frac{1}{\delta_a|\mathcal{N}(a)|} - \frac{1}{|\mathcal{N}(a)|}\Biggr) = 1 - \delta_a.\]
We also know that from any $x' \in \text{supp}(\hat{\mu}_a)$ to any $x \in \text{supp}(\mu_a)$ there exists a length $2$ path through the node $x$. Therefore, $d_G(a, a') \leq 2$. We can then conclude 
\[W(\hat{\mu}_a, \mu_a) \leq 2\bigl(1-\delta_a\bigr).\]
With the same argument, the following bound can also be derived, 
\[W(\hat{\mu}_b, \mu_b) \leq 2\bigl(1-\delta_b\bigr).\]
Putting it all together,
\[W(\mu_a, \mu_b) \geq 3 - 2\big(2 - (\delta_a + \delta_b)\big).\]
Solving for $\kappa(a,b)$, 
\begin{align*}
    \kappa(a,b) &\leq 1 - 3 - 2\big(2 - (\delta_a + \delta_b)\big)\\
    &= -2 + 2\Bigl(2 - (\delta_a + \delta_b)\Bigr).
\end{align*}

\qed{}

\begin{restatable}[]{lemma}{limit of expectation} \label{lemma: limit of expectation}
    Suppose $X$ consists of $N$ i.i.d. samples from a uniform distribution over $\man$. Then we have,
    \[\mathbb{E}_{X \sim \text{Uniform}(\man)}\biggl[ |X|^2e^{-N_{\min}s_{\max}}\biggr] = \mathcal{O}(N^4)e^{-\Omega(N)}\]
    where $N_{\min} = \min_{p \in X}|\{(B(p, \epsilon)\setminus{{p}}) \bigcap X\}|$, which is the minimum number of neighbors of any point in $X$ before noisy edge additions. 
\end{restatable}
\noindent \textit{Proof.} Pulling out the constant yields,
\begin{align*}
    \mathbb{E}_{X \sim \text{Uniform}(\man)}\biggl[ |X|^2e^{-N_{\min}s_{\max}}\biggr] &= N^2\cdot \mathbb{E}_{X \sim \text{Uniform}(\man)}\Big[e^{-N_{\min}s_{\max}}\Big]. \\
\end{align*}
Let $Z$ be a random variable distributed as $\text{Binomial}(N, V_{\min}/\text{Vol}(\man))$, where $V_{\min} := \min_{p \in \man} \text{Vol}(B(p, \epsilon))$. We have $V_{\min} > 0$ because $M$ is compact. Note that we have,
\begin{align*}
    \mathbb{E}_{x_1, \dots, x_N \sim \text{Uniform}(\man) }\Big[e^{-N_{\min}s_{\max}}\Big] &= \sum_{n=0}^N \mathbb{P}[N_{\min} = n] \cdot e^{-ns_{\max}} \\
    &\leq \mathcal{O}(N) \sum_{n=0}^N \mathbb{P}[Z \leq n] \cdot e^{-ns_{\max}}  && \text{(by \Cref{prop: minimum neighborhood size})}\\
    &= \mathcal{O}(N) \sum_{n=0}^N \sum_{n'=0}^n\mathbb{P}[Z = n'] \cdot e^{-ns_{\max}}\\
    &= \mathcal{O}(N) \sum_{n'=0}^N \sum_{n=n'}^N\mathbb{P}[Z = n'] \cdot e^{-ns_{\max}}\\
    &= \mathcal{O}(N) \sum_{n'=0}^N \mathbb{P}[Z = n']\sum_{n=n'}^N e^{-ns_{\max}}\\
    &\leq \mathcal{O}(N) \sum_{n'=0}^N \mathbb{P}[Z = n']\cdot(N-n')e^{-n's_{\max}} \\
    &\leq \mathcal{O}(N^2) \sum_{n'=0}^N \mathbb{P}[Z = n']e^{-n's_{\max}} \\
    &= \mathcal{O}(N^2)\cdot\mathbb{E}[e^{-Z s_{\max}}]
\end{align*}
We can now evaluate the expectation above using the moment generating function of the binomial distribution, evaluated at $t = -s_{\max}$,
\begin{align*}
    \mathbb{E}_{x_1, \dots, x_N \sim \text{Uniform}(\man) }\Big[e^{-N_{\min}s_{\max}}\Big] &\leq \mathcal{O}(N^2)\cdot\bigg(1-\frac{V_{\min}}{\text{Vol}(\man)} + \frac{V_{\min}}{\text{Vol}(\man)}e^{-s_{\max}}\bigg)^N \\
    &= \mathcal{O}(N^2)\cdot C^{-N}
\end{align*}
for some $C > 1$. Plugging back in, 
\begin{align*}
    \mathbb{E}_{X \sim \text{Uniform}(\man)}\biggl[ |X|^2e^{-N_{\min}s_{\max}}\biggr] &\leq \mathcal{O}(N^4)\cdot C^{-N} \\
    &= \mathcal{O}(N^4)\cdot e^{-N\ln C} \\
    &= \mathcal{O}(N^4)e^{-\Omega(N)}.
\end{align*}
\qed{}

\begin{restatable}{lemma}{lambda asymptotics}
\label{prop: p_i}
    For $p_i$ defined in \Cref{eq: p_i} and $\delta = \epsilon N^{-1/2d}/3$ we have that $p_i = 1-\mathcal{O}(N^{-1/2d})$.
\end{restatable}
\noindent\textit{Proof.} 
Plugging in the expressions, we see
\begin{align*}
    p_i &=  \frac{\mathcal{I}_{[1 - (3\delta/2\epsilon)^2]}\Bigl(\frac{d+1}{2}, \frac{1}{2}\Bigr)}{2- \mathcal{I}_{[1 - (3\delta/2\epsilon)^2]}\Bigl(\frac{d+1}{2}, \frac{1}{2}\Bigr)} \\
    &\geq \bigg(\mathcal{I}_{[1 - (3\delta/2\epsilon)^2]}\Bigl(\frac{d+1}{2}, \frac{1}{2}\Bigr)\bigg)^2 
\end{align*}
because $x/(2-x) \geq x^2$ for $x \in [0,1]$. Now we'll apply the identity $\mathcal{I}_x(a,b) = 1 - \mathcal{I}_{1-x}(b,a)$ to say
\begin{align*}
    p_i &\geq \bigg(1 - \mathcal{I}_{[ (3\delta/2\epsilon)^2]}\Bigl(\frac{1}{2}, \frac{d+1}{2}\Bigr)\bigg)^2 \\
    &\geq 1-2\mathcal{I}_{[ (3\delta/2\epsilon)^2]}\Bigl(\frac{1}{2}, \frac{d+1}{2}\Bigr)\\
    &= 1 - \frac{2}{B\big(1/2, (d+1)/2\big)}B_{(3\delta/2\epsilon)^2}\bigg(1/2, \frac{d+1}{2}\bigg)
\end{align*}
where $B_x(a,b)$ is the incomplete beta function and $B(a,b)$ is the standard beta function \cite{NIST:DLMF}. Now we'll apply another identity
\[B_x(a,b) = \frac{x^a}{a}F(a, 1-b, a+1; x)\]
(where $F$ is the hypergeometric function) to obtain,
\begin{align*}
    p_i &\geq1 - \frac{2}{B\big(1/2, (d+1)/2\big)}\cdot\frac{3\delta/2\epsilon}{1/2}F\bigg(1/2, 1-\frac{d+1}{2}, 3/2; (3\delta/2\epsilon)^2\bigg) \\
    &= 1-\delta C \cdot F\bigg(\frac{1}{2}, 1-\frac{d+1}{2}, \frac{3}{2}; (3\delta/2\epsilon)^2\bigg)
\end{align*}
where $C$ is a constant independent of $\delta$ and $N$ \cite{NIST:DLMF}. Now we'll use the integral representation of the hypergeometric function from \cite{NIST:DLMF} to say
\begin{align*}
    p_i &\geq 1-\delta C \cdot\frac{1}{\Gamma\Big(1-\frac{d+1}{2}\Big)\Gamma\Big(\frac{1}{2} +\frac{d+1}{2}\Big)} \bigintsss_{0}^{1}\frac{t^{-(d+1)/2}(1-t)^{d/2}}{(1-(3\delta/2\epsilon)^2t)^{1/2}}dt \\
    &\geq 1-\delta C \cdot\Bigg|\frac{1}{\Gamma\Big(1-\frac{d+1}{2}\Big)\Gamma\Big(\frac{1}{2} +\frac{d+1}{2}\Big)}\Bigg| \cdot \Bigg|\bigintsss_{0}^{1}\frac{t^{-(d+1)/2}(1-t)^{d/2}}{(1-(3\delta/2\epsilon)^2t)^{1/2}}dt\Bigg|. 
\end{align*}
Since $3\delta/2\epsilon =  N^{-1/2d}/2 < 1/2$ for any $N > 0$ we have 
\begin{align*}
    p_i &\geq 1-\delta C \cdot\Bigg|\frac{1}{\Gamma\Big(1-\frac{d+1}{2}\Big)\Gamma\Big(\frac{1}{2} +\frac{d+1}{2}\Big)}\Bigg|\cdot \Bigg|\bigintsss_{0}^{1}\frac{t^{-(d+1)/2}(1-t)^{d/2}}{(1-0.25t)^{1/2}}dt\Bigg| \\
    &= 1- \mathcal{O}(\delta) \\
    &= 1 - \mathcal{O}(N^{-1/2d}).
\end{align*}
\qed{}

\begin{restatable}[]{lemma}{minimum neighborhood size} \label{prop: minimum neighborhood size}
    Let $X = x_1, \dots, x_N \sim \text{Uniform}(\man)$. Let $N_{\min} = \min_{p \in X} |\{(B(p, \epsilon)\setminus{{p}}) \bigcap X\}|$, which is the minimum number of neighbors of any point in $X$. Also let 
    \[V_{\min} = \min_{p \in \man} \text{Vol}(B(p, \epsilon)).\] 
    Then we have,
    \[\mathbb{P}\big[N_{\min} = n\big] \leq N \cdot \mathbb{P}[Z \leq n]\]
    where $Z$ is a random variable with a binomial distribution of $N$ trials with success probability $V_{\min}/\text{Vol}(\man)$, and $V_{\min} = \min_{p \in \man} \text{Vol}(B(p, \epsilon))$. 
\end{restatable}
\noindent \textit{Proof.} Rewriting the left-hand side, we obtain
\begin{align*}
    \mathbb{P}\big[N_{\min} = n\big] &\leq \mathbb{P}\big[N_{\min} \leq n\big]\\
    &=\mathbb{P}\Big[\bigcup_{i}n_{i} \leq n \Big] \\
    &\leq \sum_{i=1}^{N} \mathbb{P}\Big[n_{i} \leq n \Big] && \text{(by union bound) }
\end{align*}
where $n_i = |\{(B(x_i, \epsilon)\setminus{{x_i}}) \bigcap X\}|$. Observe that each $n_i \sim \text{Binomial}\Big(N-1, \frac{\text{Vol}(B(x_i, \epsilon))}{\text{Vol}(\man)}\Big)$. Defining the random variable $Z \sim \text{Binomial}\Big(N-1, \frac{V_{\min}}{\text{Vol}(\man)}\Big)$, where $V_{\min} = \min_{p \in \man} \text{Vol}(B(p, \epsilon))$, allows us to say
\[\mathbb{P}\Big[n_{i} \leq n \Big] \leq \mathbb{P}\Big[Z \leq n \Big]\]
for all $i$. Thus,
\begin{align*}
    \mathbb{P}\big[N_{\min} = n\big] &\leq N\cdot\mathbb{P}\Big[Z \leq n \Big].
\end{align*}
\qed{}

\end{document}